\documentclass[journal]{IEEEtran}
\usepackage{booktabs}
\usepackage{url}
\usepackage{epsfig}
\usepackage{subfigure}
\usepackage{graphics}
\usepackage{graphicx}
\usepackage{formular}
\usepackage{algorithm}
\usepackage{algorithmic}
\usepackage{multirow}
\usepackage{multicol}
\usepackage{xcolor}
\usepackage{clrscode}
\usepackage{makecell}
\usepackage{CJK}
\usepackage{amsthm}
\usepackage{balance}
\usepackage{times}
\usepackage{amsmath}
\usepackage{amsfonts}
\usepackage{stfloats}
\usepackage{array}
\usepackage{float}
\usepackage{url}
\usepackage{epstopdf}

\begin{document}
\title{Restricted Boltzmann Machines with Gaussian Visible Units Guided by Pairwise Constraints}
\author{Jielei Chu, Hongjun Wang, Hua Meng, Peng Jin and Tianrui Li,~\IEEEmembership{Senior member,~IEEE} 

\thanks{Jielei Chu, Hongjun Wang (Corresponding author), Hua Meng and Tianrui Li are with the School of Information Science and Technology, Southwest Jiaotong University, Chengdu, 611756, Sichuan, China. e-mails: \{jieleichu, wanghongjun, huameng, trli\}@swjtu.edu.cn.}
\thanks{Peng Jin is with the School of Computer Science, Leshan Normal University, 614000, Leshan, China. e-mail: jandp@pku.edu.cn}
}
\maketitle
\begin{abstract}
Restricted Boltzmann machines (RBMs) and their variants are usually trained by contrastive divergence (CD) learning, but the training procedure is an unsupervised learning approach, without any guidances of the background knowledge. To enhance the expression ability of traditional RBMs, in this paper, we propose pairwise constraints restricted Boltzmann machine with Gaussian visible units (pcGRBM) model, in which the learning procedure is guided by pairwise constraints and the process of encoding is conducted under these guidances. The pairwise constraints are encoded in hidden layer features of pcGRBM. Then, some pairwise hidden features of pcGRBM flock together and another part of them are separated by the guidances. In order to deal with real-valued data, the binary visible units are replaced by linear units with Gausian noise in the pcGRBM model. In the learning process of pcGRBM, the pairwise constraints are iterated transitions between visible and hidden units during CD learning procedure. Then, the proposed model is inferred by approximative gradient descent method and the corresponding learning algorithm is designed. In order to compare the availability of pcGRBM and traditional RBMs with Gaussian visible units, the features of the pcGRBM and RBMs hidden layer are used as input `data' for K-means, spectral clustering (SP) and affinity propagation (AP) algorithms, respectively. We also use 10-fold cross-validation strategy to train and test pcGRBM model to obtain more meaningful results with pairwise constraints which are derived from incremental sampling procedures. A thorough experimental evaluation is performed with twelve image datasets of Microsoft Research Asia Multimedia (MSRA-MM). The experimental results show that the clustering performance of K-means, SP and AP algorithms based on pcGRBM model are significantly better than traditional RBMs. In addition, the pcGRBM model for clustering tasks shows better performance than some semi-supervised clustering algorithms.\\
\end{abstract}
\IEEEpeerreviewmaketitle
\section{Introduction}
 Feature learning is presently the subject of an active research. The restricted Blotzmann machine (RBM)\cite{hinton1986learning} as an energy-based modeling paradigm is one of the most popular feature extraction models. The RBM has no lateral connections among nodes in each layer, then its learning procedure becomes much more efficient than general Blotzmann machine. So, it has powerful representation capability to obtain deep features of original data. Because of these advantages, there has been extensive research into the RBM since Hinton proposed fast learning algorithms by contrastive divergence (CD) learning\cite{hinton2002training}, \cite{carreira2005contrastive}. Several power and tractability deep networks was proposed, including deep belief networks\cite{hinton2006fast}, deep autoencoder\cite{bengio2007greedy}, deep Boltzmann machine\cite{salakhutdinov2012efficient}, deep dropout neural net\cite{srivastava2014dropout}. Until now, a large number of  successful applications built on the RBMs have appeared, e.g., classification\cite{krizhevsky2012imagenet}, \cite{ciresan2012multi}, \cite{finkfuzzy2015fuzzyclassification}, \cite{chenspectral2015SpatialClassificationDBF}, \cite{elfwing2015expected}, feature learning\cite{bengio2013representation}, facial recognition\cite{teh2001rate}, collaborative filtering\cite{vincent2010stacked}, topic modelling\cite{hinton2009replicated}, speech recognition\cite{graves2013speech}, natural language understanding\cite{sarikaya2014application}, computer vision\cite{nie2015generative}, dimensionality reduction\cite{hinton2006reducing}, voice conversion\cite{nakashika2015voice}, musical genre categorization\cite{yang2011deep}, real-time key point recognition\cite{yuan2014real}, periocular recognition\cite{nie2014periocular} and time series forecasting\cite{Kuremoto2012Time}, \cite{Hirata2016Time}, \cite{Kuremoto2014Time}.\\
 \indent However, since the learning procedures of classic RBM and its variants are unsupervised methods, their processes of feature extraction are non-directional and conducted under no guidance. To remedy these weakness, this paper proposes a pairwise constraints restricted Blotzmann machine with Gaussian visible units (pcGRBM) and the corresponding learning algorithm, where the feature learning procedure is guided by pairwise constraints which come from labels. In pcGRBM model, the pairwise constraints which are instance-level prior knowledge guide the process of encoding. Some pairwise hidden features of pcGRBM flock together and another part of them are separated by the guidance. Then the process of feature extraction is no longer non-directional. So, the background knowledge of instance-level pairwise constraints are encoded in hidden layer features of pcGRBM. In order to testify the availability of pcGRBM, we design three structures of clustering, in which the features of the hidden layer of the pcRBM are used as input `data' for unsupervised clustering algorithms. The experimental results show that the clustering performance of K-means\cite{Lloyd1982Least}, SP\cite{ng2002spectral} and AP\cite{frey2007clustering} algorithms based on pcGRBM model are significantly better than traditional RBMs. In addition, the pcGRBM model for clustering is better performance than some semi-supervised algorithms (e.g., Cop-Kmeans\cite{wagstaff2001constrained}, Semi-Spectral clustering (Semi-SP)\cite{rangapuram2015constrained} and semi-supervised affinity propagation (Semi-AP)\cite{yujian2008SAP}). \\
\indent The remainder of this paper is organized as follows. In the next section, we outline the related work and provide the preliminary in Section \uppercase\expandafter{\romannumeral3}, which includes pairwise constraints, RBM and Gauss visible units. The proposed pcGRBM model and its learning algorithm are introduced in Section \uppercase\expandafter{\romannumeral4}. Next, the remarkable performance of the pcGRBM model is affirmed by the task of clustering on MSRA-MM in Section \uppercase\expandafter{\romannumeral5}. Finally, Section \uppercase\expandafter{\romannumeral6} summarizes our contributions.
\section{Related Work}
 \indent The classic RBM has great ability of extracting hidden features from original data. More and more researchers proposed variant RBMs and their deep networks which were based on classic RBM. There are several common methods to develop standard RBM such as adding connections information between the visible units and the hidden units, changing the value type of visible or hidden units, expanding the relationships of the units between visible layer and hidden layer from constant to variable by fuzzy mathematics and constructing deep network based on autoencoder\cite{hinton2006reducing} by pairwise constraints.\\
 \indent To add connections information between the visible units into RBM is a kind of methods for developing standard RBM. Osindero and Hinton proposed a semi-restricted Boltzmann machine (SRBM)\cite{osinder2008semi-rbm} which has lateral connections between the visible units, but these lateral connections are unit-level semi-supervised information. The learning procedure includes two stages:  The first one is the visible to hidden connections which is same as a classic RBM and the second one is the lateral connections which are applied the same learning procedure as the first one. In order to enforce hidden units to be pairwise uncorrelated and to maximize the entropy, Tomczak\cite{Tomczak2015Learning} proposed to add a penalty term to the log-likelihood function. His framework of learning informative features is unit-level pairwise and for classification problems, while our model is instance-level pairwise and for clustering tasks. Zhang et al.\cite{zhang2016incremental} built a deep belief network based on SRBM for classification. Given the hidden units, the visible units of the SRBM form a Markov random field. However, the main weakness of the SRBM is that there are massive parameters for high-dimensional data, if every pair of visible units have relations. Sutskever and Hinton proposed a temporal restricted Boltzmann machine (TRBM)\cite{sutskever2007learning} by adding directed connections between previous and current states of the visible and hidden units. There are three kinds of connections of the full TRBM, e.g., connections between the visible units, connections between the hidden and visible units and connections between the hidden units. Furthermore, they proposed a recurrent TRBM (RTRBM)\cite{sutskever2009recurrent}. It is easy to compute the gradient of the log-likelihood and infer exactly. \\
 \indent By changing hidden units with relevancy is another kind of methods for developing standard RBM. Courville et al.\cite{courville2014spike} developed a spike-and-slab restricted Boltzmann machine (ssRBM). The ssRBM is defined as having each hidden unit associated with the product of a binary ``spike" latent variable and a real-valued ``slab" latent variable. In order to keep learning efficiency, as a model of natural images, the binary hidden units of the ssRBM maintain the simple conditional independence structure when they encode the conditional covariance of visible units by exploiting real-valued ``slab" latent variable.\\
 \indent In general, the relationships of the units between the visible layer and the hidden layer are restricted to be constants. In order to break through this restrictions, Chen et al.\cite{PChen2015fuzzyrbm} proposed a fuzzy restricted Boltzmann machine (FRBM) to enhance deep learning capability which can avoid the deficiency. The FRBM model parameters are replaced by fuzzy numbers and the regular RBM energy function is given by fuzzy free energy functions. Moreover, the deep networks are designed by the fuzzy RBMs to boost deep learning. Nie et al.\cite{nie2015generative} proposed to theoretically extend the conventional RBMs by introducing another term in the energy function to explicitly model the local spatial interactions in the input data.\\
 \indent Conventional RBM defines the units of visible and hidden layer to be binary, but this limitation cannot meet the needs in practice. Then one common way is to replace them by means of Gaussian linear units, that is Gaussian-Bernoulli restricted Boltzmann machines (GBRBMs)\cite{krizhevsky2009learning}. The GBRBMs have the ability to learn meaningful features both in modeling natural images and in a two-dimensional separation task. But, as we know, it is difficult to learn the GBRBMs. So, Cho et al.\cite{cho2011improved} proposed a novel method to improve their learning efficiency. The new method includes three parts, e.g., changing the energy function by different parameterizations to facilitate learning, parallel tempering learning and adaptive learning rate. Moreover, the deep networks of Gaussian-Bernoulli deep Boltzmann machine (GDBM)\cite{cho2013gaussian}, \cite{taylor2011two} have been developed by the GBRBM in recent years. The GDBM is designed by adding multiple layers of hidden units and applied to continuous data.\\
 \indent Furthermore, Zhang et al. proposed a mixed model named as a supervision guided autoencoder (SUGAR)\cite{zhang2014supervised} which includes three components: main network, auxiliary network and bridge. The main network is a sparsity-encouraging variant of the autoencoder\cite{hinton2006reducing}, that is the unsupervised autoencoder. The auxiliary network is constructed by pairwise constraints, that is the supervised learning. The two heterogeneous networks are designed and each of which encodes either unsupervised or supervised data structure respectively. The main network and auxiliary network are connected by the bridge which is used to enforce the correlation of the parameters. Comparing SUGAR with supervised learning and supervised deep networks, it has flexible utilization of supervised information and better balances the numerical tractability.\\
 \indent For many practical applications, the researchers have proposed various derivatives of RBM. Yu et al. proposed a classification RBM\cite{yu2014renyi} which is an effective classifier by extending the Conditional Log Likelihood objective. Han et al. proposed a circle convolutional restricted Boltzmann machine (CCRBM)\cite{han2016unsupervised} for extracting local features from three dimensional shapes, and it holds a new ring-like multi-layer structure with an unsupervised three dimensional local feature learning. As for analyzing unstructured events and group activities from uncontrolled web videos, Zhao et al. proposed a relevance restricted Boltzmann machine (ReRBM)\cite{zhao2016learning} which extends classic RBM by incorporating sparse Bayesian learning into RBM and replacing binary hidden units by linear units. Gao et al. proposed a centered convolutional restricted Boltzmann machine (CCRBM)\cite{Gao2016A} for scene recognition. As for predicting and modeling human behaviors in health social networks, Phan et al. proposed a social restricted Boltzmann machine (SRBM)\cite{phan2015social}, which incorporates environmental events, self-motivation and explicit social influences together into hidden, historical and visible layers. Li et al. proposed a temperature based restricted Boltzmann machine (TRBM)\cite{li2016temperature}, which introduces an essential temperature parameter to improve the performance.\\
 \indent In the work of \cite{chengang2015deep}, Chen proposed a deep network structure based on RBMs which is the most related to our work. Both the work of \cite{chengang2015deep} and our work aim to solve the similar problems, e.g., how to obtain suitable features for clustering by non-linear mapping and how to use pairwise constraints during the learning process, but the model and the solution are different. They use RBMs to initialize connection weights with CD learning. Its learning process is still unsupervised method, then the learned weights are used to incorporate pairwise constraints in the feature space by maximum margin techniques. However, our pcGRBM model is based on RBMs with Gaussian visible units. Its learning process is no longer unsupervised method, but guided by pairwise constraints.\\
 \indent There are many semi-supervised feature extraction methods which are not based on RBM. Fan et al. proposed a novel graph-based semi-supervised learning method which utilized an effective and simple graph construction method to establish the graph\cite{MingyuSemi}. The method has an advantage of ensuring the connectivity between pairwise data points. To recognize video semantic, Luo et al. proposed an adaptive semi-supervised feature learning method which incorporates a local structure into joint feature selection for learning the optimal graph simultaneously\cite{Luo2017An}. As for RNA-Seq data analysis, Liu et al. proposed a semi-supervised feature extraction with the joint $L_{1,2}$-norm constraint (L21SFE), which constructs a Laplacian matrix by using the labeled samples\cite{Liu2016A}. Dong et al. proposed a novel semi-supervised SVM with extended hidden feature (SSVM-EHF), which can address the negative impact issue of some inaccurate labeled samples. Zhu and Zhang proposed a semi-supervised dimensionality reduction algorithm by pairwise constraints between tensor images\cite{Zhu2009Semi}. In the mixture graph feature extraction, Yu et al. proposed a semi-supervised dimensionality reduction (MGSSDR) with pairwise constraints which can preserve the local structure and pairwise constraints of samples in the subspace\cite{Yu2010Mixture}. As for hyperspectral image classification, Chen and Zhang proposed a semi-supervised dimensionality reduction framework, which is based on pairwise constraints and sparse representation\cite{Chen2011Semisupervised}.

\section{Preliminaries}
In this section, the background of the pairwise constraints, RBM and Gaussian visible units is briefly summarized.
\subsection{Pairwise Constraints}
The priori knowledge of pairwise constraints is widely used in supervised and semi-supervised learning\cite{wagstaff2001constrained}, \cite{zhang2008constraint}, \cite{ZhangZhiHua2008Constraint}. There are two types of instance-level pairwise constraints: One is cannot-link constraints $\mathcal{CL}=\{(x_{i},x_{j})\}$ and the other is must-link constraints $\mathcal{ML}=\{(x_{i},x_{j})\}$, where $(x_{i},x_{j})\in\mathcal{CL}$ implies that $x_{i}$ and $x_{j}$ belong to different clusters, while $(x_{i},x_{j})\in\mathcal{ML}$ implies that $x_{i}$ and $x_{j}$ belong to the same cluster. The must-link and cannot-link constraints define an instance-level relation of transitive binary. Consequently, two types of constraints may be derived from background knowledge about data set or labeled data. In this paper, we select labeled data from different groups randomly and ensure each group has the same ratio of labeled data to be selected. Then, the must-link constraints are produced by the selected same group labeled data and the cannot-link constraints are produced by the selected different group labeled data.

\subsection{Restricted Boltzmann Machine}
   A RBM\cite{hinton1986learning}\cite{hinton2002training} is a two-layer network in which the first layer consists of visible units, and the second layer consists of hidden units. The symmetric undirected weights are used to connect the visible and hidden layers. There is no interior-layer connection with either the visible units or the hidden units. A classic RBM model is shown in Fig. 1. An energy function\cite{hopfield1982neural} of a joint configuration (\textbf{v}, \textbf{h}) between the visible layer and the hidden layer is given by
   \begin{equation}
   E(\texttt{\textbf{v}},\texttt{\textbf{h}})=-\sum\limits_{i\in \texttt{visible}}a_{i}v_{i}-\sum\limits_{j\in hidden}b_{j}h_{j}-\sum\limits_{i,j}v_{i}h_{j}w_{ij},
   \end{equation}
  where $\texttt{\textbf{v}}=(v_{1}, v_{2}\cdots v_{n})$ and $\textbf{\texttt{h}}=(h_{1}, h_{2}\cdots h_{m})$ are the visible and hidden vectors, $a_{i}$ and $b_{j}$ are their biases, $n$ and $m$ are the dimension of visible layer and hidden layer, respectively, $w_{ij}$ is the connection weight matrix between the visible layer and hidden layer. A probability distribution over vectors $\textbf{\texttt{v}}$ and $\texttt{\textbf{h}}$ is defined as
  \begin{equation}
  p(\textbf{\texttt{v}},\texttt{\textbf{h}})=\frac{1}{Z}e^{-E(\texttt{\textbf{v}},\texttt{\textbf{h}})},
  \end{equation}
  where $Z$ is a ``partition function" which is defined by summing over all possible pairs of hidden layer and visible layer:
\begin{equation}
Z=\sum\limits_{\texttt{\textbf{v}},\texttt{\textbf{h}}}e^{-E(\texttt{\textbf{v}},\texttt{\textbf{h}})}.
\end{equation}
\begin{figure}
  \centering
  \includegraphics[scale=0.345]{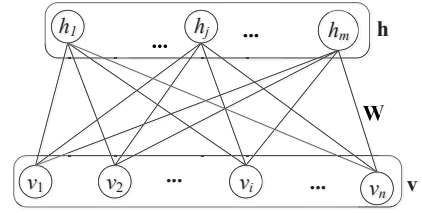}\\
  \caption{Restricted Boltzmann Machine (RBM)}
\label{rbm}
\end{figure}
\indent By means of summing over all the units of the hidden layer, the probability that the RBM assigns to the units of the visible layer $\texttt{\textbf{v}}$ is given by:
\begin{equation}
p(\texttt{\textbf{v}})=\frac{1}{Z}\sum\limits_{\texttt{\textbf{h}}}e^{-E(\texttt{\textbf{v}},\texttt{\textbf{h}})}.
\end{equation}
\indent Given a randomly selected visible layer data $\texttt{\textbf{v}}$, the binary feature of each hidden layer $h_{j}$ is equal to 1 with probability
 \begin{equation}
p(h_{j}=1|\texttt{\textbf{v}})=\sigma(b_{j}+\sum\limits_{i}v_{i}w_{ij}),
\end{equation}
where $\sigma$ is the sigmoid function.\\
\indent The partial derivative of the log probability of Eq. (4) with respect to a weight is given by
\begin{equation}
  \frac{\partial \log p(\texttt{\textbf{v}})}{\partial w_{ij}}=(<v_{i}h_{j}>_{data}-<v_{i}h_{j}>_{model}),
\end{equation}
where the angle brackets of $<v_{i}h_{j}>_{data}$ and $<v_{i}h_{j}>_{model}$ are used to denote expectations of the distribution specified by the subscript $data$ and $model$, respectively. In the log probability, a very simple learning rule for performing stochastic steepest ascent is given by:
\begin{equation}
 \Delta w_{ij} =\varepsilon(<v_{i}h_{j}>_{data}-<v_{i}h_{j}>_{model}),
\end{equation}
where $\varepsilon$ is a learning rate.\\
\indent It is easy to obtain $<v_{i}h_{j}>_{data}$ because there is no direct connection among the hidden units. However, it is difficult to get an unbiased sample of $<v_{i}h_{j}>_{model}$. Hinton\cite{hinton2002training} proposed a faster learning algorithm with the CD learning and the change of learning parameters are given by:
\begin{equation}
 \Delta w_{ij} =\varepsilon(<v_{i}h_{j}>_{data}-<v_{i}h_{j}>_{recon}),
\end{equation}
\begin{equation}
 \Delta a_{i}=\varepsilon(<v_{i}>_{data}-<v_{i}>_{recon}),
\end{equation}
\begin{equation}
 \Delta b_{j}=\varepsilon(<h_{j}>_{data}-<h_{j}>_{recon}),
\end{equation}
where $<v_{i}h_{j}>_{recon}$ can be computed efficiently than $<v_{i}h_{j}>_{model}$.
\subsection{Gaussian Visible Units}
Original RBMs were developed by binary stochastic units for the hidden and visible layers\cite{hinton2002training}. To deal with real-valued data such as natural images, one solution is that the binary visible units are replaced by linear units with independent Gaussian noise, but the hidden units remain binary, which is first suggested by Freund and Haussler in \cite{freund1994unsupervised}. The negative log probability is given by the following energy function:
\begin{equation}
\begin{aligned}
  &-\mathbf{log}p(\textbf{\texttt{v}},\textbf{\texttt{h}})=E(\texttt{\textbf{v}},\textbf{\texttt{h}})=\\
  &\sum\limits_{i\in \texttt{visible}}\frac{(v_{i}-a_{i})^2}{2\sigma_{i}^2}-\sum\limits_{j\in hidden}b_{j}h_{j}-\sum\limits_{i,j}\frac{v_{i}}{\sigma_{i}}h_{j}w_{ij},
\end{aligned}
 \end{equation}
where $\sigma_{i}$ is the standard deviation of the Gaussian noise for visible unit $i$.\\
\indent The conditional probability of visible layer is
\begin{equation}
\begin{aligned}
  p(\textbf{\texttt{v}}|\textbf{\texttt{h}})=\mathcal{N}(\sum\textbf{\texttt{h}}\mathbf{W}^T+\mathbf{a},\sigma^{2}),
\end{aligned}
 \end{equation}
 where $\mathcal{N}(\cdot)$ is a gaussian density with a mean $\sum\textbf{\texttt{h}}\mathbf{W}^T+\mathbf{a}$ and a variances $\sigma^{2}$. \\
\indent Given two divergences, $\texttt{\textbf{CD}}_{1}$\cite{carreira2005contrastive} learning is defined by:
\begin{equation}
\begin{aligned}
  \texttt{\textbf{CD}}_{1}=\texttt{\textbf{KL}}(p_{0}||p_{\infty})-\texttt{\textbf{KL}}(p_{1}||p_{\infty}),
\end{aligned}
 \end{equation}
where $\texttt{\textbf{KL}}$ is Kullback-Leibler divergence\cite{carreira2005contrastive}, $p_{o}$ is the data distribution and $p_{1}$ is the distribution of the data after running the Markov chain for one step.\\
\indent For each visible unit, it is easy to learn the variance of the noise, but it is difficult to use $\texttt{\textbf{CD}}_{1}$ because of taking long time\cite{krizhevsky2009learning}\cite{krizhevsky2010convolutional}. Therefore, in many applications, it is easy to normalise the data to have unit variance and zero mean\cite{krizhevsky2009learning}\cite{salakhutdinov2009learning}\cite{hinton2010practical}\cite{nair2010rectified}. Then the reconstructed value of Gaussian visible units is equal to its input from the binary hidden units plus its bias.
\section{pcGRBM Model and Its Learning Algorithm}
We first propose a pairwise constraints restricted Boltzmann machine with Gaussian visible units (pcGRBM) model which the binary visible units are replaced by noise-free linear units and its learning procedure is guided by pairwise constraints. Then we give exact inference of the pcGRBM optimization. Finally, the corresponding learning algorithm is presented.\\
\subsection{pcGRBM Model}
\indent Suppose that $\mathcal{V}=\{\mathbf{\textbf{v}_{1}},\mathbf{\textbf{v}_{2}},\cdots,\mathbf{\textbf{v}_{n}}\}$ is a $p$-dimensional original data set which has been normalized, $\mathcal{H}=\{\mathbf{\textbf{h}_{1}},\mathbf{\textbf{h}_{2}},\cdots,\mathbf{\textbf{h}_{n}}\}$ is a $q-$dimensional hidden code. The pairwise must-link constraints set of the reconstruction data is defined by $\mathcal{M}=\{(\mathbf{\textbf{v}_{s}},\mathbf{\textbf{v}_{t}})| \mathbf{\textbf{v}_{s}}$, $\mathbf{\textbf{v}_{t}}$ belong to the same class$\}$ and a pairwise cannot-link constraints set of the reconstruction data is given by $\mathcal{C}=\{(\mathbf{\textbf{v}_{s}},\mathbf{\textbf{v}_{t}})| \mathbf{\textbf{v}_{s}}$, $\mathbf{\textbf{v}_{t}}$ belong to the different classes$\}$.\\
\indent For training the parameters of the pcGRBM model, the first objective is that how to maximize the log probability of RBM with Gaussian visible units and the second objective is that how to maximize the distance of all pairwise vectors which come from the cannot-link set and minimize the distance of all pairwise vectors which come from the must-link set in the reconstructed visible layer. Because of using noise-free reconstruction in the model, the reconstructed value of a Gaussian visible linear unit is equal to its input from the hidden units plus its bias. The objective function is given by
\begin{equation}
\begin{aligned}
   &L(\theta,\mathcal{V})=\\
   &-\frac{\lambda}{n}\sum\limits_{v_{i}\in\mathcal{V} }\mathbf{log}p(\mathbf{\textbf{v}_{i}},\theta)+\Big[\Big(\frac{1-\lambda}{N_{M}}\sum\limits_{\mathcal{M}}\Arrowvert \mathbf{\textbf{v}}_{s}^{(1)}-\mathbf{\textbf{v}_{t}}^{(1)}\Arrowvert^2\\
   &-\frac{1-\lambda}{N_{C}}\sum\limits_{\mathcal{C}}\Arrowvert \mathbf{\textbf{v}}_{s}^{(1)}-\mathbf{\textbf{v}_{t}}^{(1)}\Arrowvert^2\Big)\Big],
\end{aligned}
\end{equation}
where $\mathbf{\textbf{v}}_{s}^{(1)}$ and $\mathbf{\textbf{v}_{t}}^{(1)}$ are the reconstructed values of visible Gaussian linear layer, respectively. So, $\mathbf{\textbf{v}}_{s}^{(1)}=\mathbf{\textbf{h}_{s}}\mathbf{W}^T+\mathbf{a}$ and $\mathbf{\textbf{v}_{t}}^{(1)}=\mathbf{\textbf{h}_{t}}\mathbf{W}^T+\mathbf{a}$. Then the objective function has another form:
\begin{equation}
\begin{aligned}
   &L(\theta,\mathcal{V})=\\
   &-\frac{\lambda}{n}\sum\limits_{v_{i}\in\mathcal{V} }\mathbf{log}p(\mathbf{\textbf{v}_{i}},\theta)+\Big[\Big(\frac{1-\lambda}{N_{M}}\sum\limits_{\mathcal{M}}\Arrowvert \mathbf{\textbf{h}_{s}}\mathbf{W}^T-\mathbf{\textbf{h}_{t}}\mathbf{W}^T\Arrowvert^2\\
   &-\frac{1-\lambda}{N_{C}}\sum\limits_{\mathcal{C}}\Arrowvert \mathbf{\textbf{h}_{s}}\mathbf{W}^T-\mathbf{\textbf{h}_{t}}\mathbf{W}^T\Arrowvert^2\Big)\Big],
\end{aligned}
\end{equation}
where $\theta=\{\mathbf{a},\mathbf{b},\mathbf{W}\}$ are the model parameters, $\mathbf{a}$ is the visible biases matrix and $\mathbf{b}$ is the hidden biases matrix. $\lambda\in(0,1)$ is a scale coefficient, $N_{M}$ and $N_{C}$ are the cardinalities of the must-link pairwise constraints set $\mathcal{M}$ and the cannot-link pairwise constraints set $\mathcal{C}$, respectively,  $\frac{1}{n}\sum\limits_{i=1}^{n}\mathbf{log}p(\mathbf{\textbf{v}_{i}};\theta)$ is the average of the log-likelihood and $\Arrowvert\cdot\Arrowvert^2$ is the square of 2-norm. \\
\indent The learning problem of the pcGRBM model is to get optimal or approximate optimal parameters $\theta$, which minimize the objective function $L(\theta,\mathcal{V})$, i.e.,
\begin{equation}
\begin{aligned}
   \textbf{\texttt{min}}\{L(\theta,\mathcal{V})\}.
\end{aligned}
\end{equation}
\begin{figure}
  \centering
  \includegraphics[scale=0.335]{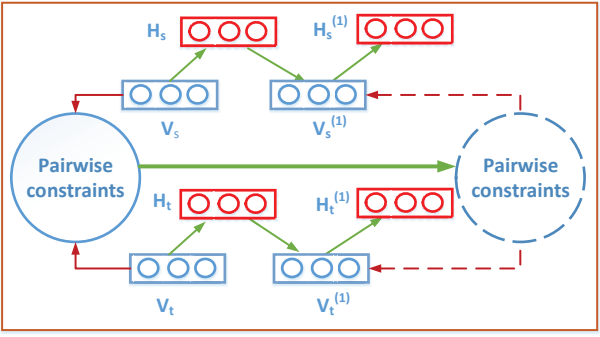}\\
  \caption{The pcGRBM model by $\textbf{\texttt{CD}}_{1}$ learning, where the pairwise constraints of the reconstructed data between $\mathbf{\textbf{v}}_{s}^{(1)}$ and $\mathbf{\textbf{v}_{t}}^{(1)}$ stem from the relationships of the original data $\mathbf{\textbf{v}}_{s}$ and $\mathbf{\textbf{v}_{t}}$.   }
\label{rbm}
\end{figure}

\subsection{pcGRBM Inference}
For our first objective, we can use the gradient descent to solve optimal problems, however, it is expensive to compute the gradient of the log probability. Recently, Karakida et al.\cite{Karakida2016Dynamical} demonstrated that $\textbf{\texttt{CD}}_{1}$ learning is simpler than ML learning in RBMs with Gaussian linear units. Then, we apply the $\textbf{\texttt{CD}}_{1}$ learning method to obtain an approximation of the log probability gradient. For our second objective, we use the method of gradient descent to solve the optimization problem. Next, the main work is that how to compute the gradient of the following equation:
\begin{equation}
\begin{aligned}
\frac{1-\lambda}{N_{M}}\sum\limits_{\mathcal{M}}\Arrowvert (\mathbf{\textbf{h}_{s}}-\mathbf{\textbf{h}_{t}})\mathbf{W}^T\Arrowvert^2-\frac{1-\lambda}{N_{C}}\sum\limits_{\mathcal{C}}\Arrowvert (\mathbf{\textbf{h}_{s}}-\mathbf{\textbf{h}_{t}})\mathbf{W}^T\Arrowvert^2.
\end{aligned}
\end{equation}

\indent Firstly, we assume that
\begin{equation}
\begin{aligned}
    J_{M}(\mathbf{W})=\frac{1}{N_{M}}\sum\limits_{\mathcal{M}}\big\Arrowvert \mathbf{\textbf{h}_{s}}\mathbf{W}^T-\mathbf{\textbf{h}_{t}}\mathbf{W}^T\big\Arrowvert^2
\end{aligned}
\end{equation}
and
\begin{equation}
\begin{aligned}
    J_{C}(\mathbf{W})=\frac{1}{N_{C}}\sum\limits_{\mathcal{C}}\Arrowvert \mathbf{\textbf{h}_{s}}\mathbf{W}^T-\mathbf{\textbf{h}_{t}}\mathbf{W}^T\Arrowvert^2.
\end{aligned}
\end{equation}
\indent Then, the gradient of the $J_{M}(\mathbf{W})$ is
\begin{equation}
\begin{aligned}
   \frac {\partial J_{M}(\mathbf{W})}{\partial w_{ij}}=&\frac{1}{N_{M}}\sum\limits_{\mathcal{M}}\Bigg[ (\mathbf{\textbf{h}_{s}}-\mathbf{\textbf{h}_{t}})\mathbf{W}^T\frac{\partial{\mathbf{W}(\mathbf{\textbf{h}_{s}}-\mathbf{\textbf{h}_{t}})^T}}{{\partial w_{ij}}}+\\      &\frac{\partial{(\mathbf{\textbf{h}_{s}}-\mathbf{\textbf{h}_{t}})\mathbf{W}^T}}{{\partial w_{ij}}}\mathbf{W}(\mathbf{\textbf{h}_{s}}-\mathbf{\textbf{h}_{t}})^T\Bigg]\\
\end{aligned}
\end{equation}
and the gradient of the $J_{C}(\mathbf{W})$ is
\begin{equation}
\begin{aligned}
   \frac {\partial J_{C}(\mathbf{W})}{\partial w_{ij}}=&\frac{1}{N_{C}}\sum\limits_{\mathcal{C}}\Bigg[ (\mathbf{\textbf{h}_{s}}-\mathbf{\textbf{h}_{t}})\mathbf{W}^T\frac{\partial{\mathbf{W}(\mathbf{\textbf{h}_{s}}-\mathbf{\textbf{h}_{t}})^T}}{{\partial w_{ij}}}+\\      &\frac{\partial{(\mathbf{\textbf{h}_{s}}-\mathbf{\textbf{h}_{t}})\mathbf{W}^T}}{{\partial w_{ij}}}\mathbf{W}(\mathbf{\textbf{h}_{s}}-\mathbf{\textbf{h}_{t}})^T\Bigg].\\
\end{aligned}
\end{equation}
\indent In order to express concisely, we suppose that
\begin{equation}
\begin{aligned}
  \mathbf{\textbf{h}^{'}}=(h_{s1}-h_{t1},\cdots,h_{sj}-h_{tj},\cdots,h_{sq}-h_{tq}),
\end{aligned}
\end{equation}
where $h_{sk}-h_{tk}=0, k\neq j, j=1,2,\cdots,q$, and $q$ is the dimension of the hidden layer.\\
\indent Then, the gradient of the $J_{M}(\mathbf{W})$ takes the form
\begin{equation}
\begin{aligned}
 &\frac {\partial J_{M}(\mathbf{W})}{\partial w_{ij}}=\frac{1}{N_{M}}\sum\limits_{\mathcal{M}}\Big[(\mathbf{\textbf{h}_{s}}-\mathbf{\textbf{h}_{t}})\mathbf{W}^T(\mathbf{\textbf{h}^{'}})^T+\mathbf{\textbf{h}^{'}}\mathbf{W}(\mathbf{\textbf{h}_{s}}-\mathbf{\textbf{h}_{t}})^T\Big].
\end{aligned}
\end{equation}
Similarly, the gradient of the $J_{C}(\mathbf{W})$ takes the form
\begin{equation}
\begin{aligned}
 &\frac {\partial J_{C}(\mathbf{W})}{\partial w_{ij}}=\frac{1}{N_{C}}\sum\limits_{\mathcal{C}}\Big[(\mathbf{\textbf{h}_{s}}-\mathbf{\textbf{h}_{t}})\mathbf{W}^T(\mathbf{\textbf{h}^{'}})^T+\mathbf{\textbf{h}^{'}}\mathbf{W}(\mathbf{\textbf{h}_{s}}-\mathbf{\textbf{h}_{t}})^T\Big].
\end{aligned}
\end{equation}
\indent So, the gradient of the objective function is as follows.
\begin{equation}
\begin{aligned}
   \bigtriangledown w_{ij}=&\lambda\varepsilon(<v_{i}h_{j}>_{data}-<v_{i}h_{j}>_{recon})\\
   &+(1-\lambda)(F_{ij}^{(M)}-F_{ij}^{(C)}),
\end{aligned}
\end{equation}
where $F_{ij}^{(M)}=\frac{1}{N_{M}}\sum\limits_{\mathcal{M}}\Big[(\mathbf{\textbf{h}_{s}}-\mathbf{\textbf{h}_{t}})\mathbf{W}^T(\mathbf{\textbf{h}^{'}})^T+\mathbf{\textbf{h}^{'}}\mathbf{W}(\mathbf{\textbf{h}_{s}}-\mathbf{\textbf{h}_{t}})^T\Big] $ and $F_{ij}^{(C)}=\frac{1}{N_{C}}\sum\limits_{\mathcal{C}}\Big[(\mathbf{\textbf{h}_{s}}-\mathbf{\textbf{h}_{t}})\mathbf{W}^T(\mathbf{\textbf{h}^{'}})^T+\mathbf{\textbf{h}^{'}}\mathbf{W}(\mathbf{\textbf{h}_{s}}-\mathbf{\textbf{h}_{t}})^T\Big] $\\
\indent It is obvious that $\frac{\partial J_{M}(\mathbf{W})}{\partial a_{i}}=0$, $\frac{\partial J_{C}(\mathbf{W})}{\partial a_{i}}=0$, $\frac{\partial J_{M}(\mathbf{W})}{\partial b_{j}}=0$ and $\frac{\partial J_{C}(\mathbf{W})}{\partial b_{j}}=0$. So, in the pcGRBM model, we use Eq. (9) and Eq. (10) to update the biases $a_{i}$ and $b_{j}$.\\
\indent Finally, the updating rulers of connection weights $\mathbf{W}$ of the pcGRBM model takes the form
\begin{equation}
\begin{aligned}
   w_{ij}^{(\tau+1)}=&w_{ij}^{(\tau)}+\lambda\varepsilon(<v_{i}h_{j}>_{data}-<v_{i}h_{j}>_{recon})\\
   &+(1-\lambda)(F_{ij}^{(M)}-F_{ij}^{(C)}).
   \end{aligned}
\end{equation}
\subsection{pcGRBM Learning Algorithm}
According to the above inference, the learning algorithm for pcGRBM is summarized as follows.\\
\textbf{Algorithm 1 Learning for pcGRBM}\\
\noindent\line(1,0){250}\\
\textbf{Input}: $\varepsilon$ is the learning rate;\\
\indent \indent \indent $\mathcal{V}$ is a $p$-dimensional data set;\\
\indent \indent\indent $\lambda$ is a scale coefficient;\\
\indent \indent\indent $N_{M}$ is the cardinality of the must-link pairwise\\
\indent \indent\indent constraints set;\\
\indent \indent \indent $N_{C}$ is the cardinality of the cannot-link pairwise \\
\indent \indent \indent constraints set;\\
\indent \indent \indent $\mathcal{M}$ is the must-link pairwise constraints set;\\
\indent \indent \indent $\mathcal{C}$ is the must-link pairwise constraints set.\\
\noindent\line(1,0){250}\\
\textbf{Output}: $\theta=\{\mathbf{a},\mathbf{b},\mathbf{W}\}$, $\mathbf{W}$ is connection weights matrix, $\mathbf{a}$ \\
\indent \indent \indent \indent is visible biases matrix, $\mathbf{b}$ is hidden biases matrix.\\
\noindent\line(1,0){250}\\
\indent Initializing $\varepsilon$, $\lambda$, $N_{M}$, $N_{C}$, $\mathbf{W}$, $\mathbf{a}$, $\mathbf{b}$;\\
\indent \textbf{For} each iteration \textbf{do}\\
\indent \indent \textbf{For} all hidden units j \textbf{do}\\
\indent \indent \indent compute $p(h_{j}=1|\mathbf{v})=\sigma(b_{j}+\sum\limits_{i}v_{i}w_{ij})$, where $\sigma$ is\\
\indent \indent \indent a sigmoid function;\\
\indent \indent \indent sample $h_{j}\in\{0,1\}$ from $p(h_{j}=1|\mathbf{v})$ ;\\
\indent \indent \textbf{End For} \\
\indent \indent \textbf{For} all visible units i \textbf{do}\\
\indent \indent \indent compute reconstructed value $v_{i}=a_{i}+\sum\limits_{j}h_{j}w_{ij}$;\\
\indent \indent \textbf{End For} \\
\indent \indent compute the gradient of the $\frac {\partial J_{M}(\mathbf{W})}{\partial w_{ij}}$ by using Eq. (23);\\
\indent \indent compute the gradient of the $\frac {\partial J_{C}(\mathbf{W})}{\partial w_{ij}}$ by using Eq. (24);\\
\indent \indent update connection weights matrix $\mathbf{W}$ by using Eq. (26);\\
\indent \indent update visible biases matrix $\mathbf{a}$ by using Eq. (9);\\
\indent \indent update hidden biases matrix $\mathbf{b}$ by using Eq. (10);\\
\indent \textbf{End For}\\
\indent  \textbf{return} $\mathbf{W}$, $\mathbf{a}$, $\mathbf{b}$.\\
\noindent\line(1,0){250}\\


\section{Results and Discussion}
In this section, we introduce the datasets, define the experimental setup, and discuss experimental results.
\begin{table}
\begin{center}
\caption{Summary of the data sets.}
 \label{imagedata}
\scalebox{1}{
\begin{tabular}{lcccc}
\toprule[1.5pt] 
\textsf{\bf{No.}} & \textsf{\bf{Dataset}} &  {classes}&  {Instances} &  {features} \\
\hline
\textsf{1} &{alph} & {3} & {814}  & {892}\\
\textsf{2} &{alphabet} & {3} & {814}  & {892} \\
\textsf{3} &{aquarium} & {3} & {922} &{892} \\
\textsf{4} &{bed} & {3} & {888}  & {892} \\
\textsf{5} &{beer} & {3} & {870} &{892} \\
\textsf{6} &{beverage} & {3} & {873} &{892} \\
\textsf{7} &{breakfast} & {3} & {895} &{892} \\
\textsf{8} &{virus} & {3} & {871} &{899} \\
\textsf{9} &{webcam}& {3} & {790} & {899} \\
\textsf{10} &{weddi}& {3} & {883} &{899} \\
\textsf{11} &{wii}& {3} & {726} &{899}\\
\textsf{12} &{wing}& {3} & {856} &{899}\\
\bottomrule[1pt] 
\end{tabular}}
\end{center}
\end{table}
\subsection{DataSets}
In the following experiments, we use the Microsoft Research Asia Multimedia (MSRA-MM)\cite{li2009msra} dataset which contains two sub-datasets, e.g., a video dataset and an image dataset. The image part contains 1,011,738 images and the video part contains 23,517 videos. To evaluate our pcGRBM model, we use two kinds of image datasets with different features from image part of the MSRA-MM for our experiments. The first kind of datasets have 892 features and the second kind of datasets have 899 features. The summary of all datasets is listed in Table I.
\subsection{Experimental Setup}
\begin{figure}
  \centering
  \includegraphics[scale=0.505]{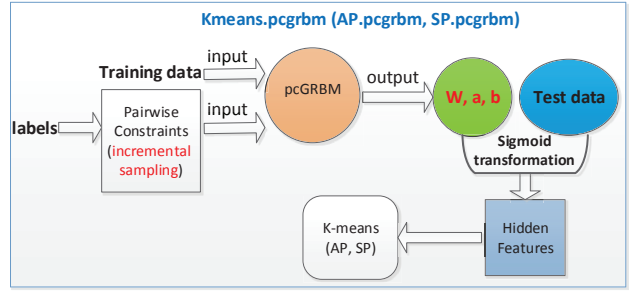}\\
  \caption{Three structures of Kmeans.pcgrbm, AP.pcgrbm and SP.pcgrbm clustering algorithms based on pcGRBM model where pairwise constraints are derived from the incremental sampling procedure. The pcGRBM mode is trained and tested by 10-fold cross-validation strategy.  Nine subsamples of each dataset are the training data, and the remaining one subsample is the test data. The hidden features for testing are from the sigmoid transformation of test data with the parameters of pcGRBM model ($\mathbf{W},\mathbf{a},\mathbf{b}$). }
\label{pcgrbm}
\end{figure}
In our experiments, we use the incremental sampling method to obtain pairwise constraints from 1\% to 8\% for all semi-supervised algorithms and our pcGRBM model. The 10-fold cross-validation strategy is used to test the performance of clustering algorithms based on the pcGRBM model. To make the comparisons of algorithms more reasonable, we use the test data of cross-validation as the input of all unsupervised clustering algorithms. The scale coefficient of pcGRBM model is set to 0.7 ($\lambda$=0.7) and the learning rates of traditional RBM and our pcGRBM model are set to $10^{-8}$ ($\varepsilon=10^{-8}$). The goal of the experiments is to study the following aspects:
\begin{itemize}
  \item Do the pairwise constraints guide the encoding procedure of traditional RBM?
  \item How do unsupervised clustering algorithms based on pcGRBM model compare with these algorithms based on traditional RBM?
  \item How do unsupervised clustering algorithms based on pcGRBM model compare with their semi-supervised clustering algorithms?
  \item How do semi-supervised clustering algorithms based on pcGRBM model compare with their classic semi-supervised clustering algorithms or unsupervised clustering algorithms based on pcGRBM model?
\end{itemize}
 To verify the features of pcGRBM contain guiding information whether or not, we use the output of pcGRBM as the input of unsupervised clustering algorithm. The twelve different datsets are used to train the pcGRBM model independently and each dataset mappings a feature set, then we obtain 12 different features for clustering task. In our experiments, we choose K-means, AP, SP clustering algorithms as examples. Then, we present three algorithms which based on pcGRBM model for clustering tasks, termed as Kmeans.pcgrbm, AP.pcgrbm and SP.pcgrbm. Their structures are shown in Fig. 3. The different datasets as the input of pcGRBM, then we obtain different semi-supervised features. All structures of clustering algorithm in Fig. 3 are similar. The only difference is the clustering method. Kmeans.pcgrbm, SP.pcgrbm and AP.pcgrbm all use the same features of our pcGRBM model as the input by K-means, SP and AP clustering algorithm, respectively. Similarly, we also present three algorithms which are based on traditional RBM with Gaussian visible units for clustering tasks, called as Kmeans.grbm, AP.grbm and SP.grbm. In fact, Kmeans.pcgrbm, AP.pcgrbm and SP.pcgrbm are semi-supervised clustering algorithms with instance-level guiding of pairwise constraints, but Kmeans.grbm, AP.grbm and SP.grbm are unsupervised methods. It is natural to use the features of our pcGRBM model as the input of semi-supervised clustering algorithms (Cop-Kmeans, Semi-SP and Semi-AP), termed as Cop-Kmeans.pcgrbm, Semi-AP.pcgrbm and Semi-SP.pcgrbm, respectively. \\
 \indent Firstly, we compare the clustering performance of the proposed algorithms (Kmeans.pcgrbm, AP.pcgrbm and SP.pcgrbm) with original K-means, AP and SP clustering algorithms, respectively. Secondly, the proposed algorithms are used to compare with Cop-Kmeans\cite{wagstaff2001constrained}, Semi-SP\cite{rangapuram2015constrained} and Semi-AP\cite{yujian2008SAP}, respectively. Thirdly, we use unsupervised algorithms that are Kmeans.grbm, AP.grbm and SP.grbm to compare with the proposed algorithms. Finally, we compare the clustering performance of the proposed algorithms with Cop-Kmeans.pcgrbm, Semi-SP.pcgrbm and Semi-AP.pcgrbm, respectively.\\
 \indent In order to obtain more meaningful results, we use 10-fold cross-validation strategy to partition each dataset into 10 subsamples randomly. Nine subsamples of each dataset are used to train our pcGRBM model, and the remaining one subsample is the test data. The cross-validation process is repeated 10 times. Each of the 10 subsamples is used exactly once as the validation data. The hidden features for testing come from the sigmoid transformation of the test data with the parameters of pcGRBM model ($\mathbf{W},\mathbf{a},\mathbf{b}$).\\
\indent To evaluate the performance of the clustering algorithms, we adopt three widely used metrics: clustering accuracy\cite{cai2005document}, clustering purity\cite{ding2006orthogonal} and Friedman Aligned Ranks test\cite{garcia2010advanced} as the evaluation measures.\\
\indent Give an instance $x_{i}$, let $s_{i}$ and $r_{i}$ be the true label and the obtained cluster label, respectively. The clustering accuracy is defined by:
\begin{equation}
\begin{aligned}
    accuracy=\frac{\sum\limits_{i=1}\delta(s_{i},map(r_{i}))}{n},
\end{aligned}
\end{equation}
where $n$ is the total number of instances, $\delta(x,y)$ equals to one if $x=y$ and zero otherwise, and $map(r_{i})$ maps each cluster label $r_{i}$ to the equivalent label from the data set. \\
\indent Purity is a transparent external evaluation measure for cluster quality and it measures the extent of each cluster contained data points from primarily one class. The purity of a clustering is given by
 \begin{equation}
  purity=\sum_{i=1}^K\frac{n_i}{n}P(s_i), P(s_i)=\frac{1}{n_i}\max_j(n_i^j),
  \end{equation}
  where $s_i$ is a particular cluster size of $n_i$ and $n_i^j$  is the number of the $i$-th input class assigned to the $j$-th cluster.\\
\indent The Friedman Aligned Ranks test\cite{garcia2010advanced} is based on $n$ data sets and $m$ algorithms of ranks. It is given by
 \begin{equation}
\begin{aligned}
  T=\frac{(G-1)\Big[\sum\limits_{j=1}^{G}\widehat{R}_{.j}^2-(GD^2/4)(GD+1)^2\Big]}{\{[GD(GD+1)(2GD+1)]/6]\}-(1/G)\sum\limits_{i=1}^{D}\widehat{R}_{i.}^2},
\end{aligned}
\end{equation}
 where $\widehat{R}_{i.}$ is the rank sum of the $j$th algorithm, $\widehat{R}_{.j}$ is the rank sum of the $i$th data set, $D$ is the number of data set and $G$ is the number of algorithm.\\
\subsection{Results}
\begin{figure*}
\vspace{1mm} \centering
    \includegraphics[scale=0.4225]{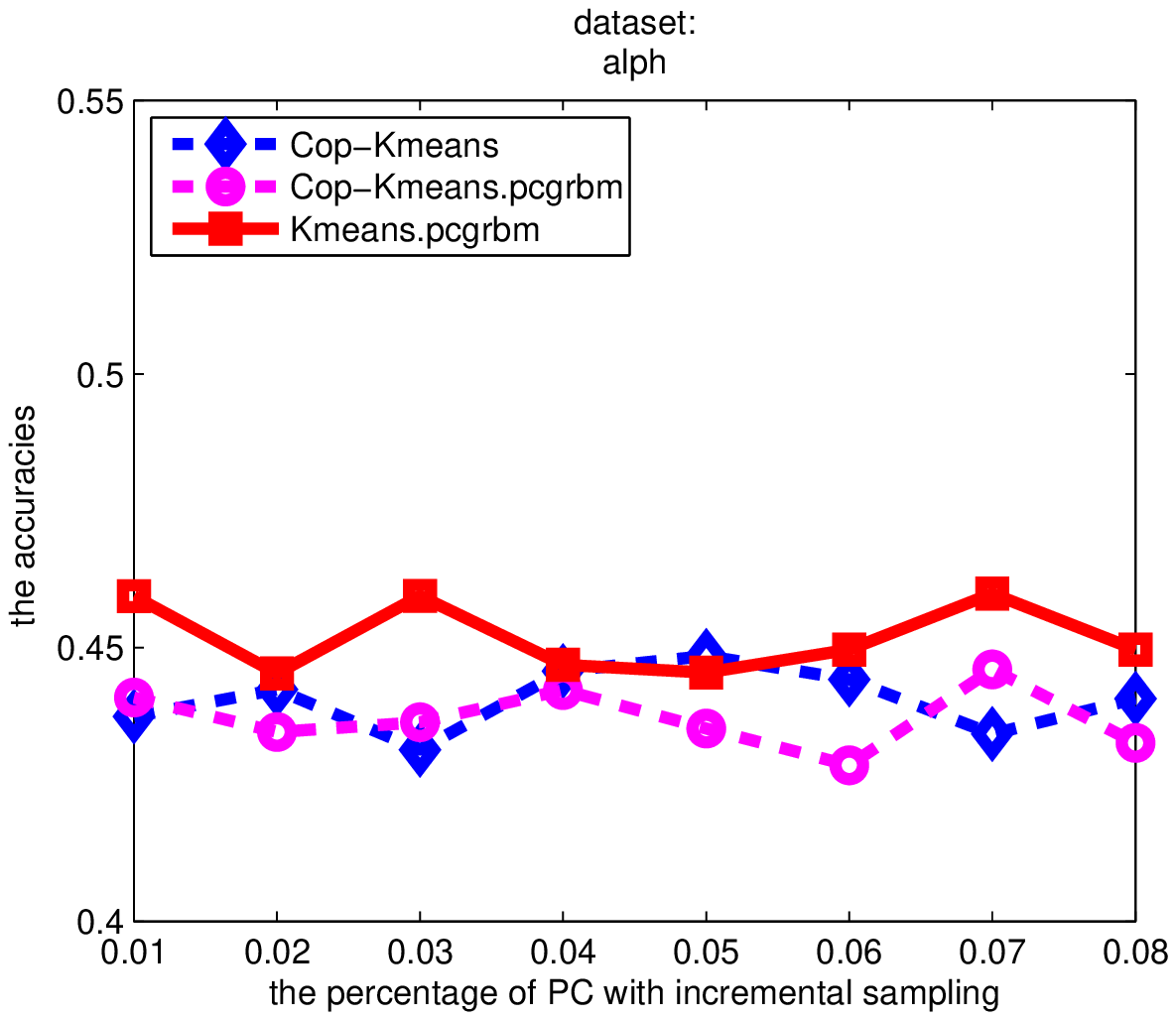}
    \includegraphics[scale=0.4225]{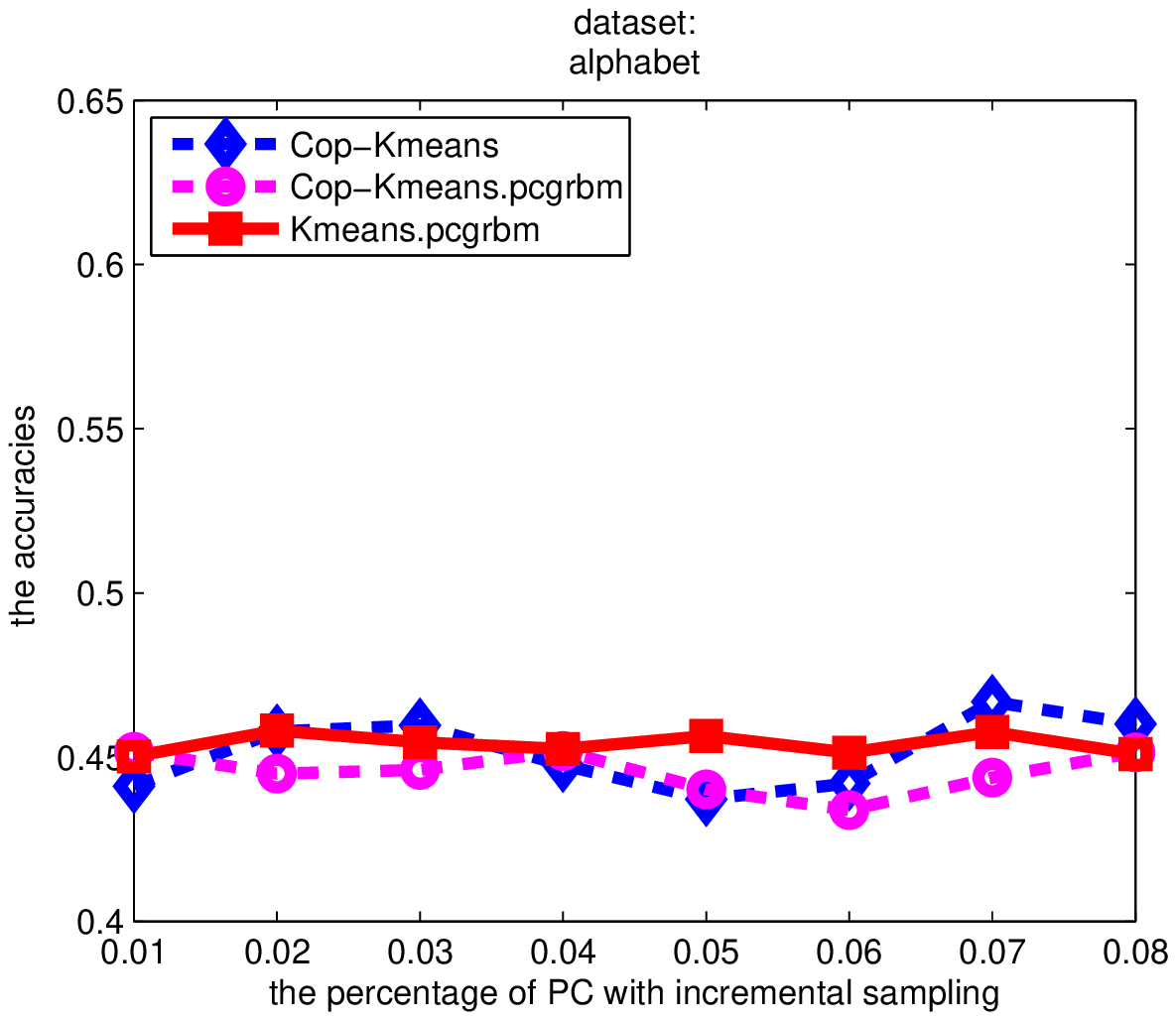}
    \includegraphics[scale=0.4225]{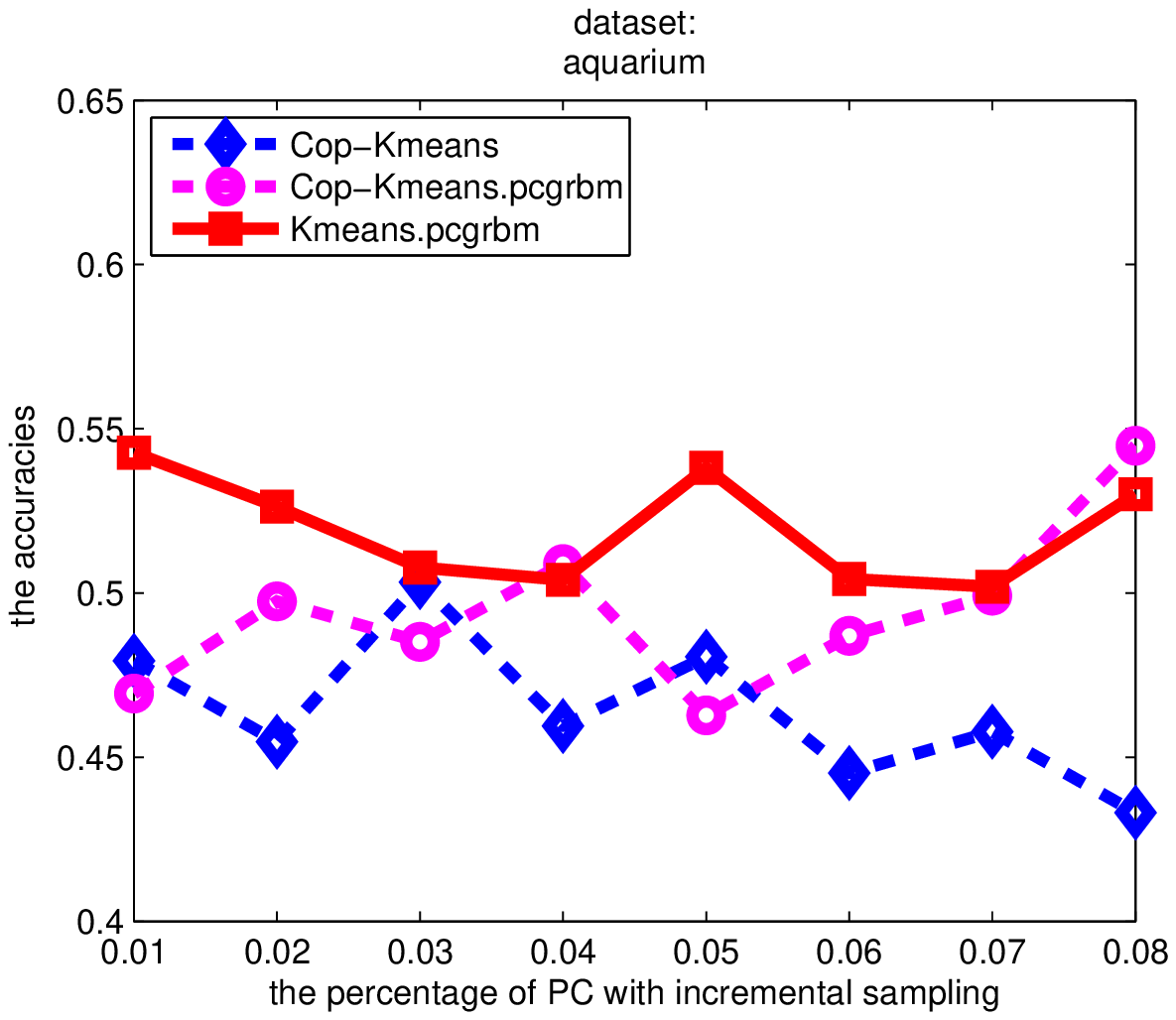}
    \includegraphics[scale=0.4225]{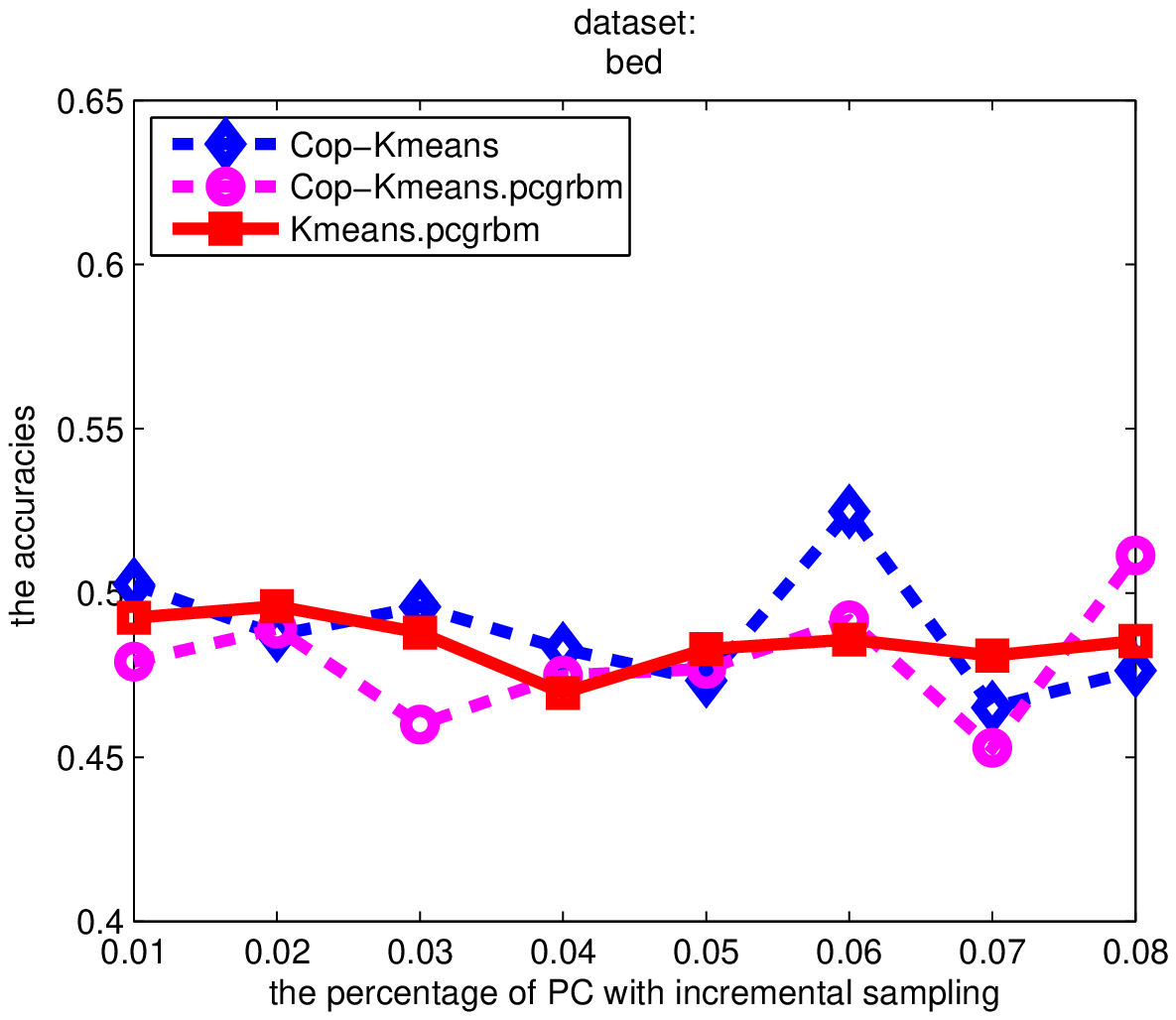}
    \includegraphics[scale=0.4225]{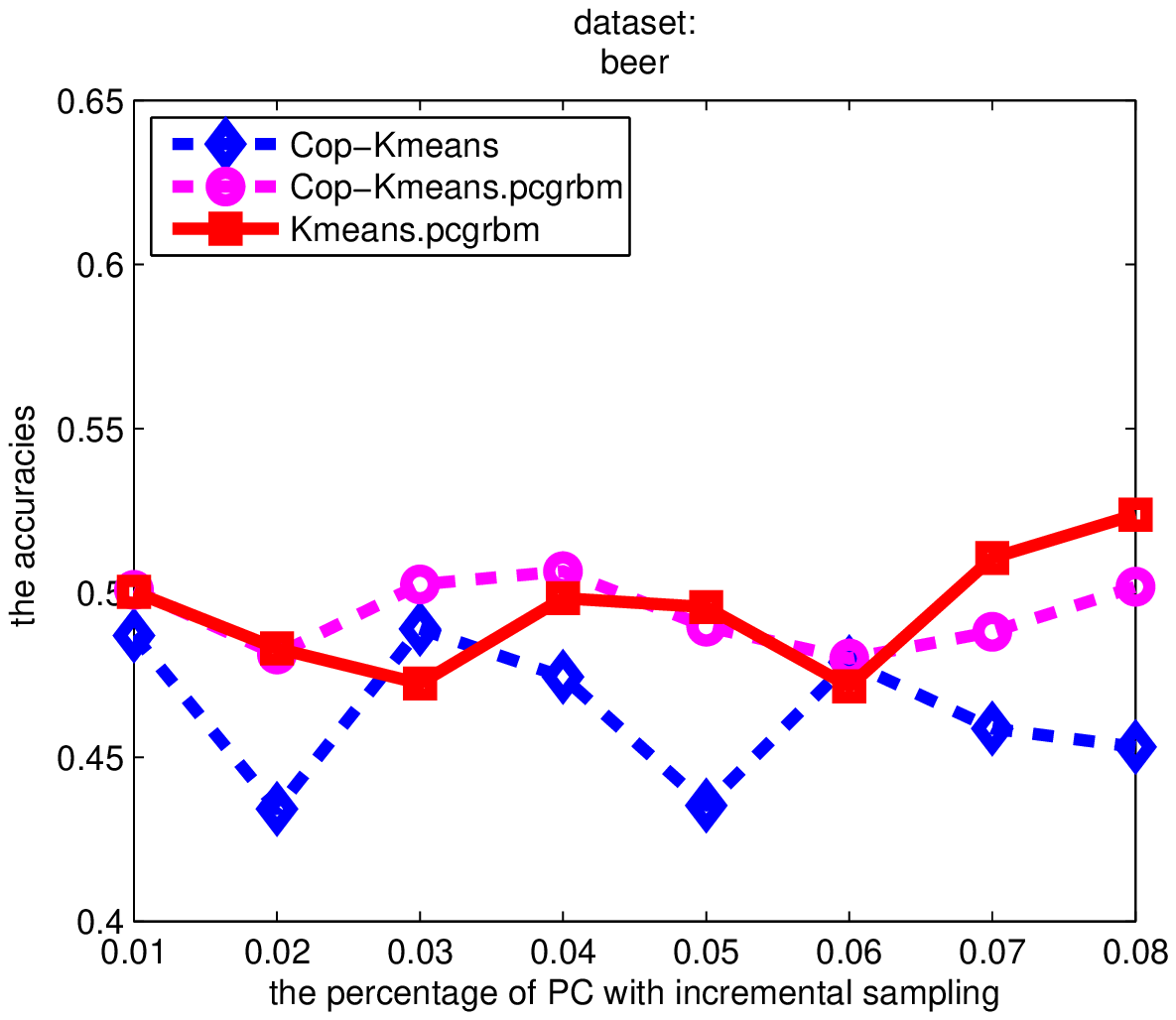}
    \includegraphics[scale=0.4225]{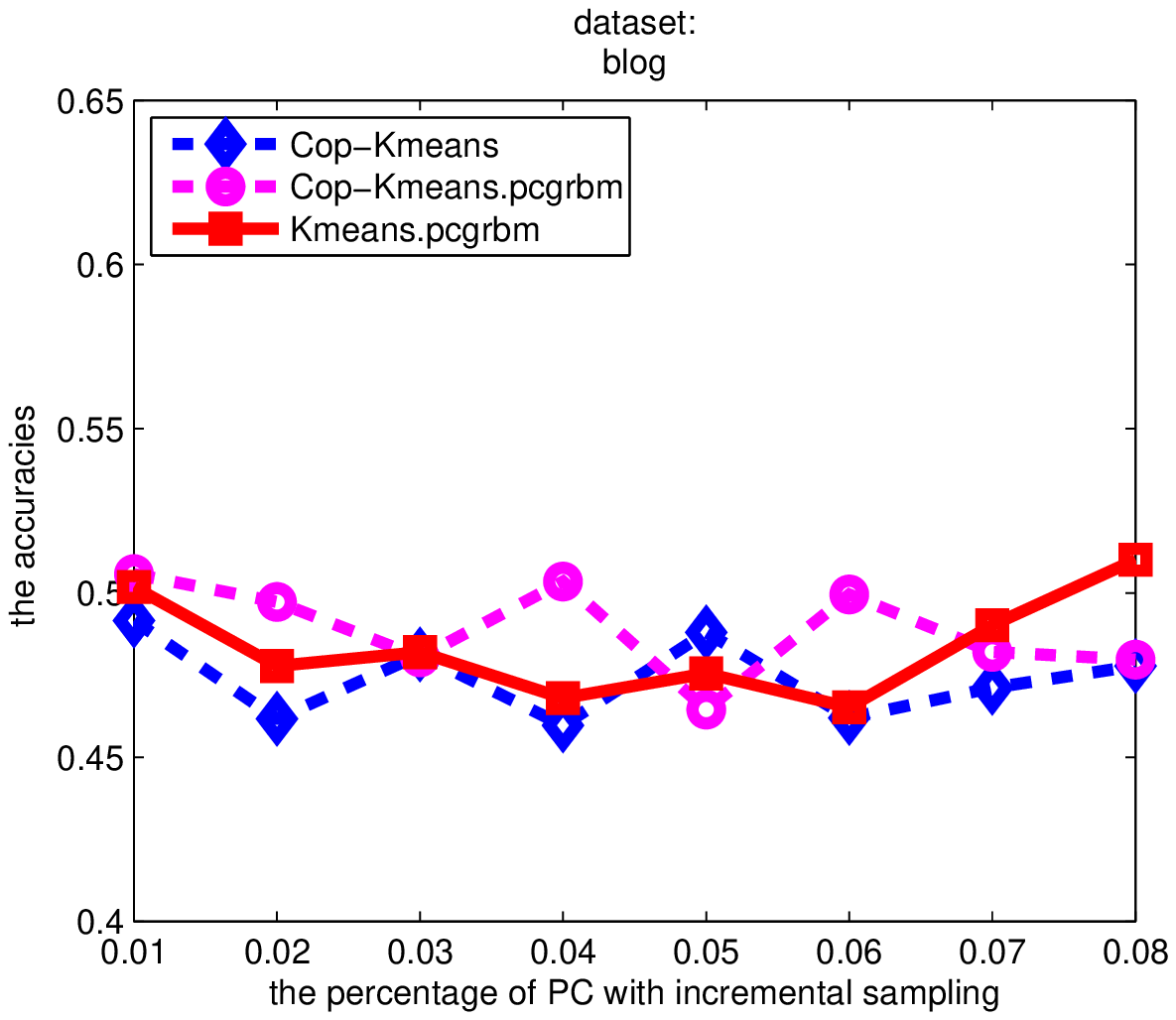}
    \includegraphics[scale=0.4225]{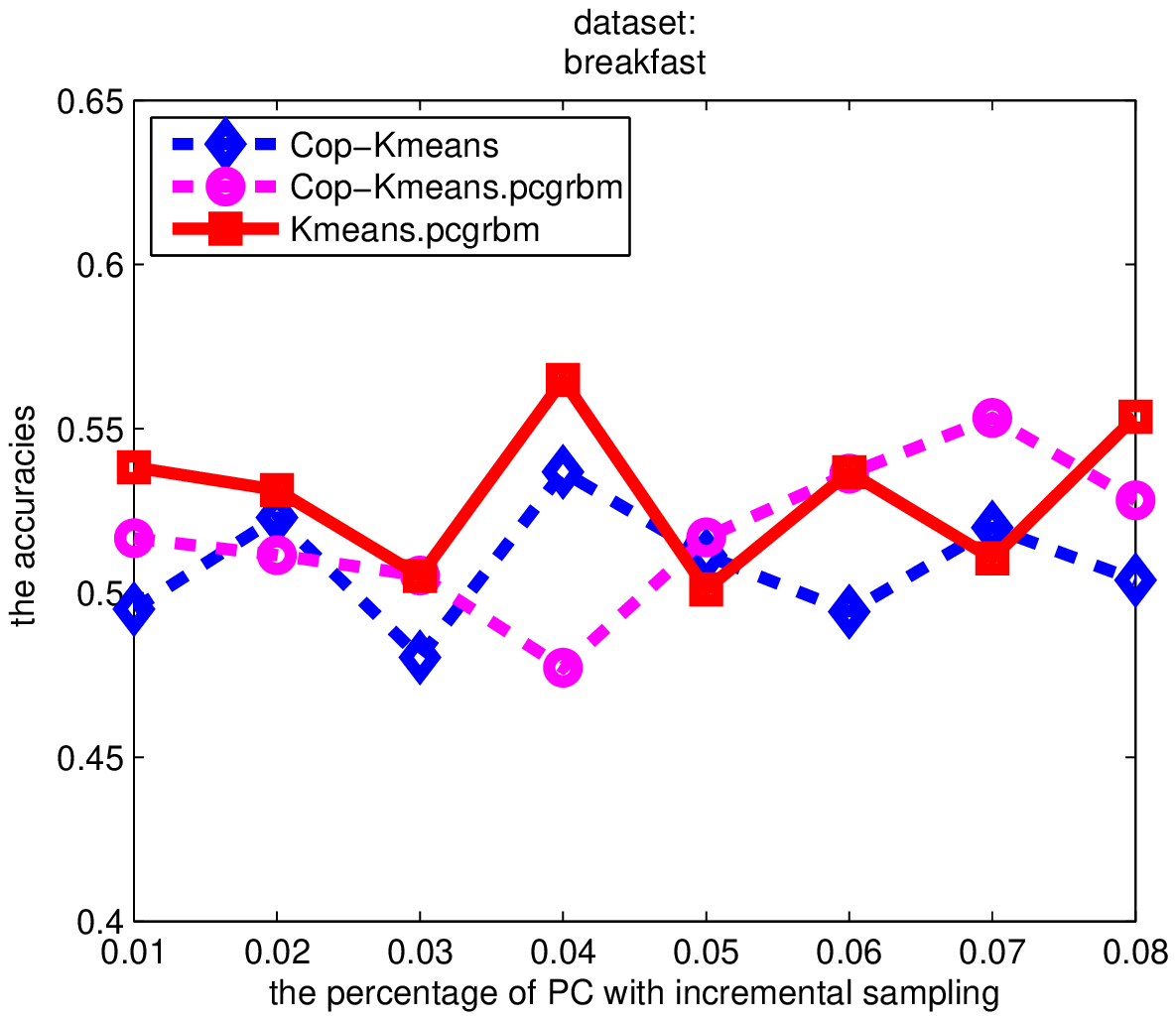}
    \includegraphics[scale=0.4225]{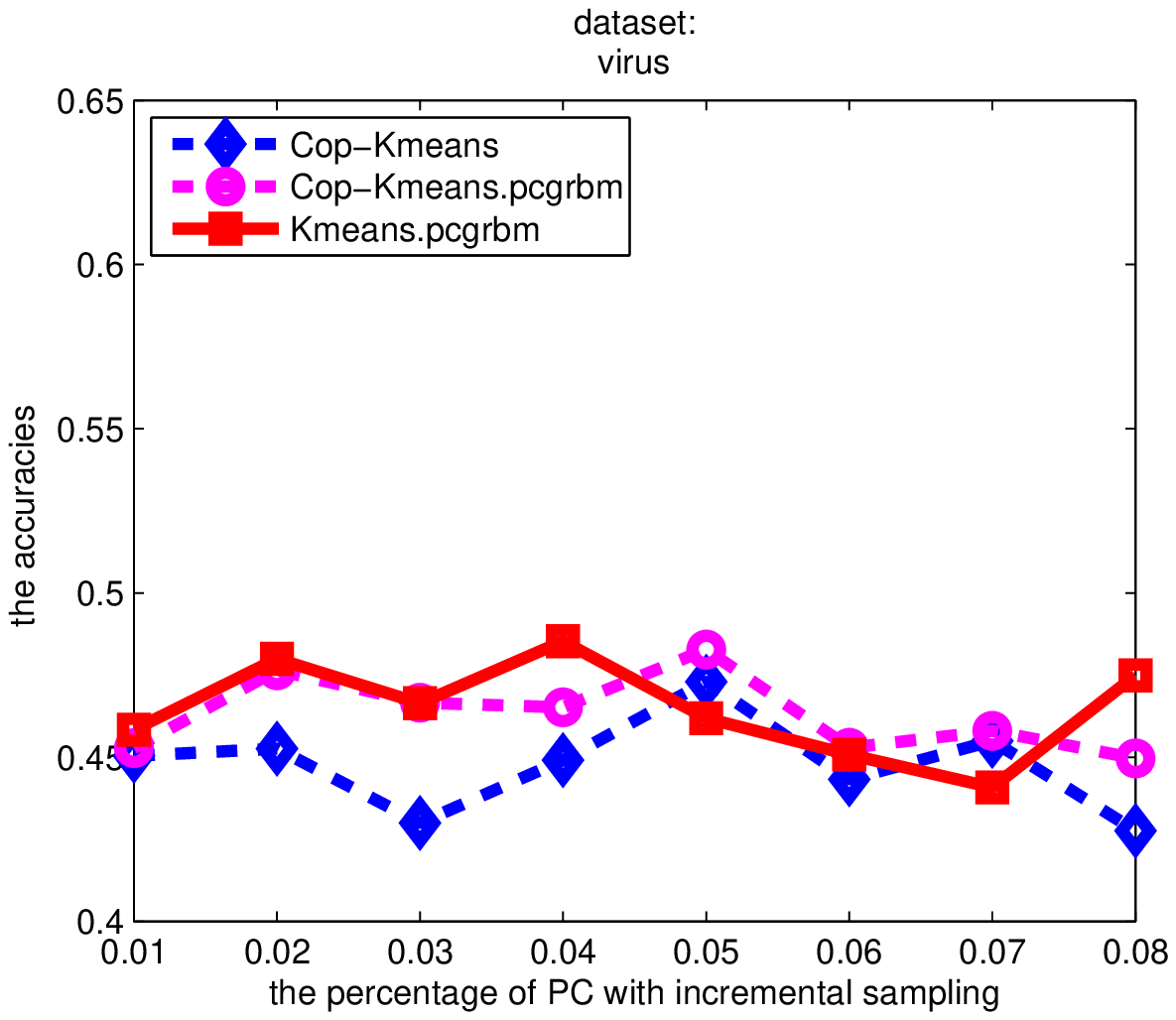}
    \includegraphics[scale=0.4225]{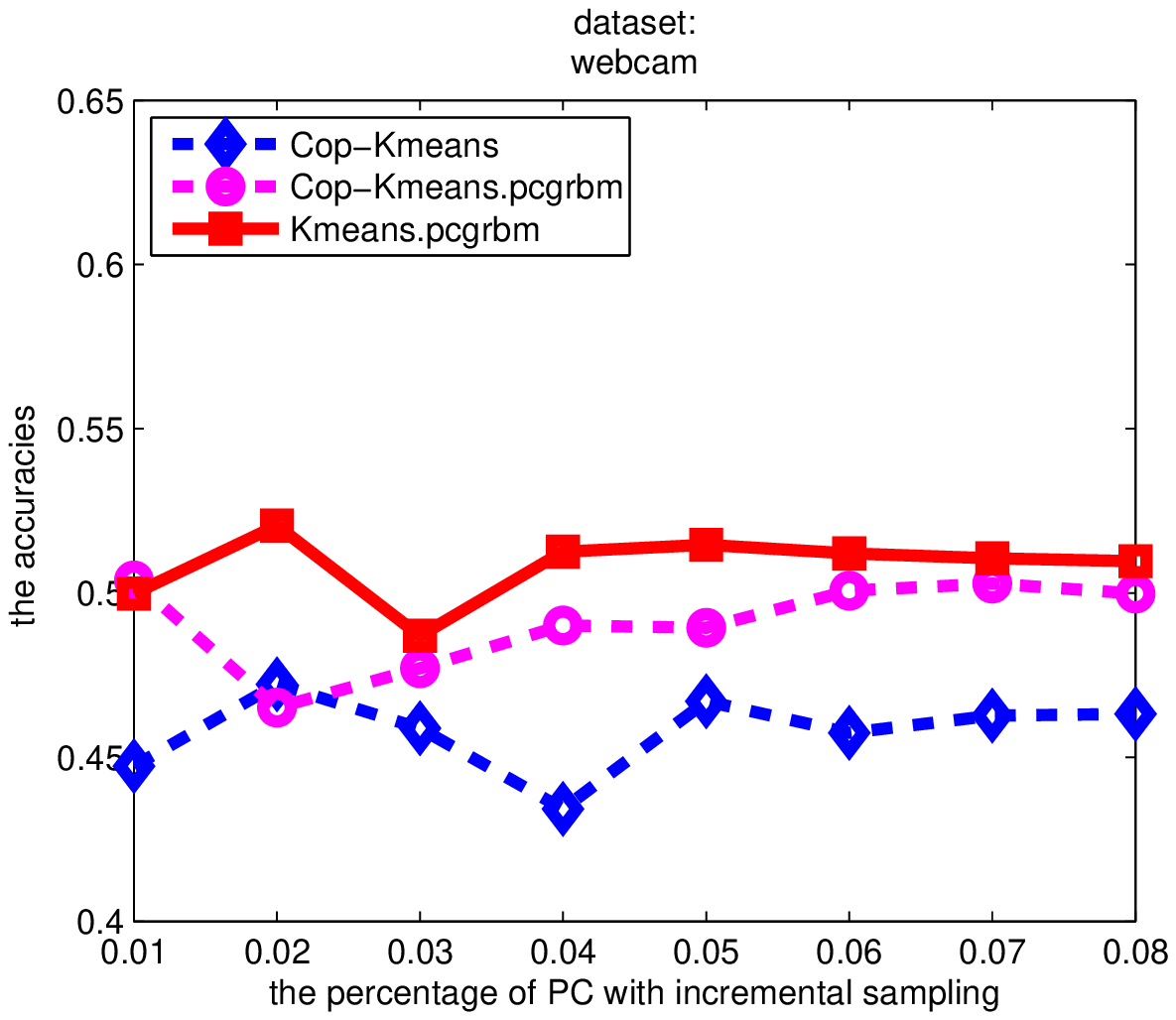}
    \includegraphics[scale=0.4225]{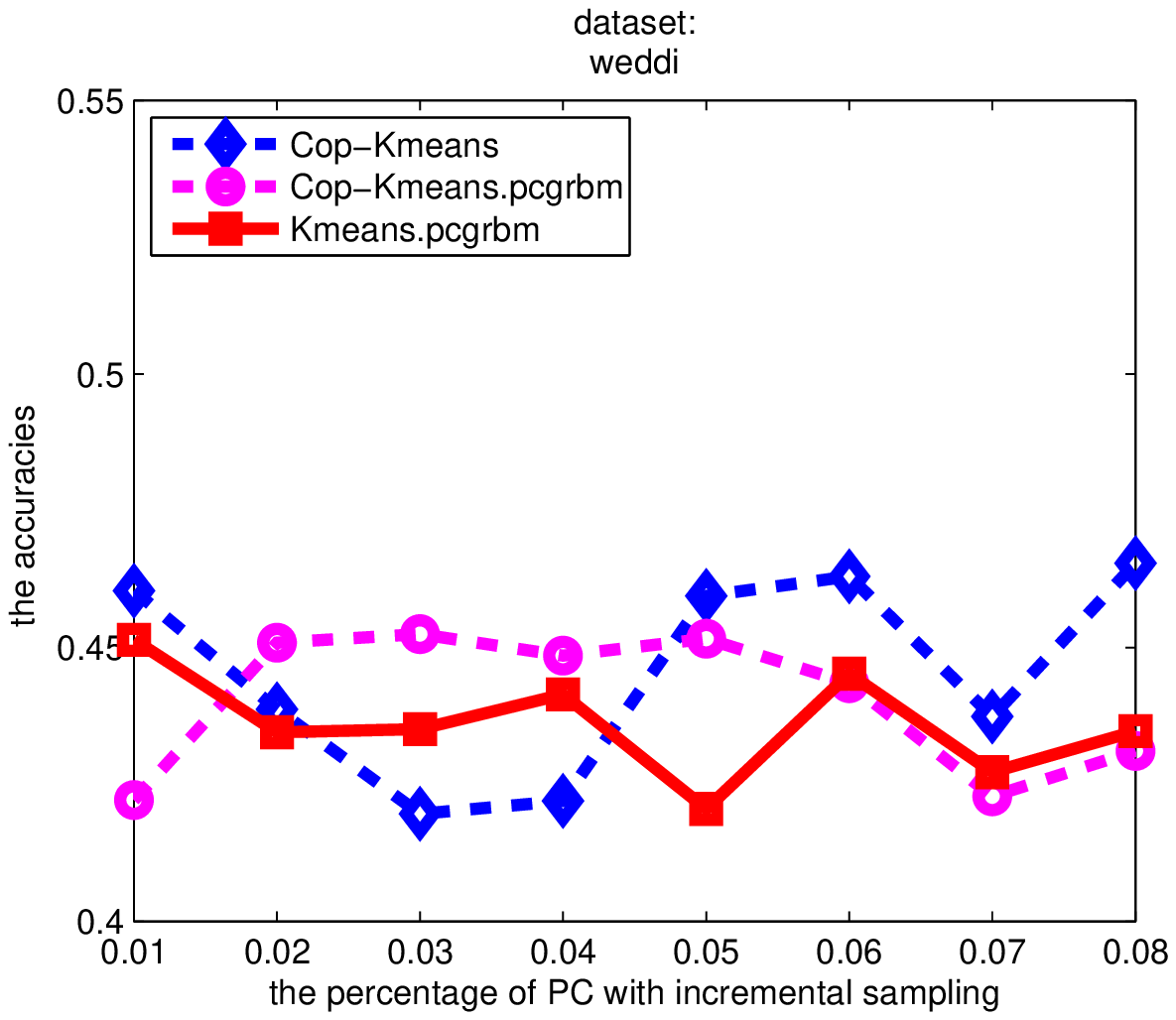}
    \includegraphics[scale=0.4225]{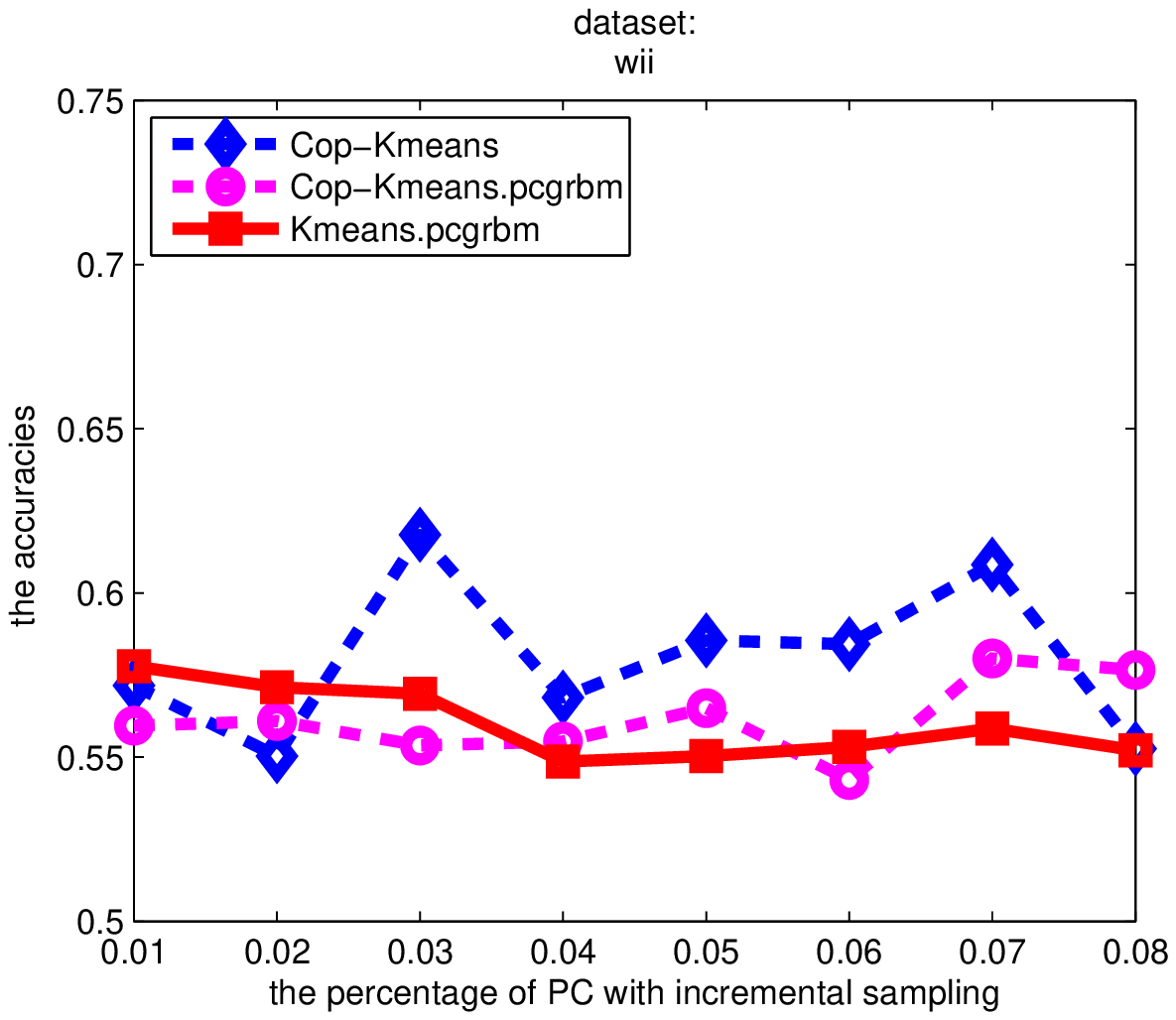}
    \includegraphics[scale=0.4225]{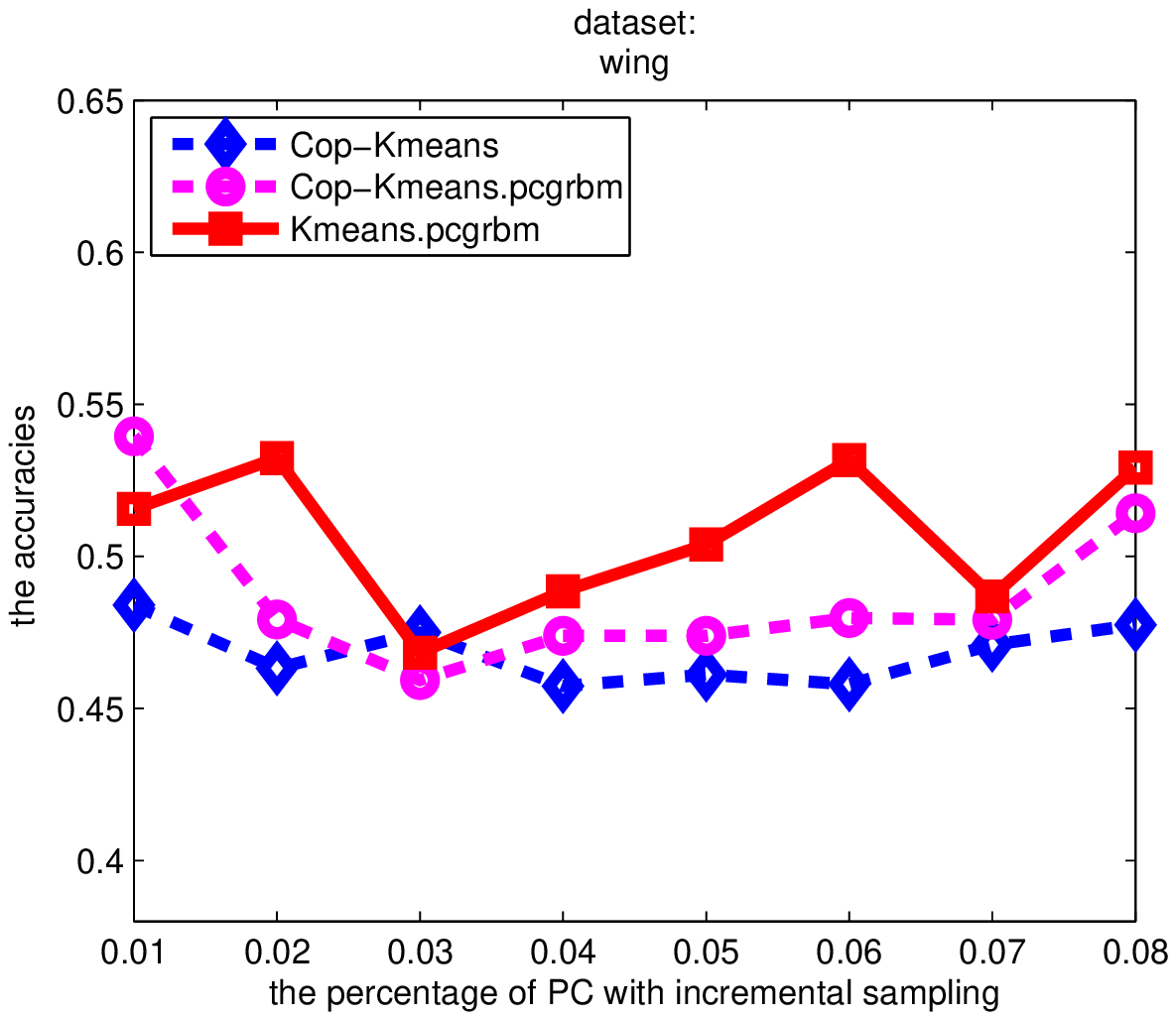}
\\
\caption{Cop-Kmeans, Cop-Kmeans.pcgrbm and Kmeans.pcgrbm results with an increasing percentage of pairwise constraints (PC) from 1\% to 8\% in steps of 1\% by the incremental sampling method.
} \label{fig:1}
\end{figure*}
\begin{figure*}
\vspace{1mm} \centering
    \includegraphics[scale=0.4225]{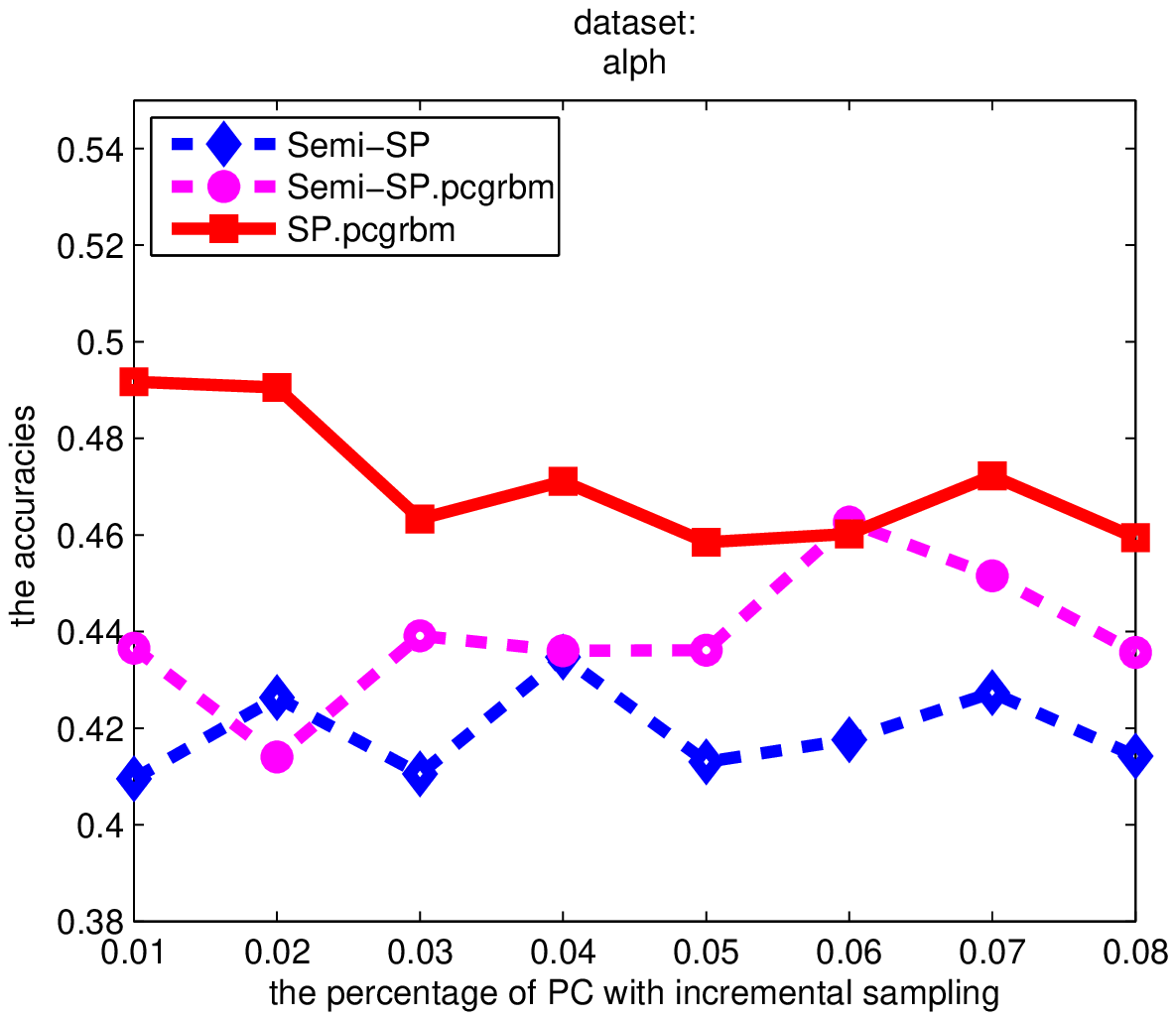}
     \includegraphics[scale=0.4225]{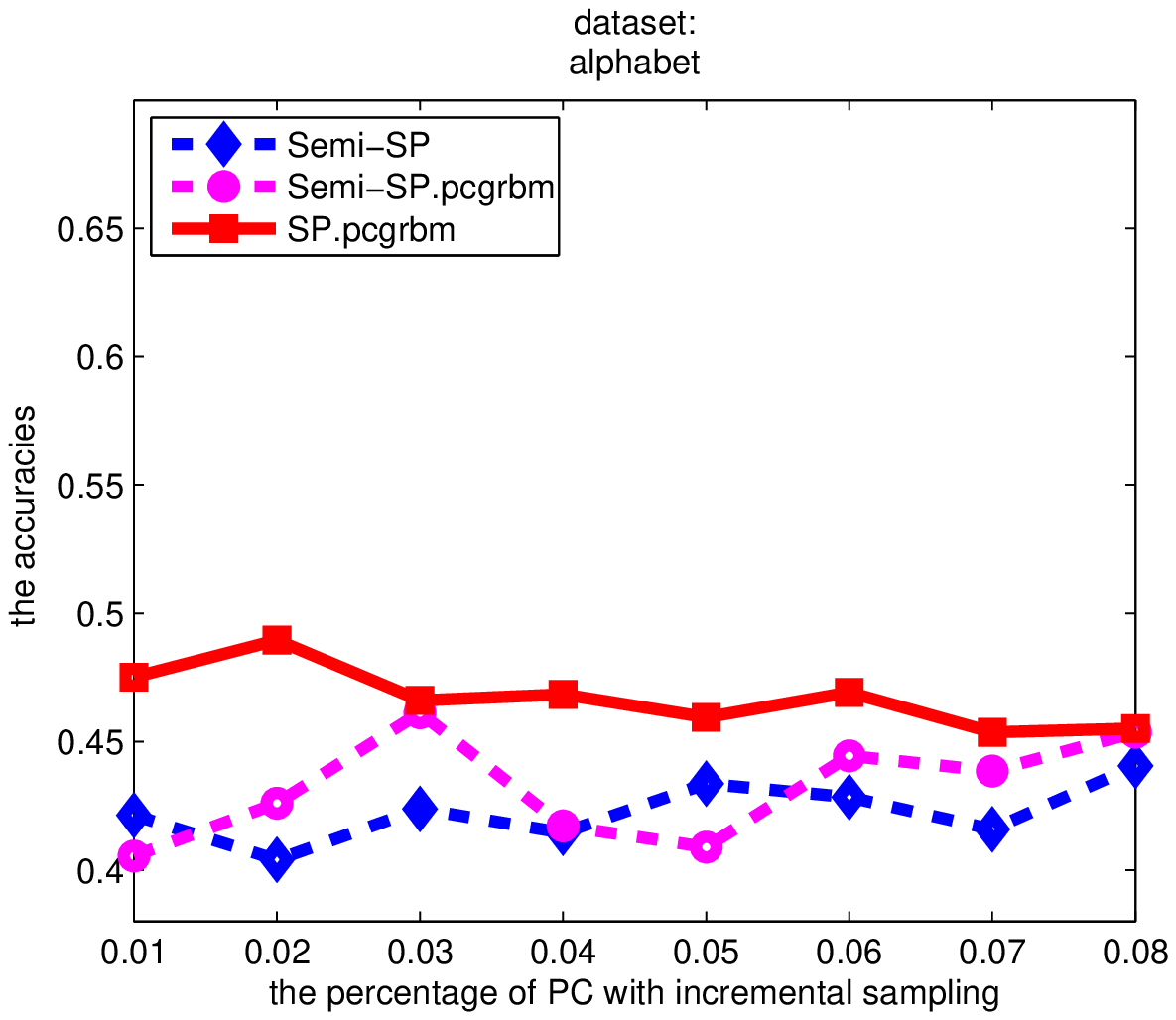}
    \includegraphics[scale=0.4225]{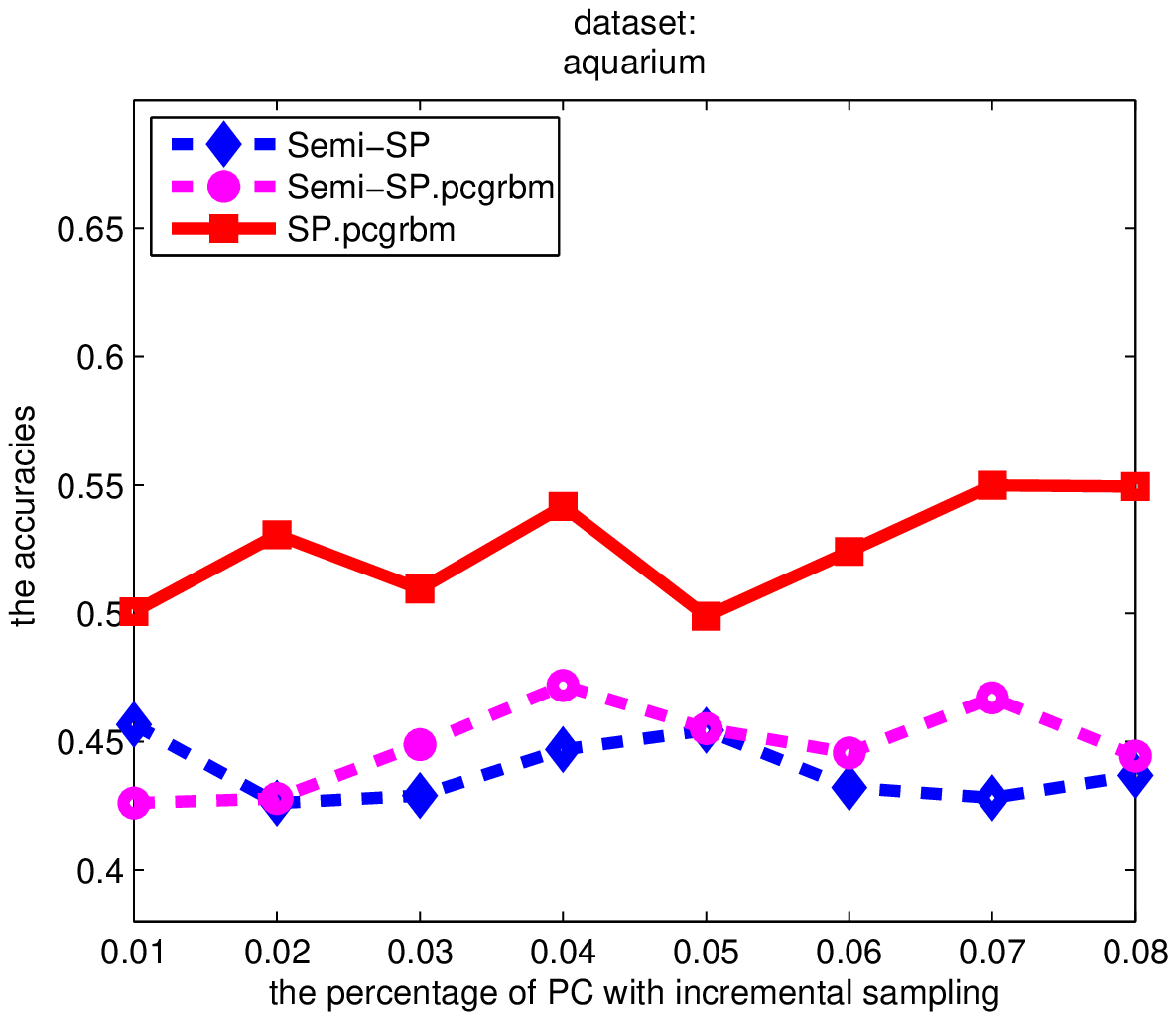}
    \includegraphics[scale=0.4225]{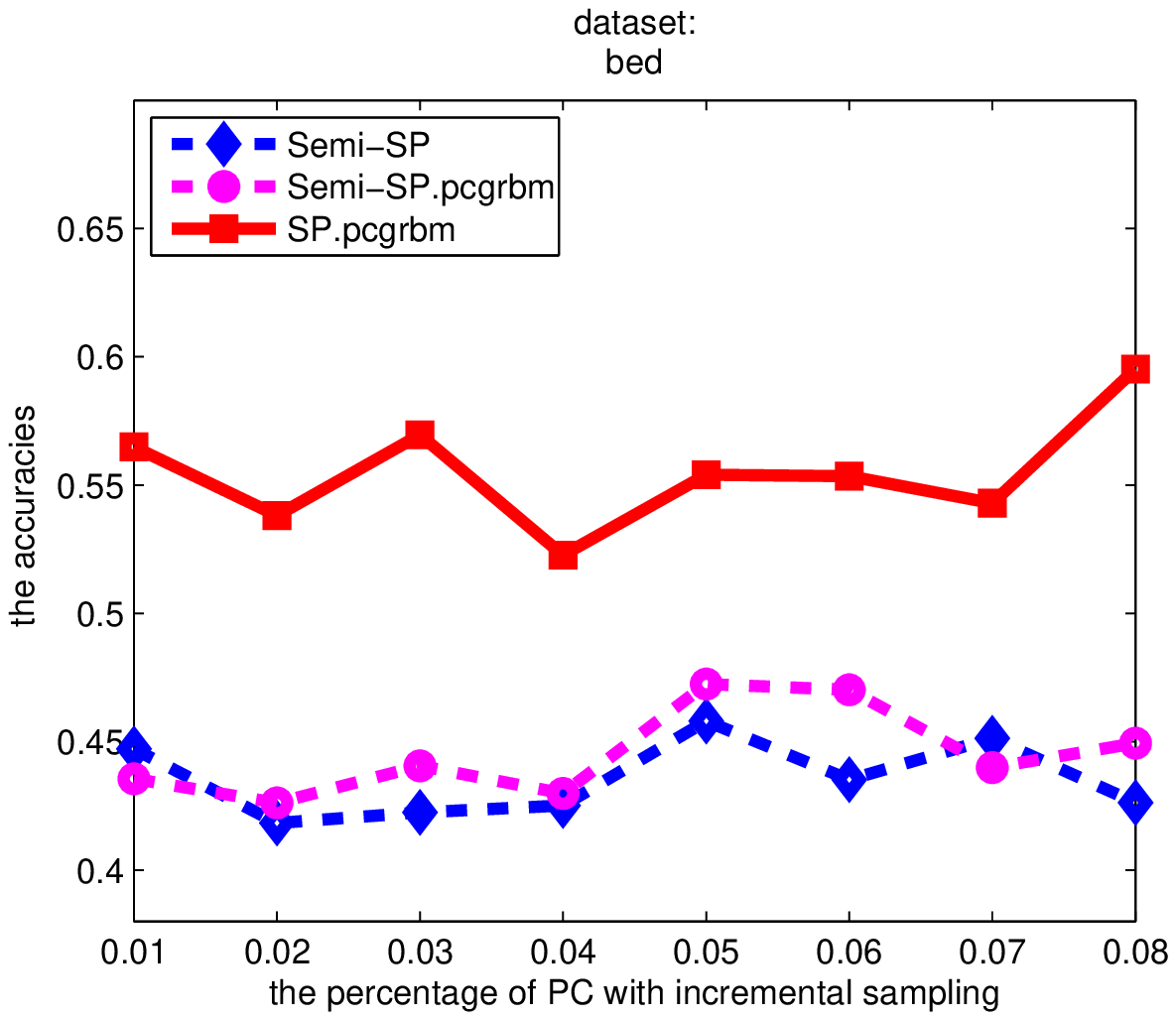}
    \includegraphics[scale=0.4225]{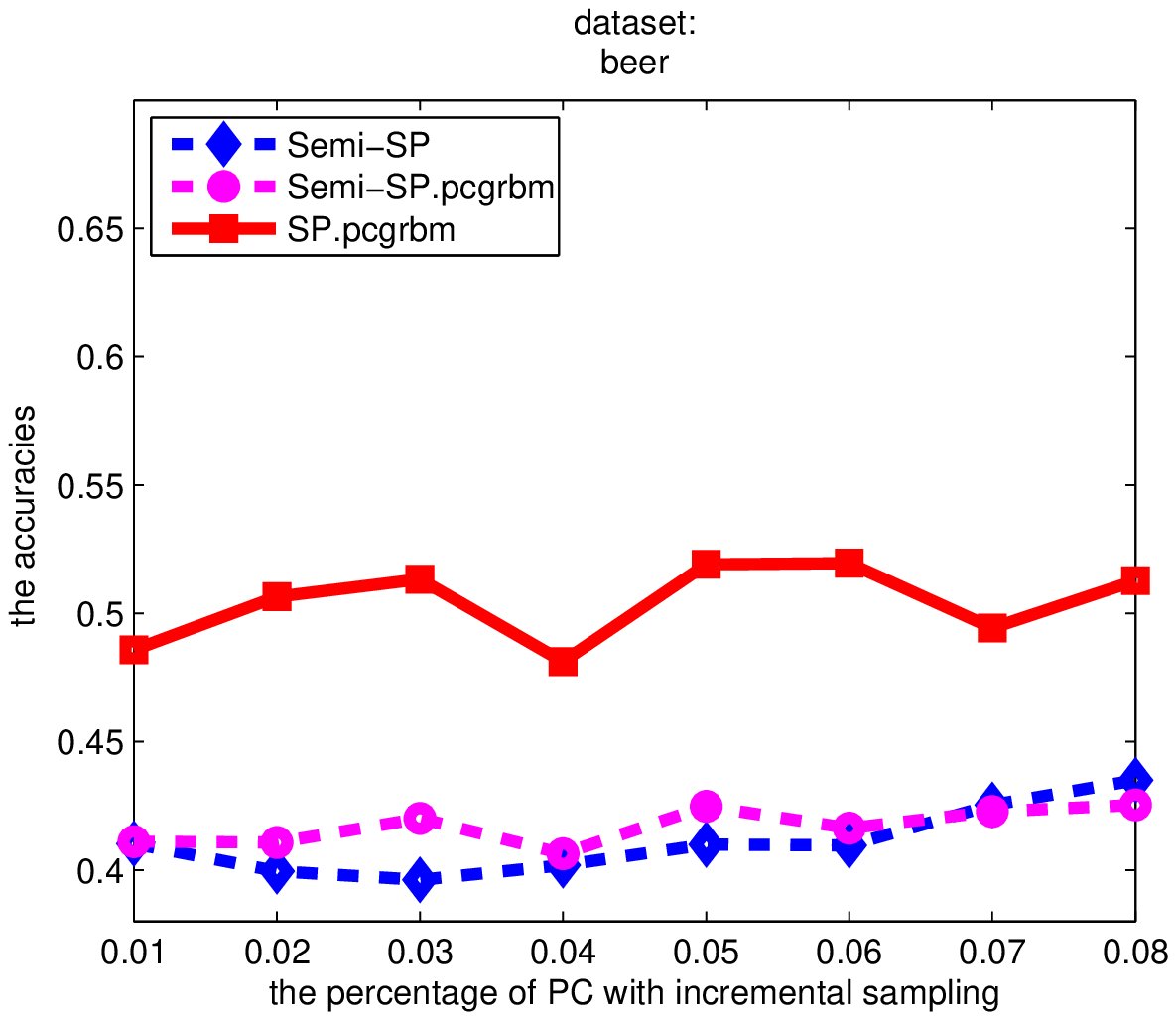}
    \includegraphics[scale=0.4225]{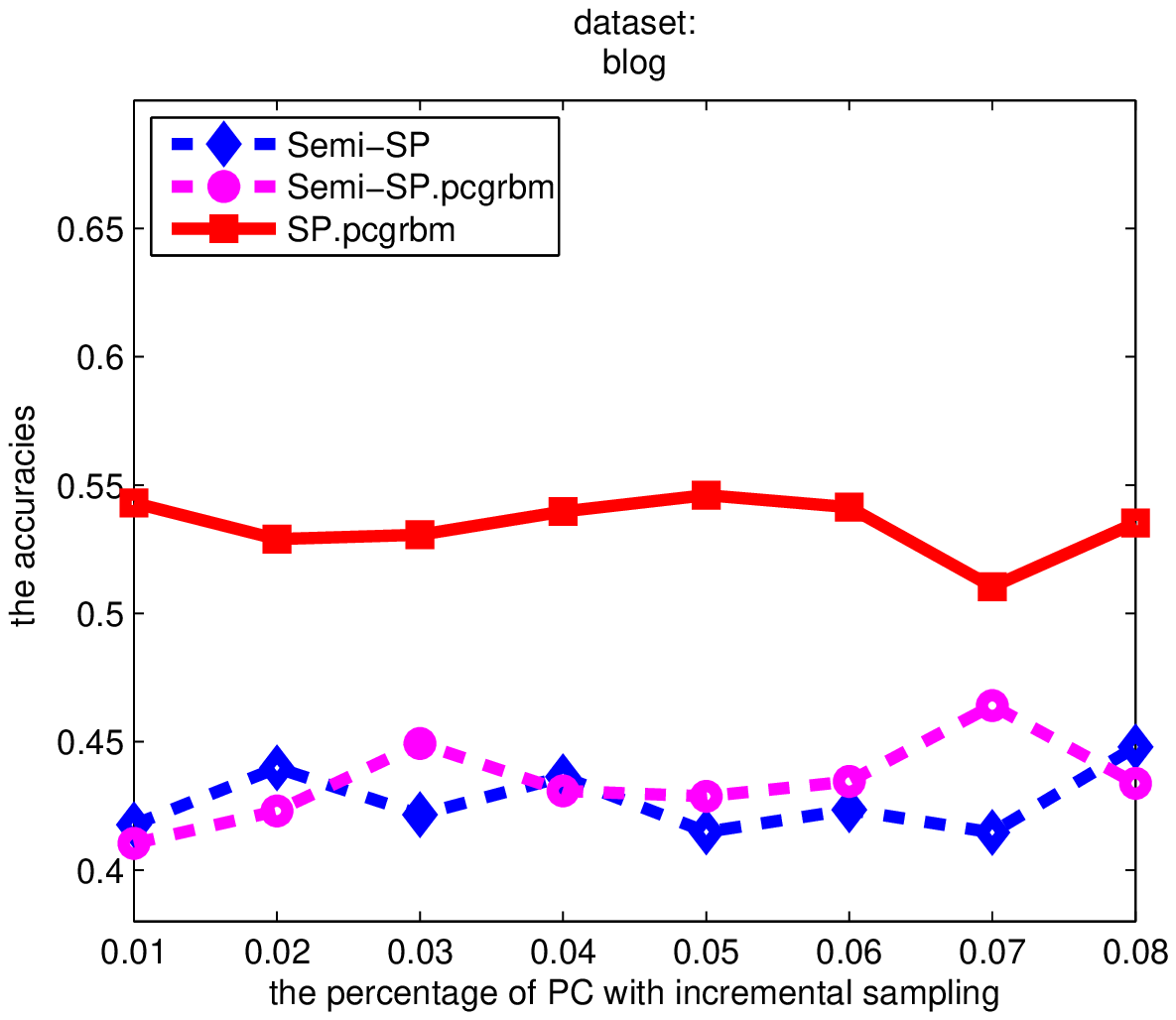}
    \includegraphics[scale=0.4225]{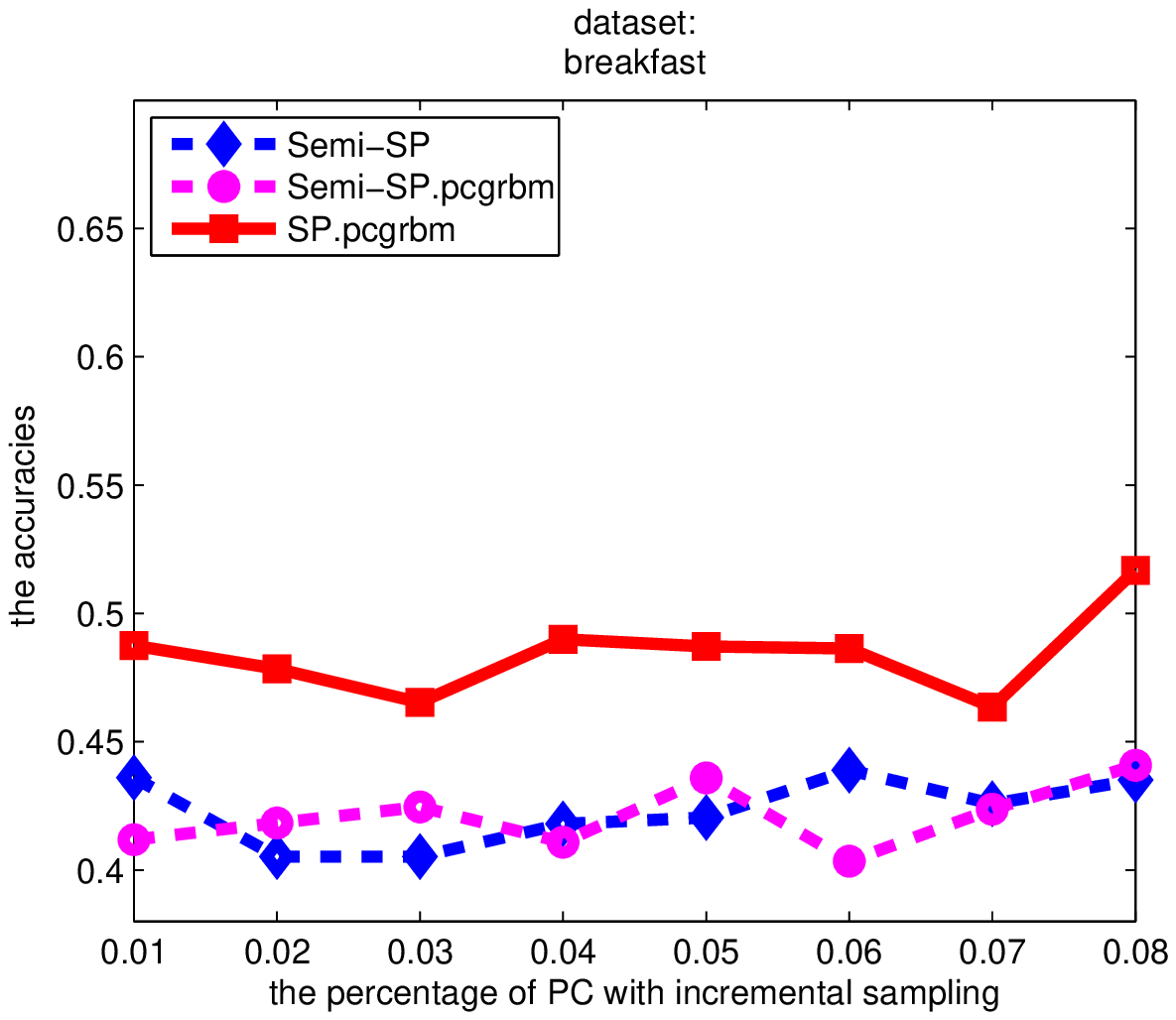}
    \includegraphics[scale=0.4225]{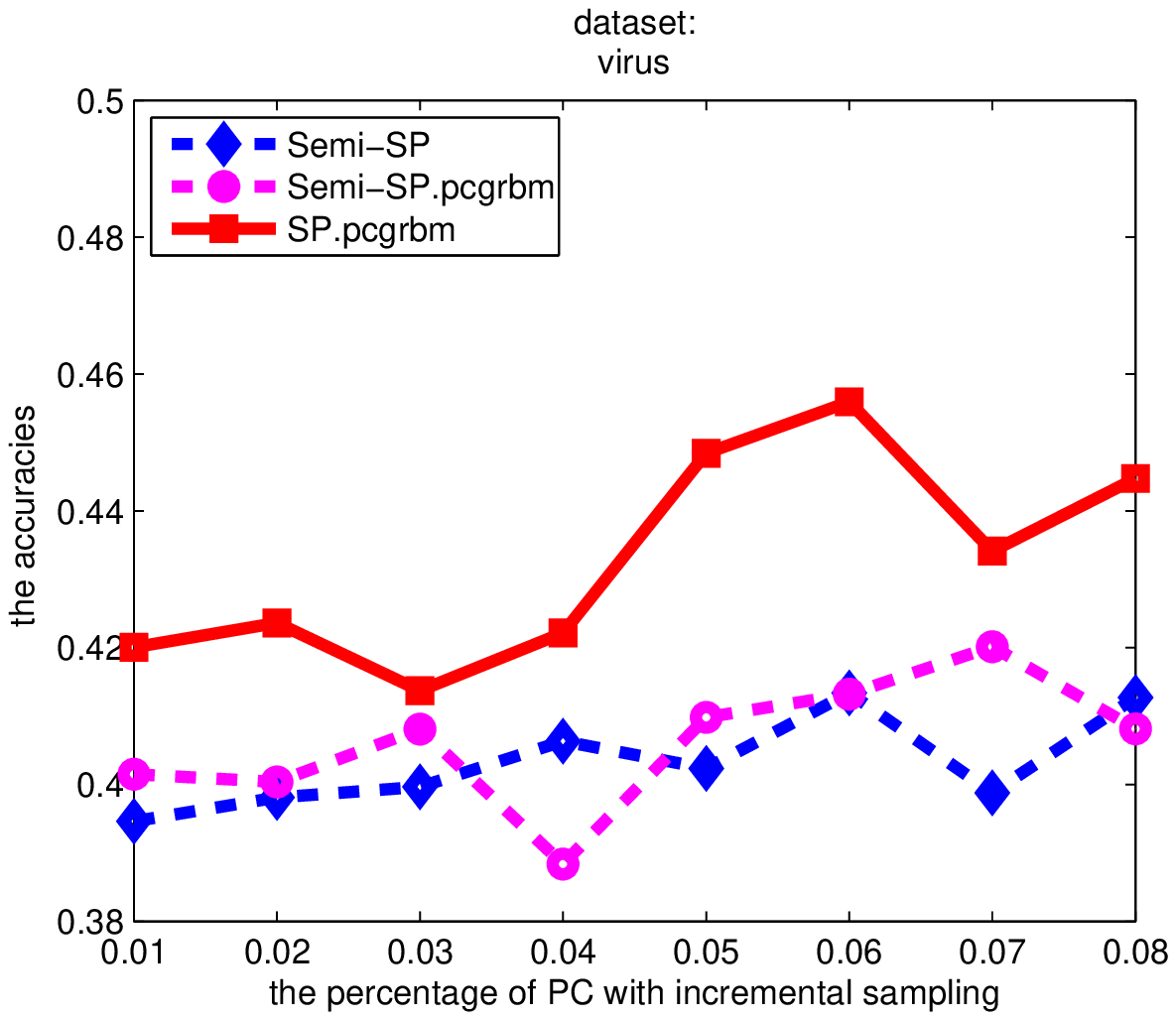}
    \includegraphics[scale=0.4225]{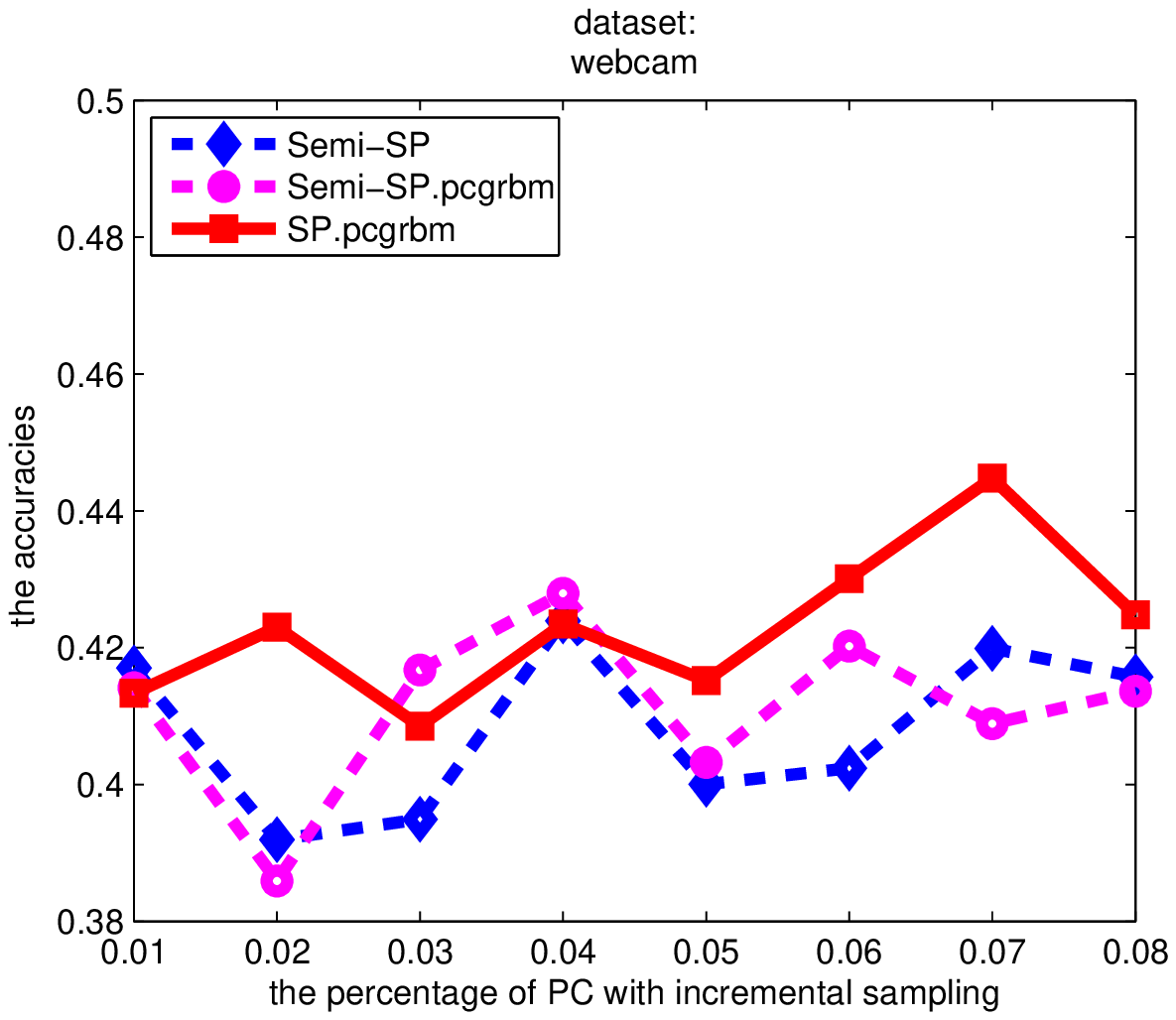}
    \includegraphics[scale=0.4225]{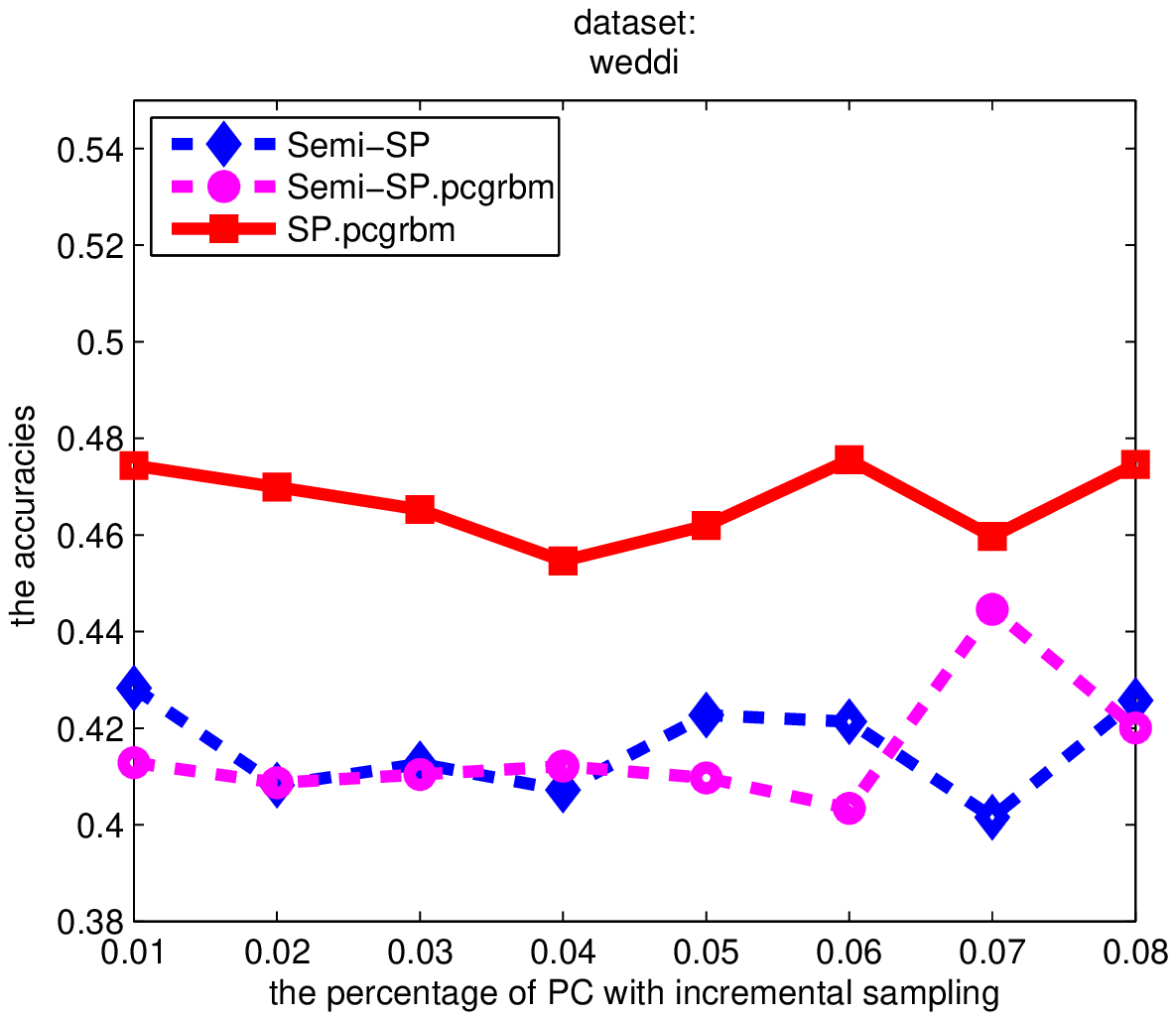}
    \includegraphics[scale=0.4225]{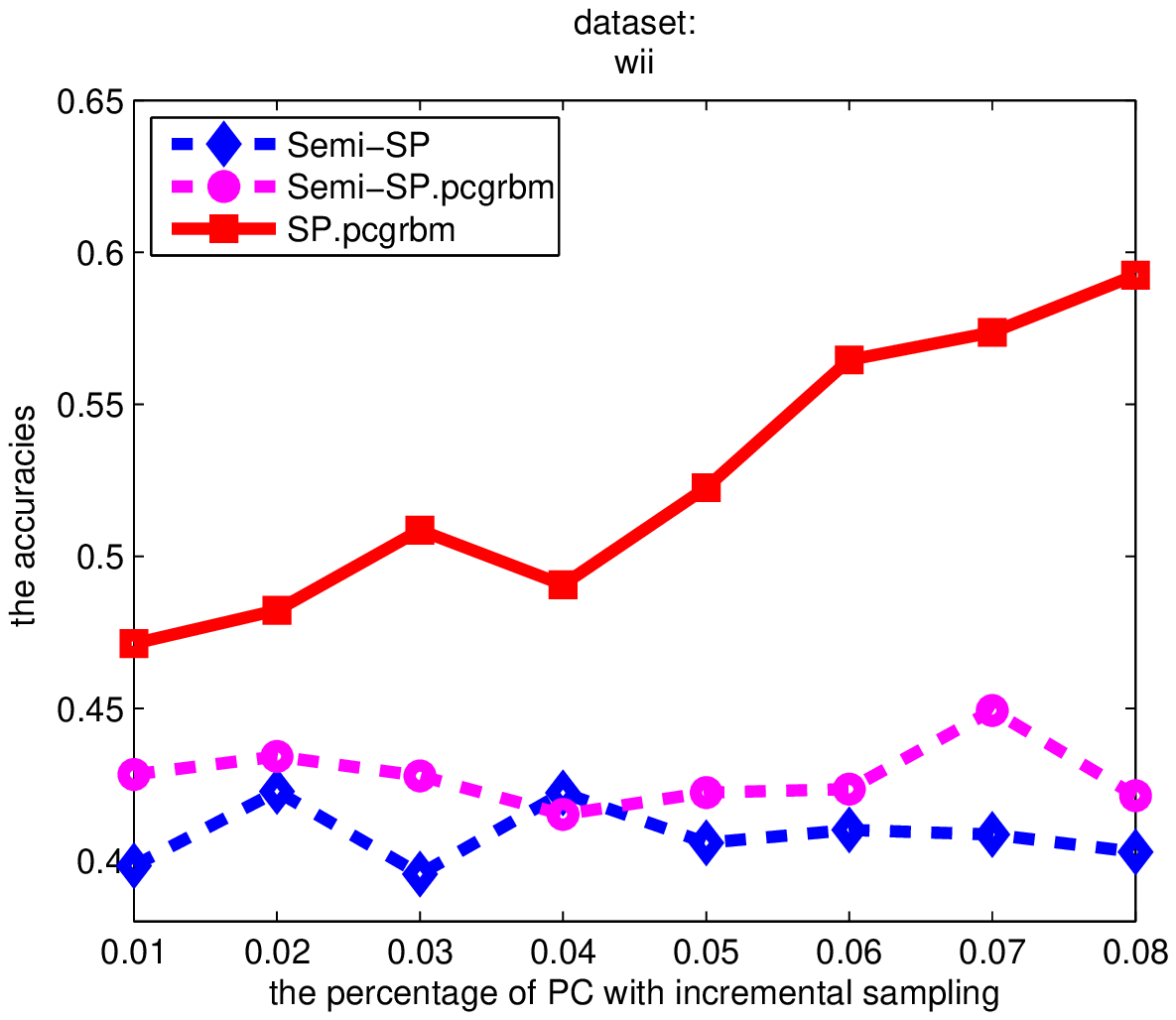}
    \includegraphics[scale=0.4225]{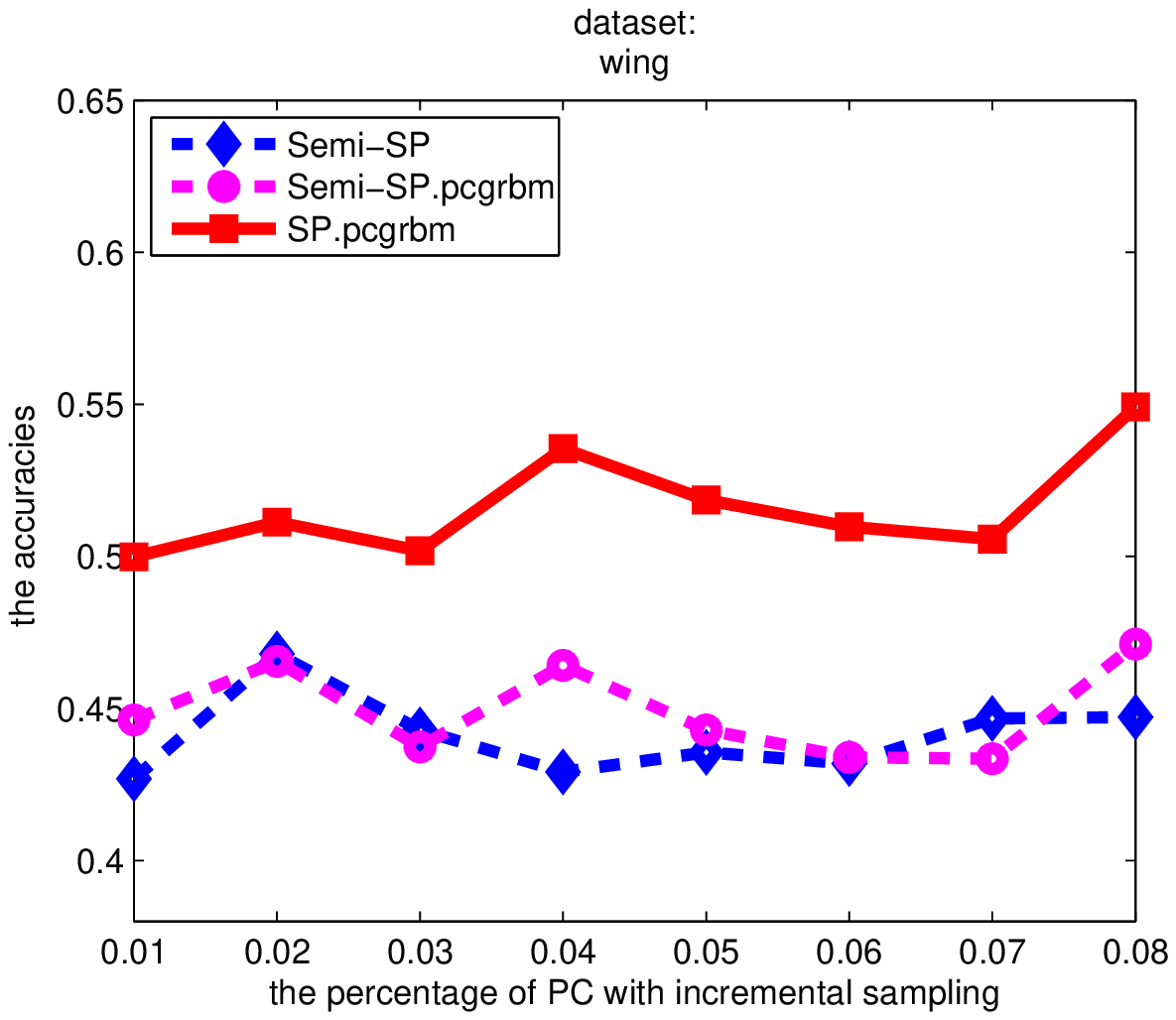}
\\
\caption{Semi-SP, Semi-SP.pcgrbm and SP.pcgrbm results with an increasing percentage of pairwise constraints (PC) from 1\% to 8\% in steps of 1\% by the incremental sampling method.
} \label{fig:1}
\end{figure*}
\begin{figure*}
\vspace{1mm} \centering
    \includegraphics[scale=0.4225]{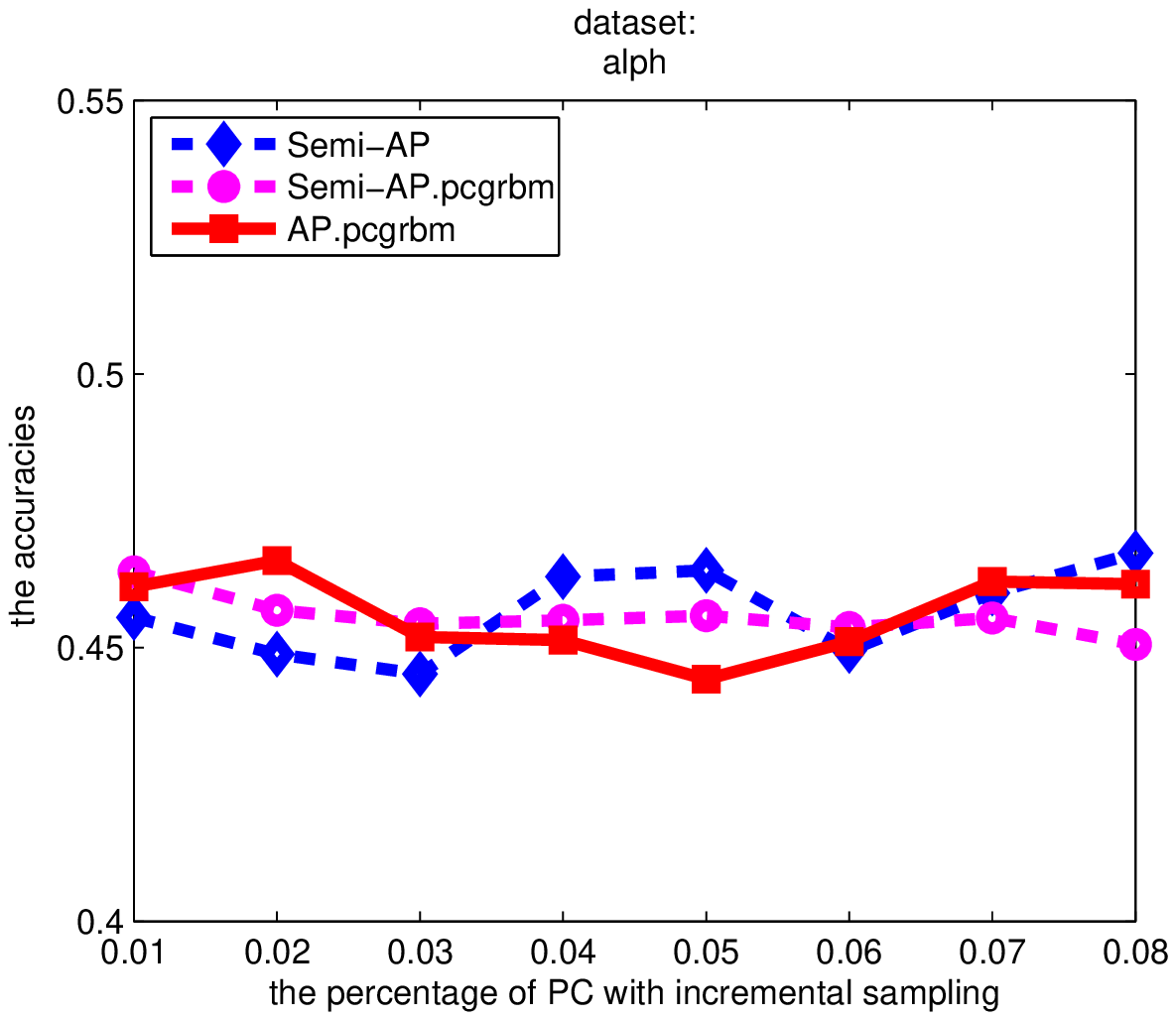}
   \includegraphics[scale=0.4225]{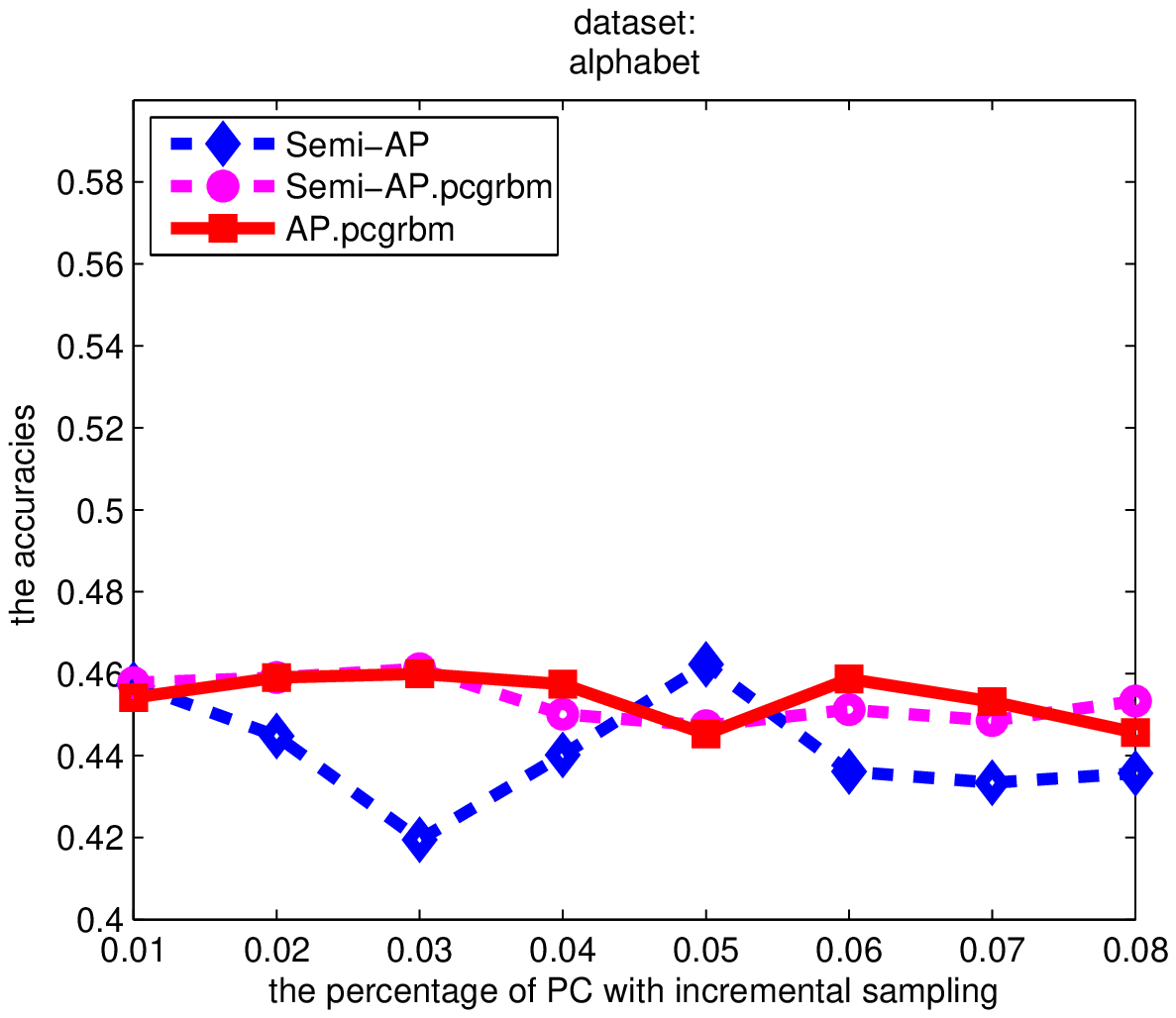}
    \includegraphics[scale=0.4225]{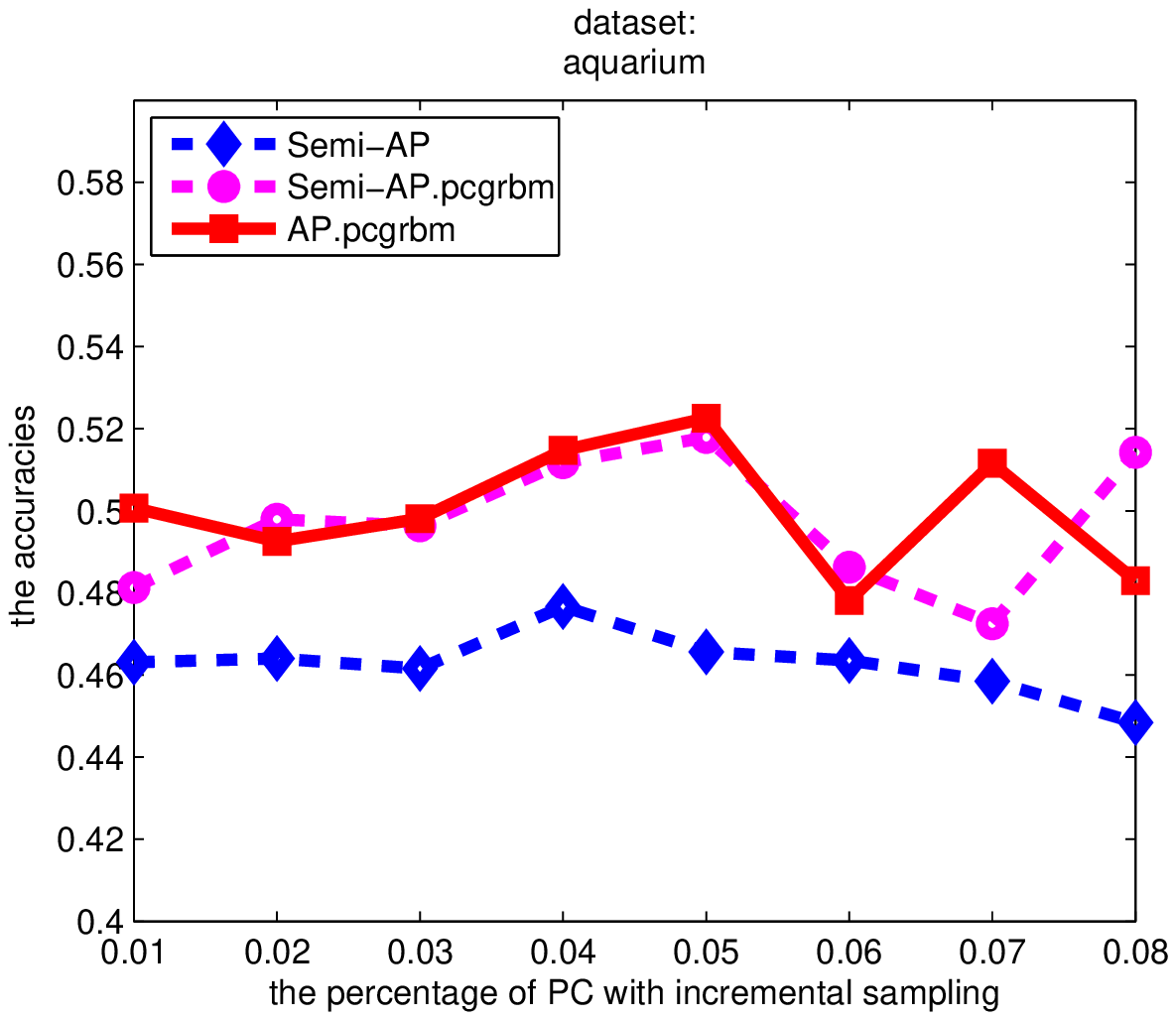}
    \includegraphics[scale=0.4225]{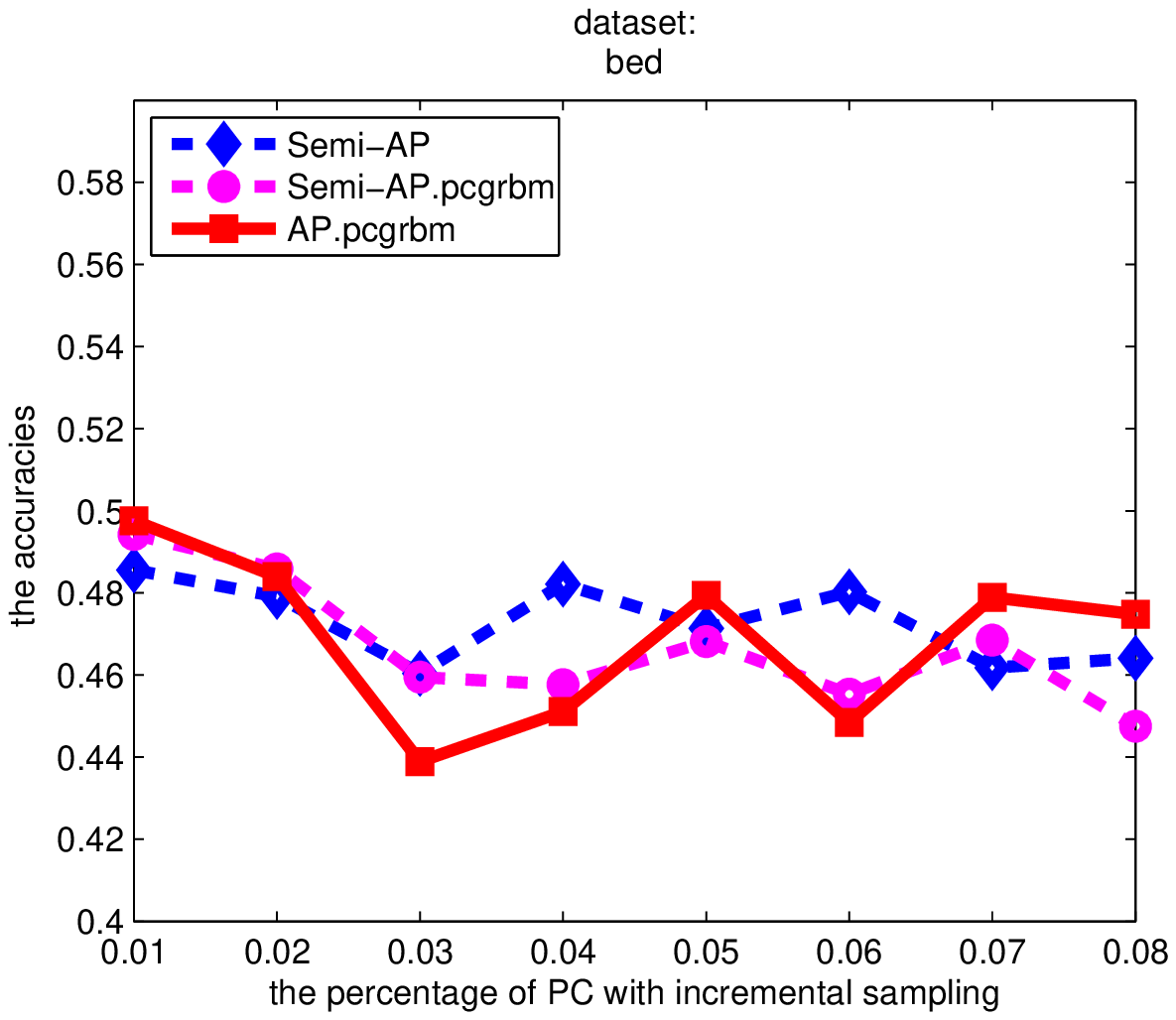}
    \includegraphics[scale=0.4225]{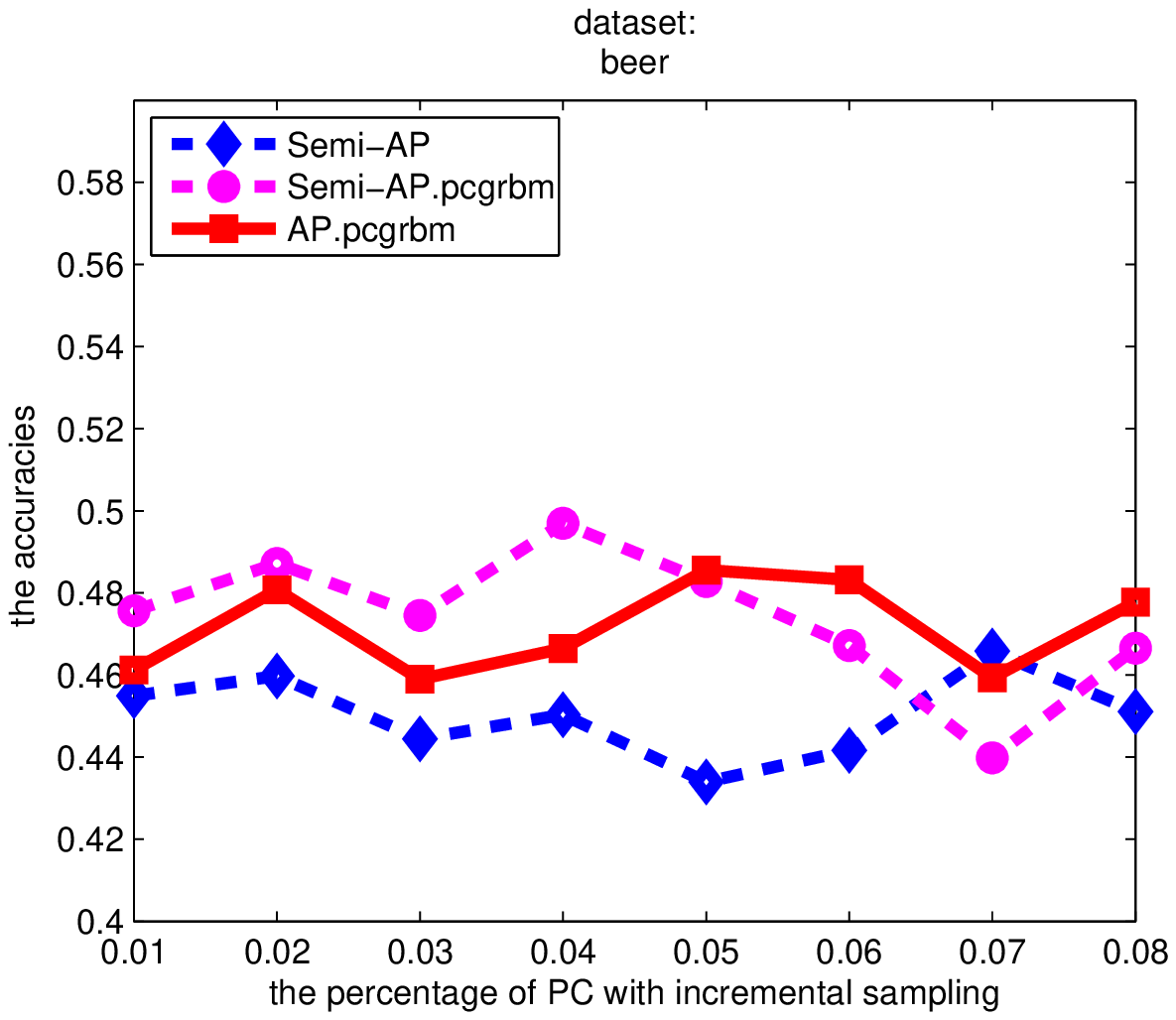}
    \includegraphics[scale=0.4225]{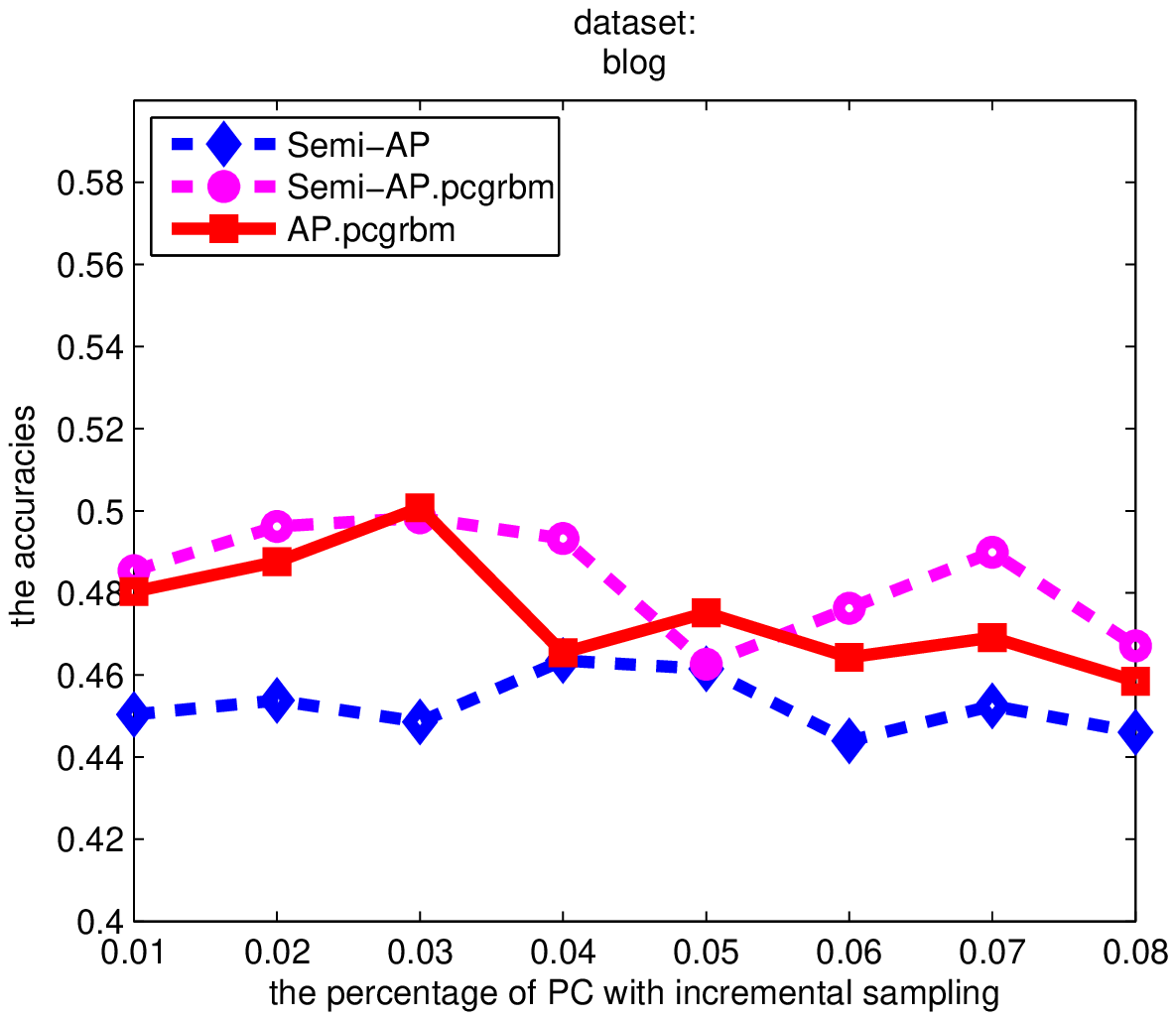}
    \includegraphics[scale=0.4225]{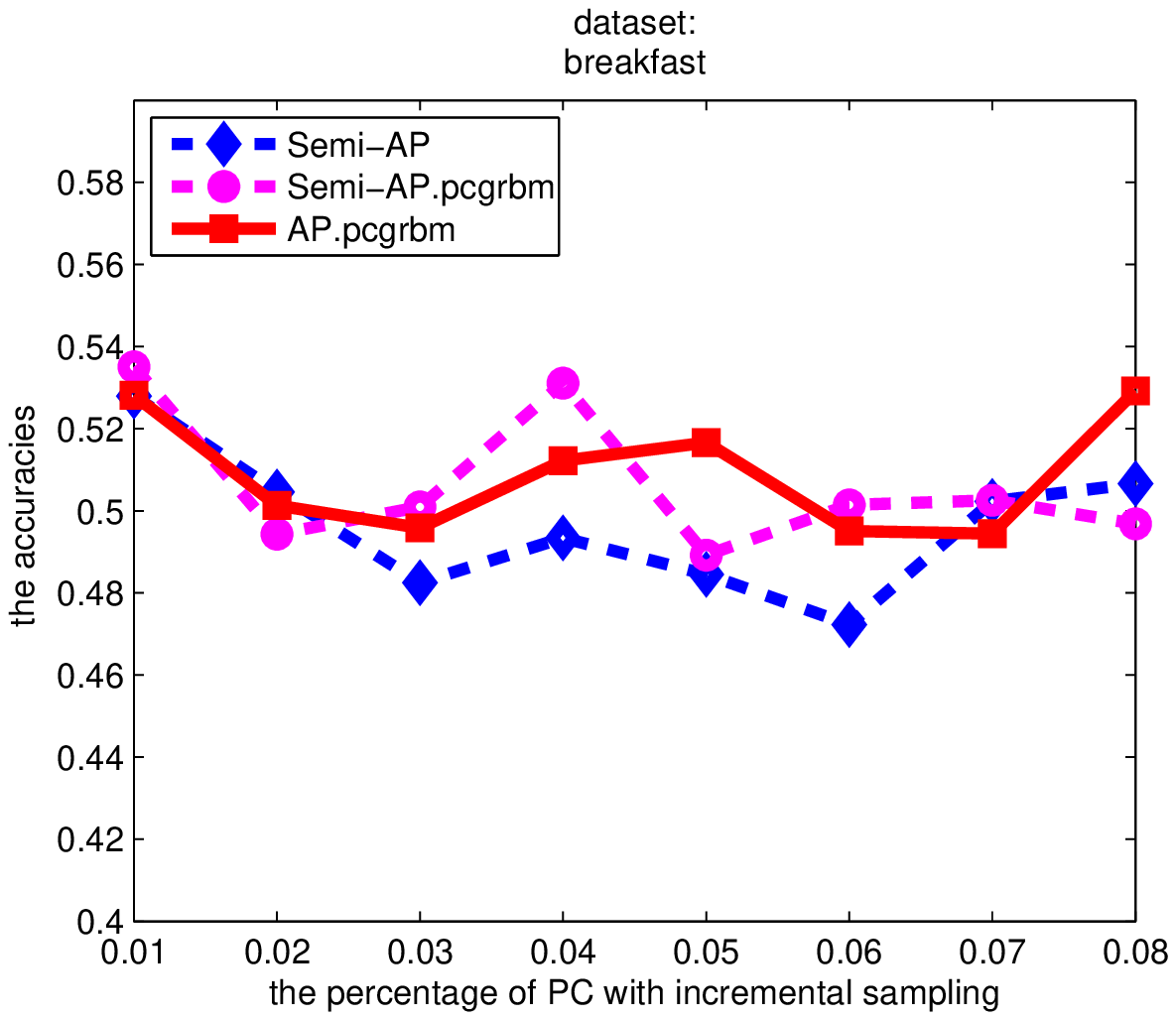}
    \includegraphics[scale=0.4225]{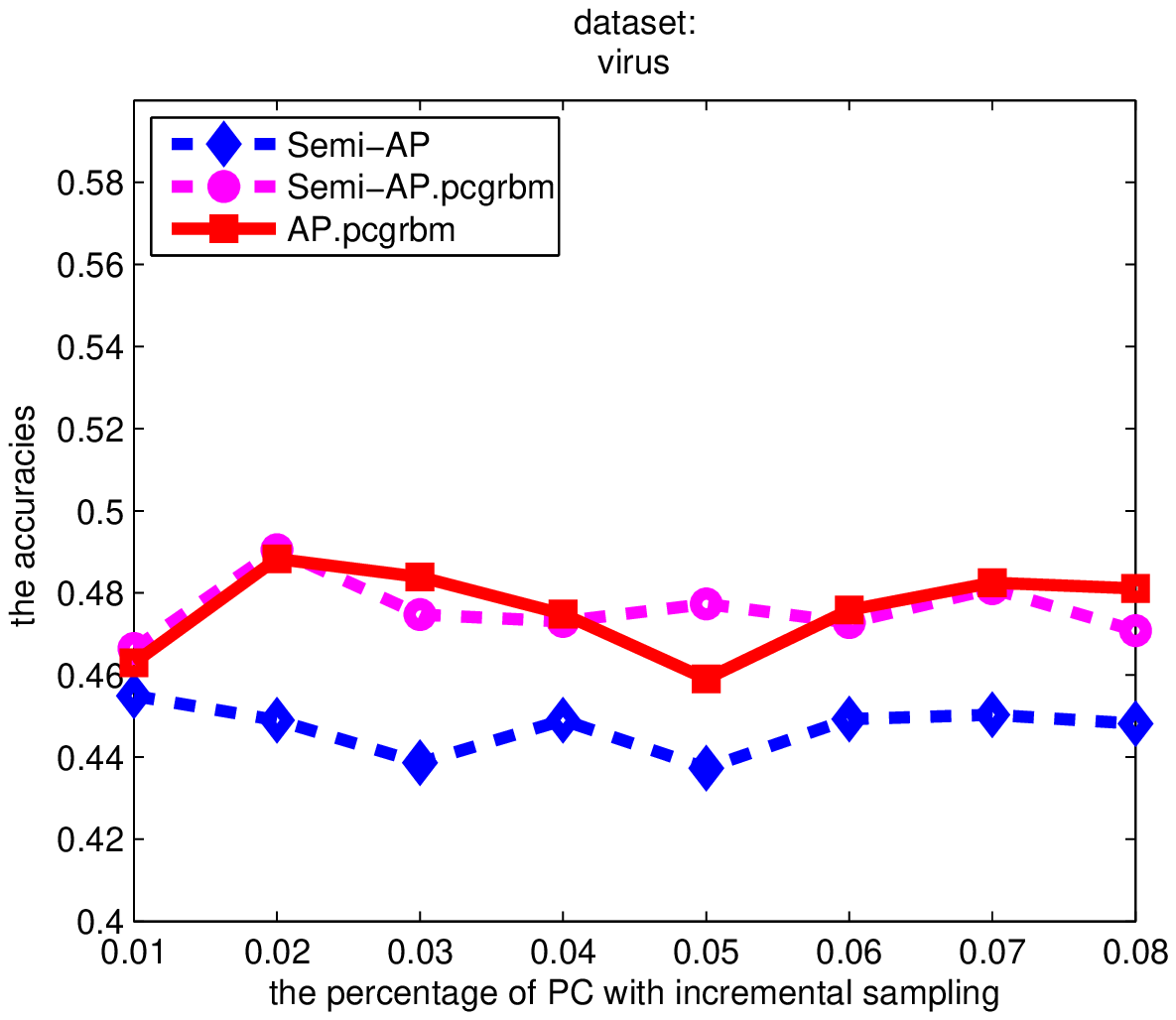}
    \includegraphics[scale=0.4225]{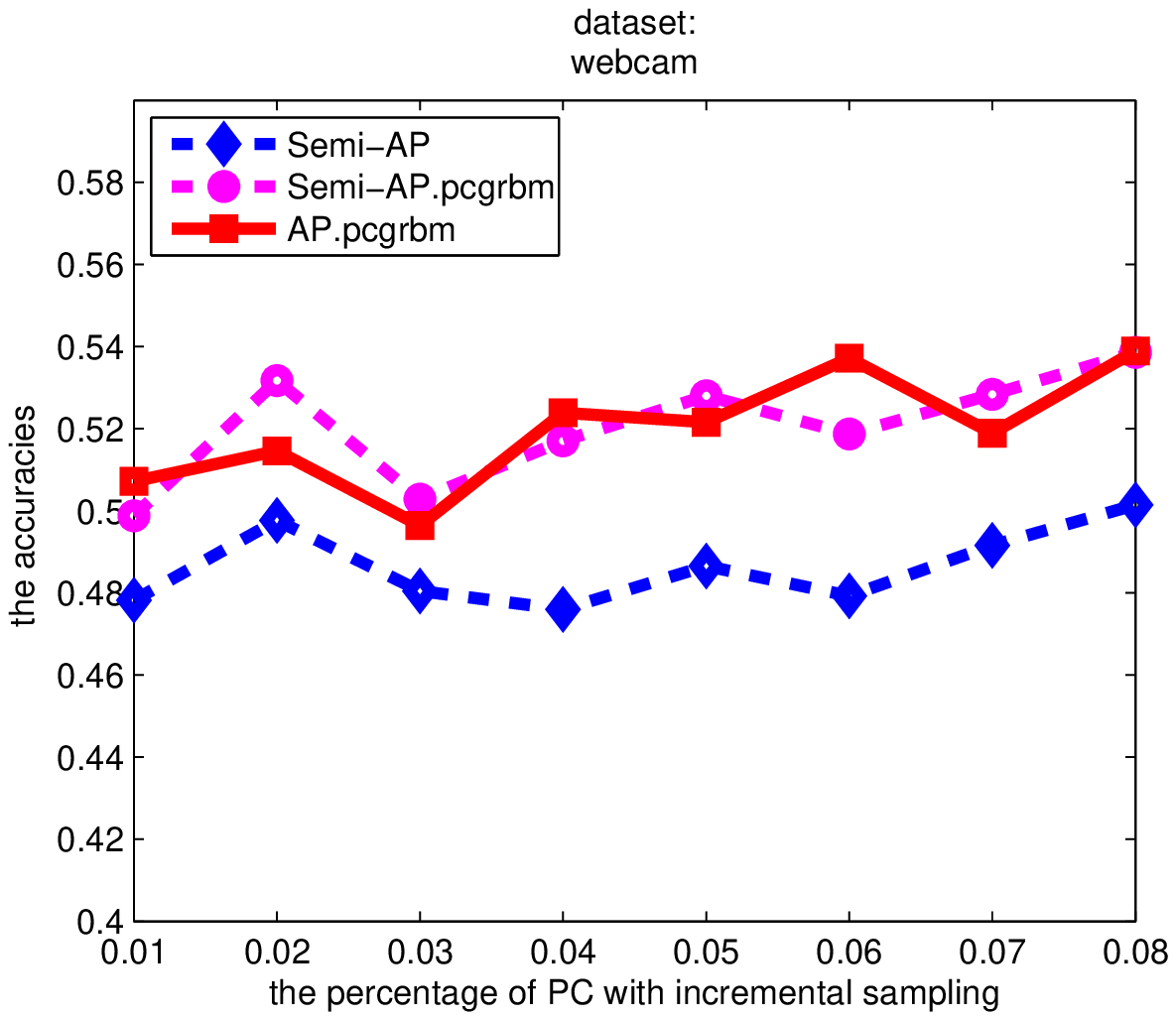}
    \includegraphics[scale=0.4225]{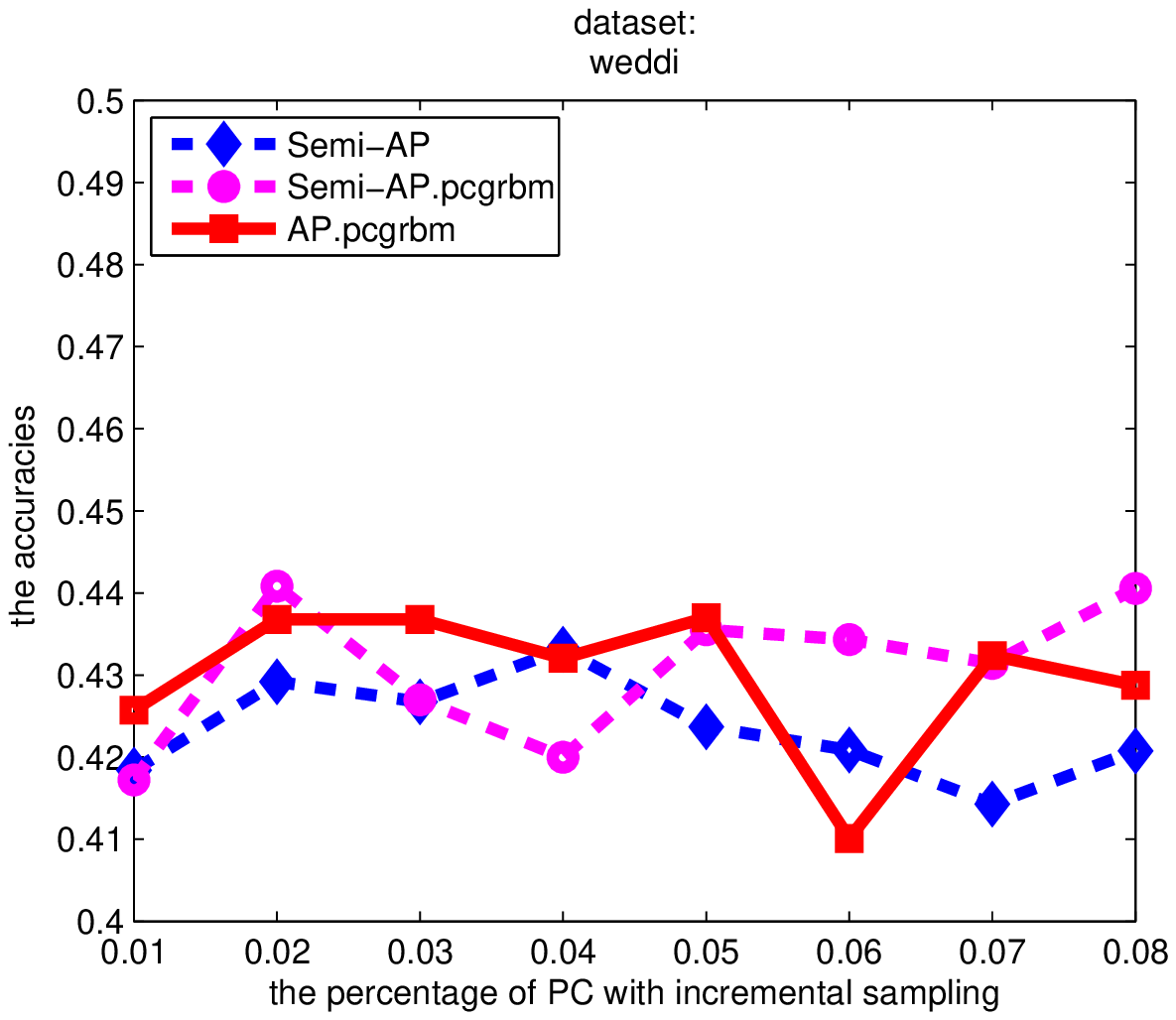}
    \includegraphics[scale=0.4225]{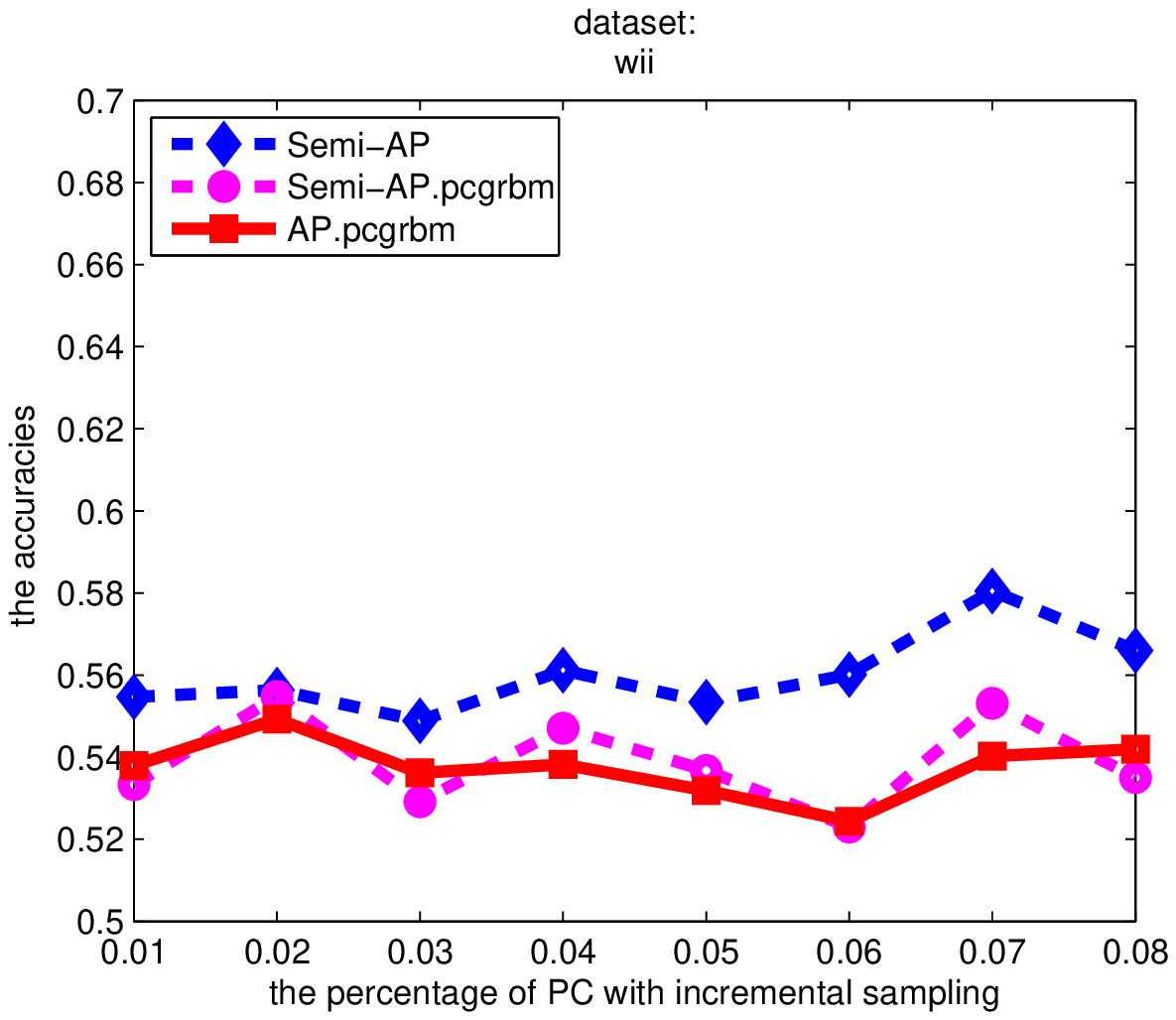}
    \includegraphics[scale=0.4225]{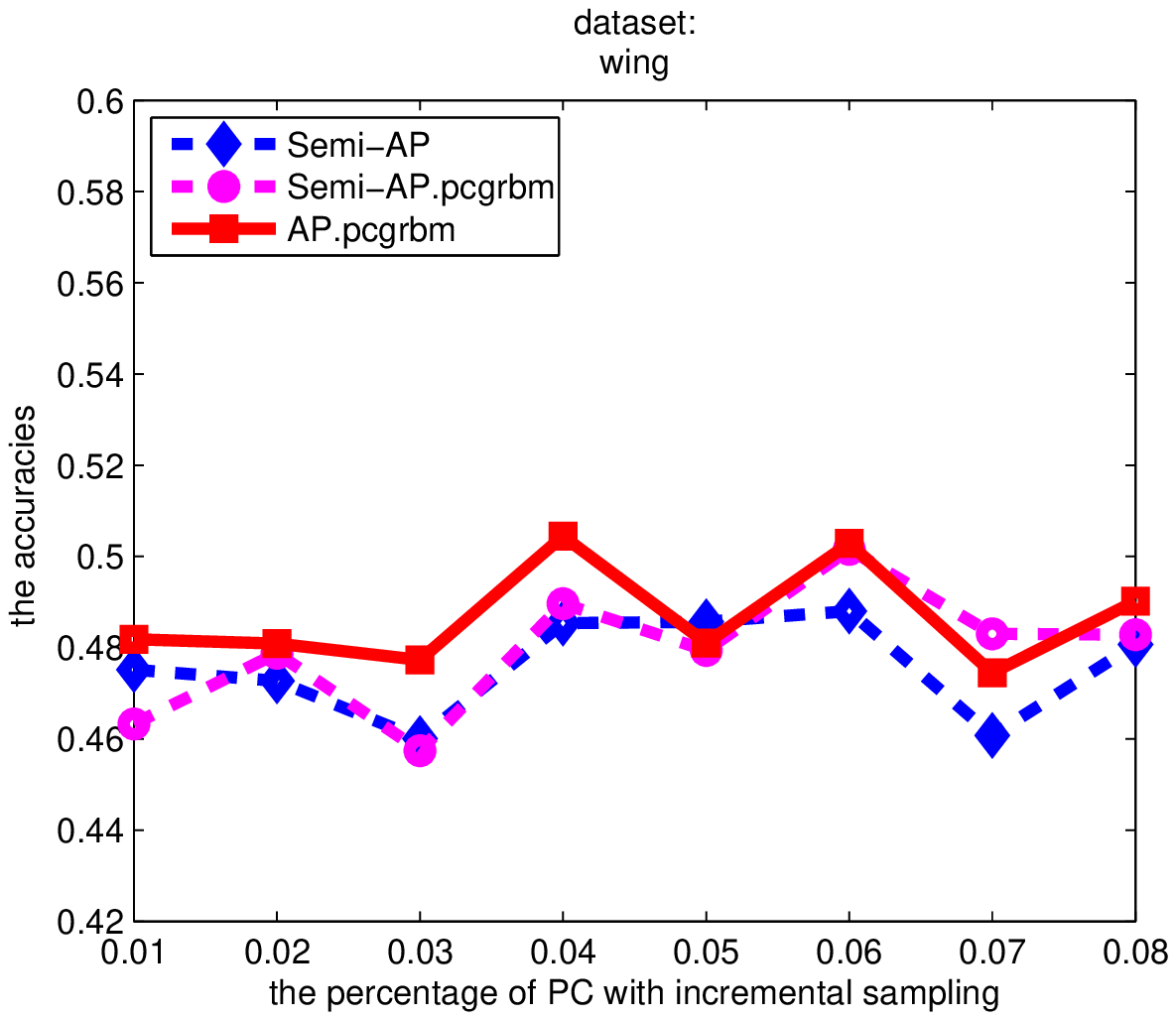}
\\
\caption{Semi-AP, Semi-AP.pcgrbm and AP.pcgrbm results with an increasing percentage of pairwise constraints (PC) from 1\% to 8\% in steps of 1\% by the incremental sampling method.
} \label{fig:1}
\end{figure*}

\begin{figure*}
\vspace{1mm} \centering
    \includegraphics[scale=0.4225]{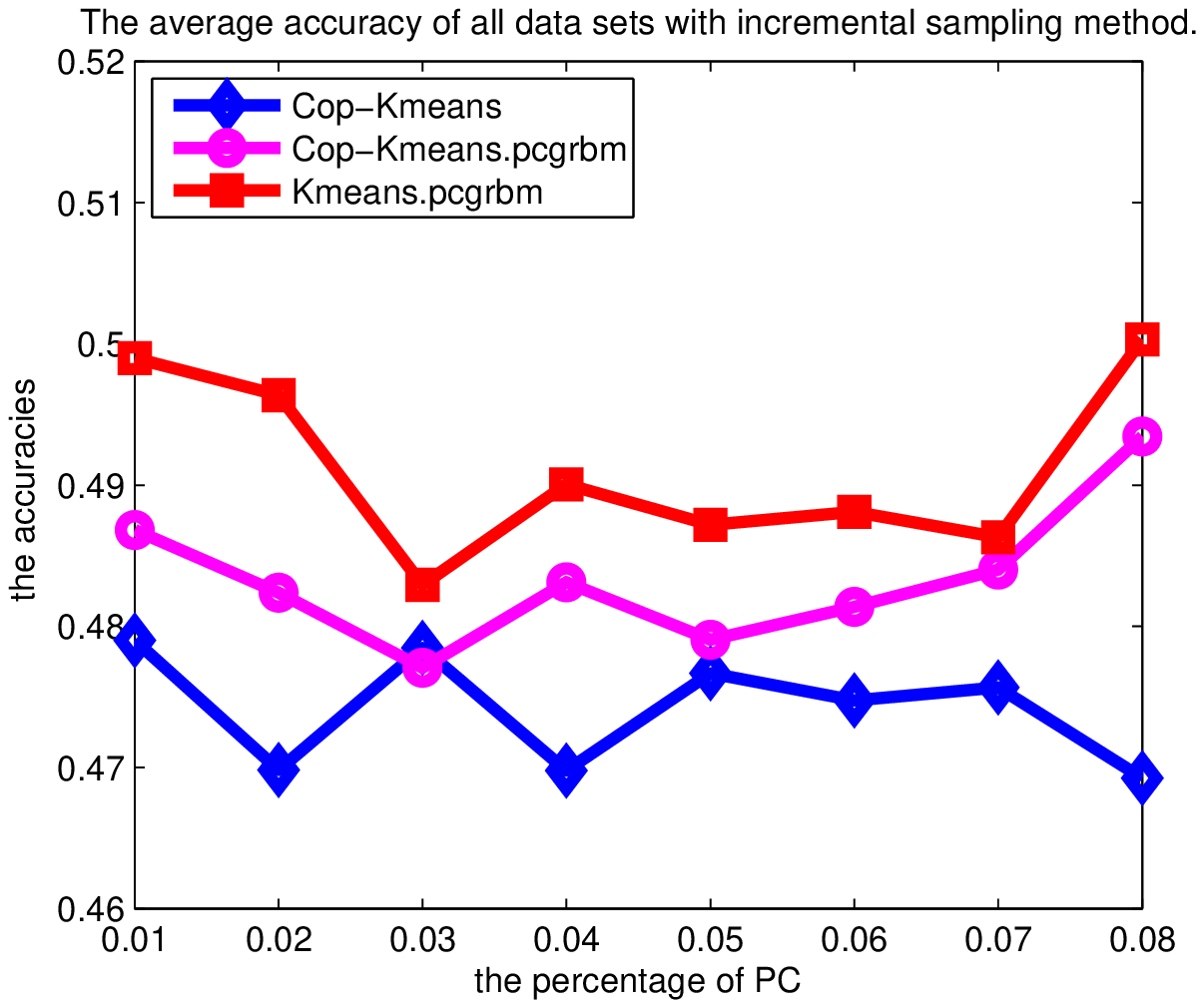}
    \includegraphics[scale=0.4225]{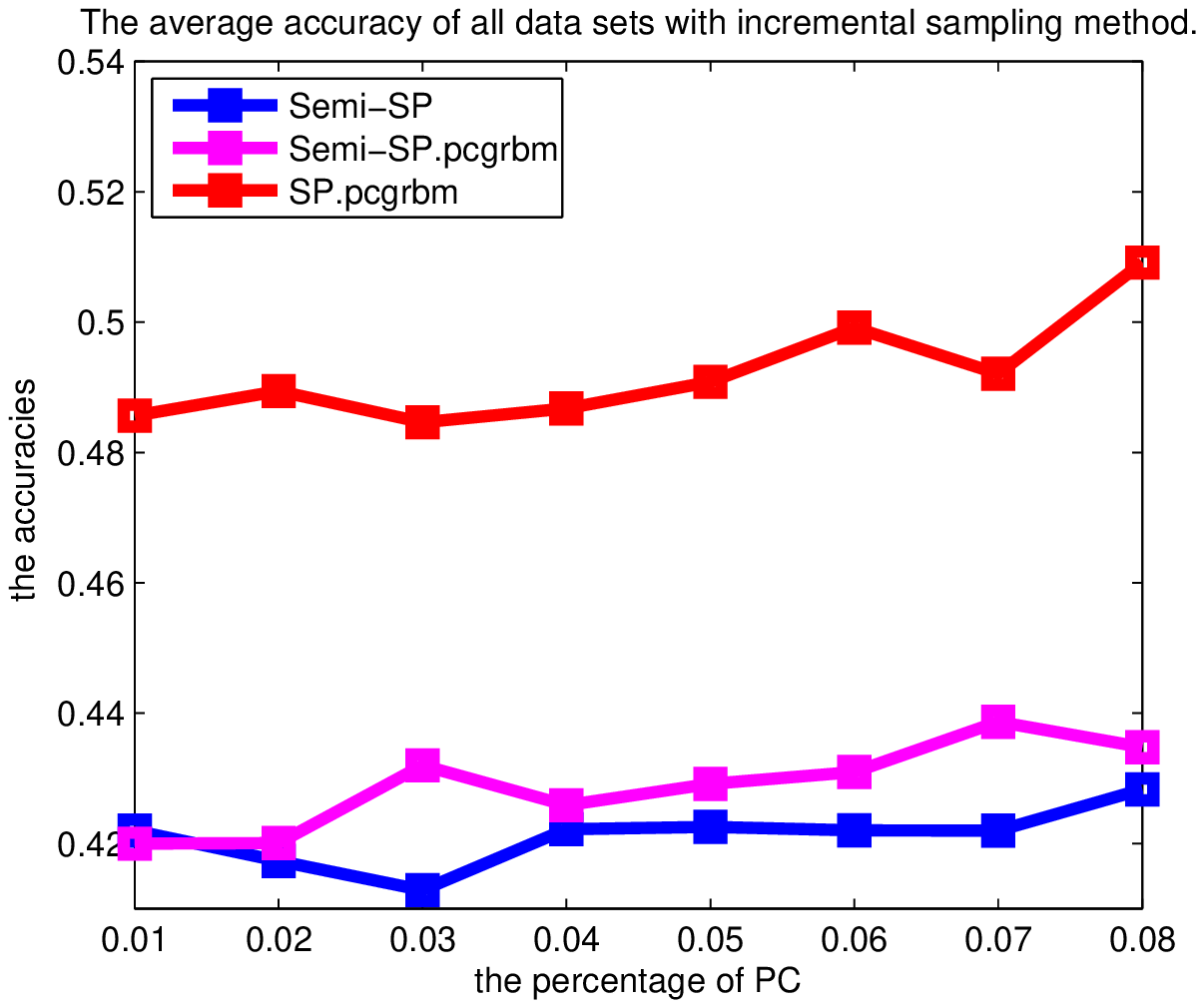}
    \includegraphics[scale=0.4225]{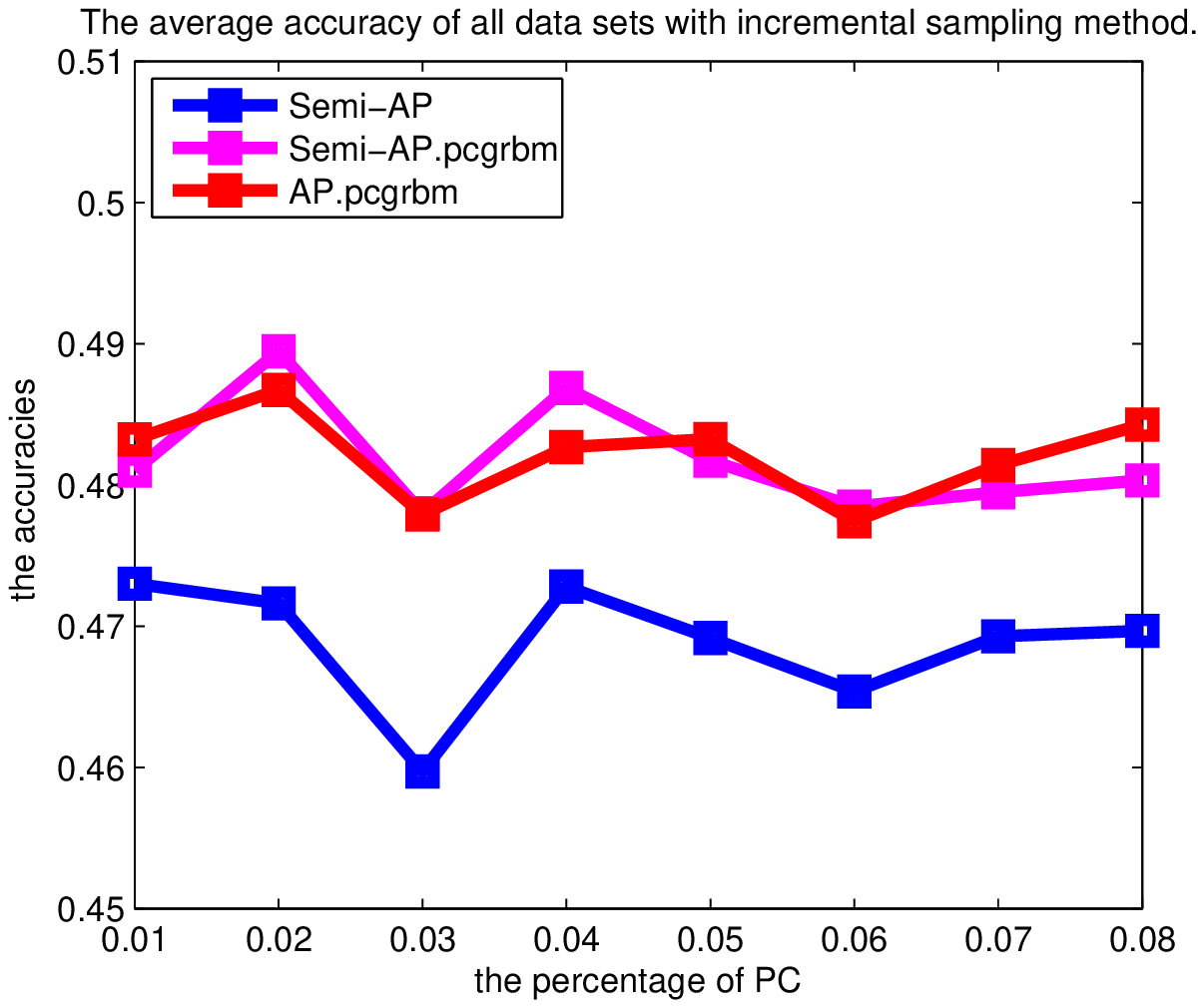}
\\
\caption{The average accuracy of all data sets of  Cop-Kmeans, Semi-SP, Semi-AP, Cop-Kmeans.pcgrbm, Semi-SP.pcgrbm, Semi-AP.pcgrbm, Kmeans.pcgbrm, SP.pcgrbm and AP.pcgrbm algorithms with an increasing percentage of pairwise constraints (PC) from 1\% to 8\% in steps of 1\% by the incremental sampling method.
} \label{fig:1}
\end{figure*}
\subsubsection{The pcGRBM for Clustering VS Unsupervised Algorithms}
In this section, we compare unsupervised clustering of K-means, SP and AP with Kmeans.pcgrbm, SP.pcgrbm and AP.pcgrbm which are based on the pcGRBM by evaluation of accuracy, rank and purity. In Table II and Table IV, the average accuracies of K-means, SP and AP algorithms are 46.57\%, 43.18\% and 46.61\%, respectively, but the average accuracies of Kmeans.pcgrbm, SP.pcgrbm and AP.pcgrbm algorithms raise to 50.03\%, 50.91\% and 48.43\%, respectively. In Table III ant Table VI, the average purities of K-means, SP and AP algorithms are 0.7733, 0.7709 and 0.7705, respectively, but the average purities of Kmeans.pcgrbm, SP.pcgrbm and AP.pcgrbm algorithms raise to 0.7885, 0.8106 and 0.7866, respectively. A greater accuracy and purity indicates a better algorithm. As a whole, it is obvious that the performances of clustering algorithms based on the pcGRBM model are better than the original unsupervised clustering.\\
\subsubsection{The pcGRBM VS RBM with Gaussian Visible Units for Clustering}
The pcGRBM and RBM with Gaussian visible have ability to extract features, however, which one shows better performance for clustering tasks? In order to compare the representation capability between the pcGRBM and RBM without any guiding of pairwise constraints, we design a structure of clustering algorithm in which the features of RBM with Gaussian visible units are used as the input of unsupervised clustering. In our experiments, we use three clustering algorithms based on this structure which are termed as Kmeans.grbm, SP.grbm and AP.grbm algorithms to compare with Kmeans.pcgrbm, SP.pcgrbm and AP.pcgrbm, respectively. In Table II, the average accuracies of kmeans.grbm, SP.grbm and AP.grbm algorithms are 46.31\%, 42.77\% and 47.80\%, respectively, however, Kmeans.pcgrbm, SP.pcgrbm and AP.pcgrbm algorithms raise the average accuracies by 3.72\%, 8.14\% and 0.63\%, respectively. Table III shows the average purities of kmeans.grbm, SP.grbm and AP.grbm. The values are 0.7747, 0.8013 and 0.7718, respectively. As a whole, it is obvious that the pcGRBM is better than RBM for clustering from all above results.\\
\subsubsection{The pcGRBM for Clustering VS Semi-supervised Algorithms}
In this section, we make further comparison among semi-supervised clustering of Cop-kmeans, Semi-SP and Semi-AP with Kmeans.pcgrbm, SP.pcgrbm and AP.pcgrbm by evaluation of accuracy, rank and purity. In Table IV, the average accuracies of Cop-kmeans, Semi-SP and Semi-AP with Kmeans.pcgrbm algorithms are 46.93\%, 42.83\% and 46.97\%, respectively. But the average accuracies of Kmeans.pcgrbm, SP.pcgrbm and AP.pcgrbm algorithms raise to 50.03\%, 50.91\% and 48.43\%, respectively. In Table V, the average ranks of Cop-kmeans, Semi-SP and Semi-AP algorithms are 60.8333, 92.7500 and 62.0000, respectively, however, the average ranks of Kmeans.pcgrbm, SP.pcgrbm and AP.pcgrbm algorithms are 28.1667, 26.4167 and 45.4167, respectively. The smaller the rank value means the better the algorithm. The average purities of Cop-kmeans, Semi-SP and Semi-AP algorithms are shown in Table VI. Their values are 0.7826, 0.8018 and 0.7824, respectively. From all above results, the performances of algorithms based on the pcGRBM model for clustering are better than the semi-supervised clustering.\\
\indent More detailed and intuitive comparisons about accuracy of each data set between semi-supervised algorithms and clustering algorithms based on pcGRBM model are shown in Fig. 4, Fig. 5 and Fig. 6. We plot the experimental results with the increasing percentage of pairwise constraints which ranges from 1\% to 8\% in steps of 1\% for Cop-Kmeans and Kmeans.pcgrbm in Fig. 4, Semi-SP and SP.pcgrbm in Fig. 5 and Semi-AP and AP.pcgrbm in Fig. 6, respectively. For most of these data sets, the performance of algorithms based on the pcGRBM model are better than semi-supervised algorithms. We plot the average accuracy of all data sets of  Cop-Kmeans, Semi-SP, Semi-AP, Cop-Kmeans.pcgrbm, Semi-SP.pcgrbm and Semi-AP.pcgrbm algorithms with an increasing percentage of pairwise constraints (PC) from 1\% to 8\% in steps of 1\% by the incremental sampling method in Fig. 7. As a whole, the clustering algorithms based on pcGRBM model perform better than semi-supervised algorithms.
\subsubsection{The Features of the pcGRBM for Semi-supervised Clustering}
In Table IV, we can see that the average accuracies of Cop-Kmeans.pcgrbm, Semi-SP.pcgrbm and Semi-AP.pcgrbm are 0.4935, 0.4348 and 0.4804, respectively. As show in 4, Semi-SP.pcgrbm algorithm is better than Semi-SP algorithm, but worse than SP.pcgrbm. Similarly, Semi-AP.pcgrbm algorithm is better than Semi-AP algorithm, but worse than AP.pcgrbm. In Table VI, the purity of Cop-Kmeans.pcgrbm, Semi-SP.pcgrbm and Semi-AP.pcgrbm algorithms are 0.7866, 0.8025 and 0.7835, respectively. They are worse than Kmeans.pcgrbm, SP.pcgrbm and AP.pcgrbm, respectively. The average rank of Cop-Kmeans.pcgrbm, Semi-SP.pcgrbm and Semi-AP.pcgrbm algorithms are 36.1667, 89.0833 and 49.6667. We plot the average accuracy of all data sets of Cop-Kmeans.pcgrbm, Semi-SP.pcgrbm and Semi-AP.pcgrbm algorithms with an increasing percentage of pairwise constraints (PC) from 1\% to 8\% in steps of 1\% by the incremental sampling method in Fig. 7. As a whole, the Cop-Kmeans.pcgrbm, Semi-SP.pcgrbm and Semi-AP.pcgrbm algorithms are better than Cop-kmeans, Semi-SP and Semi-AP algorithm, respectively, but worse than Kmeans.pcgrbm, SP.pcgrbm and AP.pcgrbm algorithms, respectively. Therefore, pairwise constraints may influence our pcGRBM model both positively and negatively since the constraints are ``soft''. Meanwhile, the same pairwise constraints also may influence semi-supervised clustering algorithms, both positively and negatively.\\
\subsubsection{The Rank and Friedman Aligned Ranks Test}
Nine algorithms are compared with twelve data sets by means of the Friedman Aligned test. We check whether the measured sum of the ranks is significantly different from the average value of the total ranks $\widehat{R}_j=654$ expected under the null hypothesis:$\sum_{j=1}^k\widehat{R}^2_{.,j}=4523296, \sum_{j=1}^k\widehat{R}^2_{i,.}=2908948$ and $T=52.5741$. \emph{T} is the chi-square distribution with 8 degrees of freedom because we use nine algorithms and twelve data sets. For one tailed test, the \emph{p}-value is 0.000000002 which is computed by $\chi^2(8)$ distribution and the \emph{p}-value is 0.000000004 for two-tailed test. Then, the null hypothesis is rejected at a high level significance. The experimental results of algorithms are significantly different because the \emph{p}-values are far less than 0.05.\\
\subsubsection{Time Complexity}
The time complexity of the proposed pcGRBM model is $O(TNM)$, where $T$ is the number of iterations, $N$ is the number of visible units and $M$ is the number of hidden units. As we know, the time complexities of K-means, SP and AP are $O(nkt)$, $O(n^3)$ and $O(n^3)$, respectively. $n$ is the number of instance, $k$ is the number of the cluster and $t$ is the number of iterations of K-means algorithm. We use the same hidden features of our pcGRBM model as the input of K-means.pcgrbm, SP.pcgrbm and AP.pcgrm algorithms. So, their time complexities are $O(TNM+nkt)$, $O(TNM+n^3)$ and $O(TNM+n^3)$, respectively.\\

\section{Conclusions}
 In this paper, we proposed a novel pcGRBM model. Its learning procedure is guided by the pairwise constraints and its encoding process is conducted under guidance. Then, some pairwise hidden features of pcGRBM flock together and another part of them are separated by the guidances. In the process of learning pcGRBM, CD learning is used to approximate ML learning and pairwise constraints are iterated transitions between visible and hidden units. Then, the background of pairwise constraints are encoded in hidden layer features of pcGRBM. In order to testify the availability of pcGRBM, the features of the hidden layer of the pcGRBM are used as input `data' for clustering tasks. The experimental results showed that the performance of the Kmeans.pcgrbm, SP.pcgrbm and AP.pcgrbm algorithms which are based on pcGRBM for clustering tasks are better than their classic unsupervised clustering algorithms (K-means, SP, AP, Kmeans.grbm, SP.grbm and AP.grbm), semi-supervised clustering algorithms (Cop-kmeans, Semi-SP, Semi-AP) and even better than semi-supervised clustering based on pcGRBM model (Cop-Kmeans.pcgrbm, Semi-SP.pcgrbm and Semi-AP.pcgrbm). We use 10-fold cross-validation strategy to train and test pcGRBM model. From the results we obtain that the incremental sampling shows excellent performance in the process of generating pairwise constraints.\\
 \indent There are several interesting questions in our future studies. For example, how to design deep networks based on the pcGRBM. How to strengthen pairwise constraints information when the layer of the deep network becomes deeper and deeper. How many dimensions in the hidden layer can enhance the performance for clustering. How to compare the performance with other semi-supervised feature extraction methods which are not based on RBM.
\section{Acknowledgement}
This work was partially supported by the National Science Foundation of China (Nos. 61773324, 61573292).
\section{Appendix}
In the following, the accuracy and purity results of unsupervised algorithms are shown in Table II and Table III, respectively. The accuracy, purity and rank results of clustering algorithms with pairwise constraints (8\%) are shown in Table IV, Table V and Table VI, respectively. The results of clustering algorithms with pairwise constraints (from 1\% to 7\%) are shown in the supplementary materials.
\begin{table*}
\begin{center}
\caption{The accuracies and variances of Kmeans, SP, AP, Kmeans.grbm, SP.grbm and AP.grbm algorithms (without pairwise constraints).}
\label{tab:results1} \scalebox{0.85}{
\begin{tabular}{|l|c|c|c|c|c|c|}
\hline
\textsf{ \bf{Dataset}} &  {K-means} & {SP}  & {AP}&{Kmeans.grbm}& {SP.grbm}& {AP.grbm} \\
\hline
\hline \textsf{alph} & {0.4434$\pm$0.00022 } & {0.4141$\pm$0.00002 } & {0.4467$\pm$ 0.00007}& {0.4587$\pm$0.00005} & {0.4069$\pm$0.00001} & {0.4486$\pm$0.00003} \\
\hline \textsf{alphabet} & {0.4370$\pm$0.00010 }& {0.4322$\pm$0.00006} & {0.4312$\pm$ 0.00018}& {0.4433$\pm$0.00012}& {0.4203$\pm$0.00011} & {0.4400$\pm$0.00005} \\
\hline \textsf{aquarium} & {0.4631$\pm$0.00036 }& {0.4162$\pm$0.00001}& {0.4527$\pm$ 0.00006}  & {0.4617$\pm$0.00011} & {0.4443$\pm$0.00027} & {0.4738$\pm$0.00025} \\
\hline \textsf{bed} & {0.4781$\pm$0.00007 }& {0.4568$\pm$0.00005} & {0.4631$\pm$ 0.00010}  & {0.4866$\pm$0.00061} & {0.4253$\pm$0.00011} & {0.4824$\pm$0.00016} \\
\hline \textsf{beer} & {0.4502$\pm$0.00039 }& {0.4486$\pm$0.00002} & {0.4402$\pm$ 0.00010} & {0.4658$\pm$0.00035} & {0.4151$\pm$0.00006} & {0.4480$\pm$0.00009} \\
\hline \textsf{beverage} & {0.4717$\pm$0.00012 }& {0.4362$\pm$0.00002}& {0.4425$\pm$ 0.00005}  & {0.4769$\pm$0.00009} & {0.4272$\pm$0.00003} & {0.4636$\pm$0.00005} \\
\hline \textsf{breakfast} & {0.5099$\pm$0.00021 }& {0.4840$\pm$0.00005}& {0.4868$\pm$ 0.00003}  &{0.5167$\pm$0.00019} & {0.4158$\pm$0.00021} & {0.4917$\pm$0.00013} \\
\hline \textsf{virus} & {0.4377$\pm$0.00008 }& {0.4338$\pm$0.00005}& {0.4371$\pm$ 0.00004}  &   {0.4367$\pm$0.00009} & {0.4182$\pm$0.00012} & {0.4437$\pm$0.00003} \\
\hline \textsf{webcam} & {0.4607$\pm$0.00009 }& {0.4601$\pm$0.00003} & {0.4764$\pm$ 0.00009}  &   {0.4768$\pm$0.00035} & {0.4092$\pm$0.00007} & {0.4986$\pm$0.00014} \\
\hline \textsf{weddi} & {0.4284$\pm$0.00031 }& {0.4004$\pm$0.00005}& {0.4134$\pm$0.00004}  &   {0.4397$\pm$0.00012} & {0.4441$\pm$0.00013} & {0.4180$\pm$0.00006} \\
\hline \textsf{wii} & {0.5648$\pm$0.00049 }& {0.5576$\pm$0.00002}& {0.5501$\pm$0.00009}   &   {0.5762$\pm$0.00028} & {0.5634$\pm$0.00017} & {0.5541$\pm$0.00008} \\
\hline \textsf{wing} & {0.4657$\pm$0.00087 }& {0.4318$\pm$0.00004}& {0.4661$\pm$ 0.00012}  &   {0.4631$\pm$0.00030} & {0.4277$\pm$0.00017} & {0.4780$\pm$0.00009} \\
\hline
\hline  {Average} & {0.4657}  & {0.4318}& {0.4661} &  {0.4631} & {0.4277} & {0.4780}\\
\hline
\end{tabular}}
\end{center}
\end{table*}

\begin{table*}
\begin{center}
\caption{The purities of K-means, SP, AP, Kmeans.grbm, SP.grbm and AP.grbm algorithms (without pairwise constraints).}
\label{tab:results2} \scalebox{0.85}{
\begin{tabular}{|l|c|c|c|c|c|c|}
\hline
\textsf{ \bf{Dataset}} &  {K-means} & {SP}  & {AP}&{Kmeans.grbm}& {SP.grbm}& {AP.grbm} \\
\hline
\hline \textsf{alph}  & {0.8014} & {0.7977}  & {0.7944} & {0.8015 } & {0.8243}& {0.7965 } \\
\hline \textsf{alphabet}  & {0.7983} & {0.7764}  & {0.7941} &   {0.7958 } &    {0.8147}& {0.7955}\\
\hline \textsf{aquarium}  & {0.6473} & {0.6389}  & {0.6514}&  {0.6438} &    {0.6695}&{0.6542} \\
\hline \textsf{bed} & {0.7091} & {0.7063}  & {0.7271}&  {0.7120} &   {0.7301}&{0.7211} \\
\hline \textsf{beer} & {0.8284} & {0.8318}  & {0.8272} &   {0.8321} &    {0.8606}&  {0.8241}\\
\hline \textsf{beverage} & {0.7281} & {0.7184}  & {0.7237}&   {0.7310} &    {0.7531} &  {0.7208} \\
\hline \textsf{breakfast}  & {0.7409} & {0.7594}  & {0.7544}&  {0.7459} &    {0.8124} &  {0.7555} \\
\hline \textsf{virus} & {0.9001}  & {0.9112}  & {0.8824}&   {0.9026} &    {0.9163} &  {0.8913}\\
\hline \textsf{webcam} & {0.8555}  & {0.8434}  & {0.8171} &   {0.8576} &    {0.9100} &  {0.8115} \\
\hline \textsf{weddi}  & {0.8362} & {0.8423}  & {0.8316} &   {0.8375} &    {0.8429} &  {0.8369} \\
\hline \textsf{wii}  & {0.6951} & {0.6954}  & {0.7072}&  {0.6953} &    {0.7269} &  {0.7132} \\
\hline \textsf{wing}  & {0.7397} & {0.7299}  & {0.7359}&   {0.7415} &    {0.7553} &  {0.7408}\\
\hline {Average} & {0.7733}  & {0.7709} & {0.7705}&    {0.7747} &  {0.8013}&  { 0.7718}\\
\hline
\end{tabular}}
\end{center}
\end{table*}

\begin{table*}
\begin{center}
\caption{The accuracies and variances of Cop-Kmeans, Semi-SP, Semi-AP, Cop-Kmeans.pcgrbm, Semi-SP.pcgrbm, Semi-AP.pcgrbm, Kmeans.pcgbrm, SP.pcgrbm and AP.pcgrbm algorithms withs pairwise constraints (8\%).}
\label{tab:results1} \scalebox{0.8}{
\begin{tabular}{|l|c|c|c|c|c|c|c|c|c|c|c|c|}
\hline
\textsf{ \bf{Dataset}} &  {Cop-Kmeans}& {Semi-SP}& {Semi-AP}& { Cop-Kmeans.pcgrbm }& {Semi-SP.pcgrbm} & {Semi-AP.pcgrbm} &{Kmeans.pcgrbm}& {SP.pcgrbm}& {AP.pcgrbm}\\
\hline
\hline \textsf{alph}& {0.4407$\pm$0.00159 } & {0.4142$\pm$0.00072 } & {\textbf{0.4673$\pm$ 0.00137}} & {0.4326$\pm$0.00134} &    {0.4356$\pm$0.00094} & {0.4506$\pm$ 0.00294} & {0.4497$\pm$0.00168} &    {0.4594$\pm$0.00100} & {0.4616$\pm$0.00292} \\

\hline \textsf{alphabet} & {\textbf{0.4601$\pm$0.00146} }& {0.4405$\pm$0.00079} & {0.4357$\pm$ 0.00233} & {0.4515$\pm$0.00180} &    {0.4537$\pm$0.00118} & {0.4534$\pm$0.00265} &   {0.4509$\pm$0.00266} & {0.4551$\pm$0.00071} & {0.4456$\pm$0.00297} \\

\hline \textsf{aquarium} & {0.4331$\pm$0.00225 }& {0.4369$\pm$0.00068}& {0.4484$\pm$ 0.00140} & {0.5448$\pm$0.00649} &    {0.4444$\pm$0.00243}  & {0.5143$\pm$0.00736} &   {0.5300$\pm$0.00708} & {\textbf{0.5494$\pm$0.01030}} & {0.4830$\pm$0.00402} \\

\hline \textsf{bed} & {0.4764$\pm$0.00781 }& {0.4263$\pm$0.00439} & {0.4641$\pm$ 0.00160} & {0.5114$\pm$0.00543} &    {0.4495$\pm$0.00322} & {0.4475$\pm$0.00415} &   {0.4853$\pm$0.00387} & {\textbf{0.5952$\pm$0.00740}} & {0.4748$\pm$0.00554} \\

\hline \textsf{beer} & {0.4531$\pm$0.00234 }& {0.4350$\pm$0.00037} & {0.4511$\pm$ 0.00092} & {0.5019$\pm$0.00306} &    {0.4254$\pm$0.00125} & {0.4666$\pm$0.00125} &   {\textbf{0.5238$\pm$0.00074}} & {0.5127$\pm$0.00620} & {0.4778$\pm$0.00082} \\

\hline \textsf{beverage} & {0.4778$\pm$0.00299 }& {0.4479$\pm$0.00429}& {0.4460$\pm$ 0.00129} & {0.4797$\pm$0.00314} &    {0.4337$\pm$0.00055}  & {0.4671$\pm$0.00234} &   {0.5101$\pm$0.00423} & {\textbf{0.5352$\pm$0.00465}} & {0.4585$\pm$0.00138} \\

\hline \textsf{breakfast} & {0.5039$\pm$0.00286 }& {0.4351$\pm$0.00287}& {0.5066$\pm$ 0.00344} & {0.5282$\pm$0.00451} &    {0.4409$\pm$0.00168}  & {0.4968$\pm$0.00221} &   {\textbf{0.5536$\pm$0.00726}} & {0.5168$\pm$0.00576} & {0.5292$\pm$0.00510} \\

\hline \textsf{virus} & {0.4276$\pm$0.00222 }& {0.4124$\pm$0.00043}& {0.4481$\pm$ 0.00121} & {0.4496$\pm$0.00125} &    {0.4081$\pm$0.00064}  & {0.4708$\pm$0.00295} &   {0.4749$\pm$0.00237} & {0.4447$\pm$0.00112} & {\textbf{0.4811$\pm$0.00262}} \\

\hline \textsf{webcam} & {0.4632$\pm$0.00526 }& {0.4157$\pm$0.00153} & {0.5014$\pm$ 0.00455} & {0.4998$\pm$0.00154} &    {0.4136$\pm$0.00182} & {0.5386$\pm$0.00332} &   {0.5096$\pm$0.00322} & {0.4248$\pm$0.00112} & {\textbf{0.5389$\pm$0.00376}} \\

\hline \textsf{weddi} & {0.4654$\pm$0.00182 }& {0.4258$\pm$0.00219}& {0.4208$\pm$0.00110}  & {0.4311$\pm$0.00154} &   {0.4200$\pm$0.00100}  & {0.4405$\pm$0.00181} &   {0.4347$\pm$0.00093} & {\textbf{0.4746$\pm$0.00102}} & {0.4288$\pm$0.00221} \\

\hline \textsf{wii} & {0.5526$\pm$0.00316 }& {0.4028$\pm$0.00040}& {0.5660$\pm$0.00239}  & {0.5765$\pm$0.00309} &   {0.4213$\pm$0.00067}  & {0.5350$\pm$0.00100} &   {0.5521$\pm$0.00167} & {\textbf{0.5925$\pm$0.00690}} & {0.5419$\pm$0.00089} \\

\hline \textsf{wing} & {0.4775$\pm$0.00532 }& {0.4472$\pm$0.0.00130}& {0.4807$\pm$ 0.00518} & {0.5142$\pm$0.00341} &    {0.4711$\pm$0.00519}  & {0.4829$\pm$0.00697} &   {0.5292$\pm$0.00446} & {\textbf{0.5492$\pm$0.00061}} & {0.4902$\pm$0.00520} \\
\hline
\hline  {Average} & {0.4693}  & {0.4283}& {0.4697} &   {0.4935} &    {0.4348} &   {0.4804} &    {0.5003} & {0.5091} &  {0.4843}\\
\hline
\end{tabular}}
\end{center}
\end{table*}
\begin{table*}
\begin{center}
\caption{The ranks of Cop-Kmeans, Semi-SP, Semi-AP, Cop-Kmeans.pcgrbm, Semi-SP.pcgrbm, Semi-AP.pcgrbm, Kmeans.pcgbrm, SP.pcgrbm and AP.pcgrbm algorithms with pairwise constraints (8\%).}
\label{tab:results2} \scalebox{0.8}{
\begin{tabular}{|l|c|c|c|c|c|c|c|c|c|c|c|c|c|}
\hline
\textsf{ \bf{Dataset}} & {Cop-Kmeans}& {Semi-SP}& {Semi-AP}& { Cop-Kmeans.pcgrbm }& {Semi-SP.pcgrbm} & {Semi-AP.pcgrbm} &{Kmeans.pcgrbm}& {SP.pcgrbm}& {AP.pcgrbm}&{Total}\\
\hline
\hline \textsf{alph}  & {-0.0051(63)} & {-0.0316(88)}  & {0.0215(31)} &   {-0.0131(74)}&   {-0.0101(70)} &    {0.0048(45)}& {0.0039(48) } &    {0.0137(37)}& {0.0159(34) }&{490} \\
\hline \textsf{alphabet}  & {0.0105(38)} & {-0.0091(68)}  & {-0.0139(75)} &   {0.0018(53)}&   {0.0041(47)} &    {0.0037(49)}& {0.0013(55) } &    {0.0055(42)}& {-0.0040(59)} & {486}\\
\hline \textsf{aquarium}  & {-0.0540(101)} & {-0.0503(99)}  & {-0.0387(94)}&   {0.0577(7)} &    {-0.0427(96)}& {0.0271(25) } &   {0.0429(12)} &    {0.0623(4)}&{-0.0042(60)} & {498}\\
\hline \textsf{bed} & {-0.0048(62)} & {-0.0549(102)}  & {-0.0171(79)}&   {0.0303(20)} &    {-0.0317(89)}& {-0.0337(90) } &   {0.0042(46)} &    {0.1141(1)}&  {-0.0064(66)}& {555}\\
\hline \textsf{beer} & {-0.0189(83)} & {-0.0369(92)}  & {-0.0208(84)}&   {0.0300(21)} &    {-0.0465(98)}& {-0.0054(64) } &   {0.0519(10)} &    {0.0408(13)} &  {0.0058(41)} & { 506}\\
\hline \textsf{beverage}  & {0.0049(44)} & {-0.0250(86)}  & {0.0268(87)}&   {0.0068(40)} &    {-0.0391(95)}& {-0.0058(65) } &   {0.0372(15)} &    {0.0623(3)} &  {-0.0144(76)} & {511}\\
\hline \textsf{breakfast} & {0.0027(51)}  & {-0.0662(106)}  & {0.0054(43)}&   {0.0270(26)} &    {-0.0604(103)}& {-0.0044(61) } &   {0.0524(9)} &    {0.0155(35)} &  {0.0280(23)}& {457}\\
\hline \textsf{virus} & {-0.0188(82)}  & {-0.0337(91)}  & {0.0017(54)}&   {0.0032(50)} &    {-0.0383(93)}& {0.0244(29) } &   {0.0285(22)} &    {-0.0017(56)} &  {0.0347(18)} & {495} \\
\hline \textsf{webcam}  & {-0.0152(77)} & {-0.0627(104)}  & {0.0230(30)}&   {0.0214(32)} &    {-0.0648(105)}& {0.0602(6) } &   {0.0312(19)} &    {-0.0536(100)} &  {0.0605(5)}&{478} \\
\hline \textsf{weddi}  & {0.0274(24)} & {-0.0122(72)}  & {-0.0172(80)}&   {-0.0068(67)} &    {-0.0179(81)}& {0.0026(52) } &   {-0.0032(57)} &    {0.0366(16)} &  {-0.0092(69)} &{518} \\
\hline \textsf{wii}  & {0.0258(27)} & {-0.1240(108)}  & {0.0392(14)}&   {0.0498(11)} &    {-0.1054(107)}& {0.0083(39) } &   {0.0253(28)} &    {0.0657(2)} &  {0.0152(36)}& {372}\\
\hline \textsf{wing} & {-0.0161(78)}  & {-0.0464(97)}  & {-0.0129(73)}&   {0.0206(33)} &    {-0.0225(85)}& {-0.0107(71) } &   {0.0356(17)} &    {0.0556(8)} &  {-0.0034(58)}&  {520} \\
\hline {Total} & {730}  & {1113} & {744} &  {434} & {1069}&  {596} &  {338} & {317}&{ 545} &{5886}\\
\hline {Average rank} & {60.8333}  & {92.7500} & {62.0000} &    {36.1667} &  {89.0833}& {49.6667}& {28.1667} &    {26.4167} &  {45.4167}&{}\\
\hline
\end{tabular}}
\end{center}
\end{table*}
\begin{table*}
\begin{center}
\caption{The purities of Cop-Kmeans, Semi-SP, Semi-AP, Cop-Kmeans.pcgrbm, Semi-SP.pcgrbm, Semi-AP.pcgrbm, Kmeans.pcgbrm, SP.pcgrbm and AP.pcgrbm algorithms with pairwise constraints (8\%).}
\label{tab:results2} \scalebox{0.8}{
\begin{tabular}{|l|c|c|c|c|c|c|c|c|c|c|c|c|}
\hline
\textsf{ \bf{Dataset}} &  {Cop-Kmeans}& {Semi-SP}& {Semi-AP}& { Cop-Kmeans.pcgrbm }& {Semi-SP.pcgrbm} & {Semi-AP.pcgrbm} &{Kmeans.pcgrbm}& {SP.pcgrbm}& {AP.pcgrbm}\\
\hline
\hline \textsf{alph}  & {0.7939} & {0.8123}  & {0.8018} &   {0.8186}&   {0.8243} &    {0.8126}& {\textbf{0.8250} } &    {0.8143}& {0.8109}\\
\hline \textsf{alphabet}  & {0.8050} & {\textbf{0.8224}}  & {0.8059}&   {0.8025} &    {0.8094}& {0.8090 } &   {0.8058} &    {0.8150}&{0.8064} \\
\hline \textsf{aquarium} & {0.6595} & {0.6560}  & {0.6630}&   {0.6576} &    {0.6603}& {0.6688 } &   {0.6634} &    {\textbf{0.6897}}&{0.6718} \\
\hline \textsf{bed} & {0.7236} & {0.7126}  & {0.7341}&   {0.7363} &    {0.7145}& {0.7381 } &   {0.7396} &    {0.7413}&  {\textbf{0.7514}}\\
\hline \textsf{beer} & {0.8352} & {0.8479}  & {0.8457}&   {0.8450} &    {0.8539}& {0.8355 } &   {0.8370} &    {\textbf{0.8710}} &  {0.8328} \\
\hline \textsf{beverage}  & {0.7393} & {0.7334}  & {0.7417}&   {0.7376} &    {0.7420}& {0.7382 } &   {0.7396} &    {\textbf{0.7478}} &  {0.7412} \\
\hline \textsf{breakfast} & {0.7734}  & {0.8077}  & {0.7679}&   {0.7730} &    {0.8061}& {0.7670 } &   {0.7557} &    {\textbf{0.8248}} &  {0.7661}\\
\hline \textsf{virus} & {0.9030}  & {0.9128}  & {0.8886}&   {0.8852} &    {\textbf{0.9214}}& {0.8929 } &   {0.9039} &    {0.9088} &  {0.8959}\\
\hline \textsf{webcam}  & {0.8461} & {0.9027}  & {0.8272}&   {0.8578} &    {0.8928}& {0.8411} &   {0.8562} &    {\textbf{0.9248}} &  {0.8451} \\
\hline \textsf{weddi}  & {0.8458} & {0.8365}  & {0.8398}&   {0.8547} &    {0.8205}& {0.8445 } &   {0.8652} &    {\textbf{0.8669}} &  {0.8536} \\
\hline \textsf{wii}  & {0.7204} & {\textbf{0.8399}}  & {0.7249}&   {0.7225} &    {0.8383}& {0.7059 } &   {0.7210} &    {0.7633} &  {0.7208}\\
\hline \textsf{wing}  & {0.7463} & {0.7373}  & {0.7482} &   {0.7486}&   {0.7462} &    {0.7480}& {0.7494 } &    {\textbf{0.7597}}& {0.7431 }\\
\hline {Average} & {0.7826}  & {0.8018} & {0.7824} &    {0.7866} &  {0.8025}&  {0.7835} &    {0.7885} &  {0.8106}&  { 0.7866}\\
\hline
\end{tabular}}
\end{center}
\end{table*}

\bibliography{rbm}
\bibliographystyle{IEEEtran}
\begin{IEEEbiography}[{\includegraphics [width=1in,height=1in,keepaspectratio]{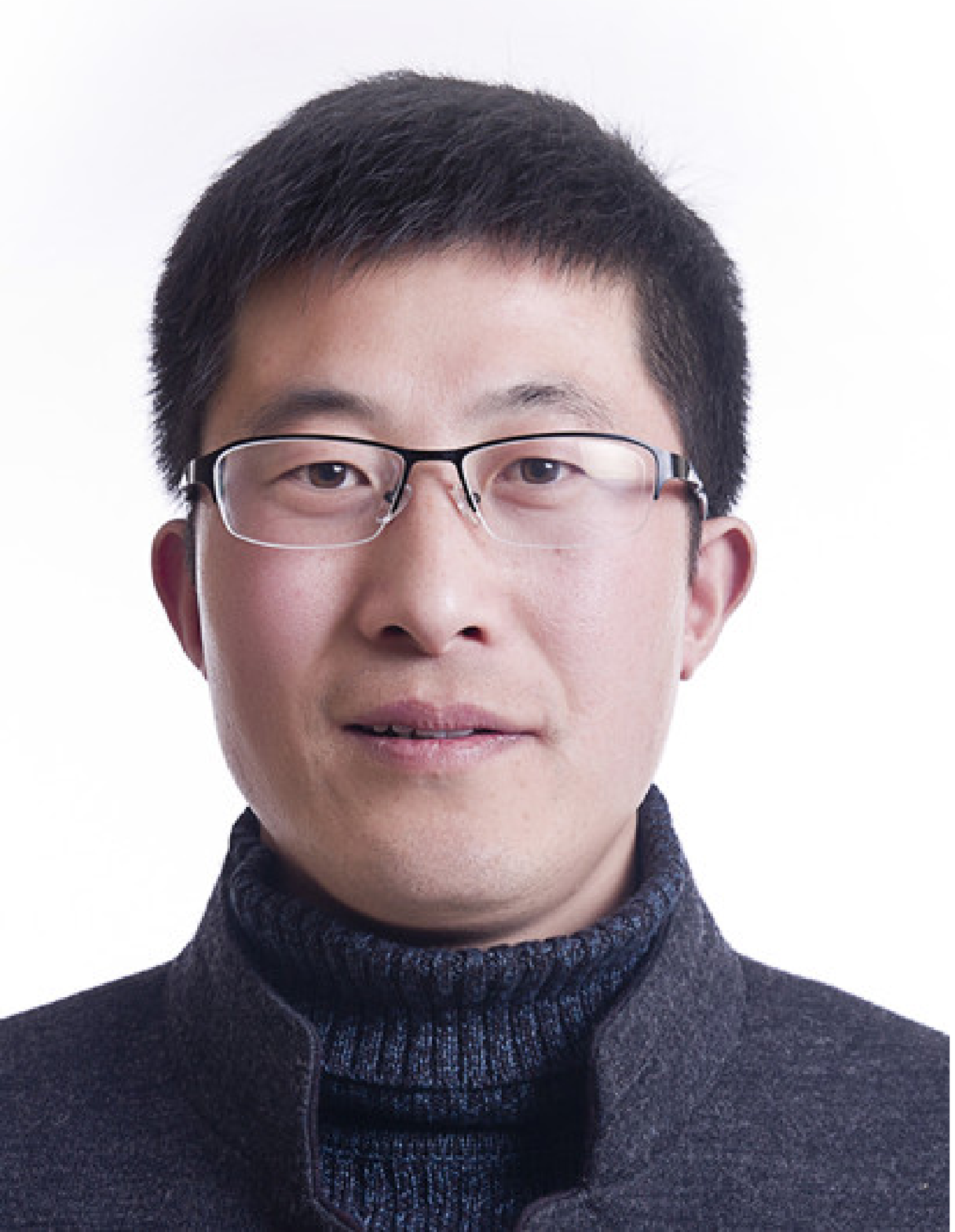}}]{Jielei Chu}
 received the B.S. degree from Southwest Jiaotong University, Chengdu, China in 2008, and is currently working toward the Ph.D. degree at Southwest Jiaotong University. His research interests are machine learning, data mining, semi-supervised learning and ensemble learning.
\end{IEEEbiography}
\begin{IEEEbiography}[{\includegraphics [width=1in,height=1in,clip,keepaspectratio]{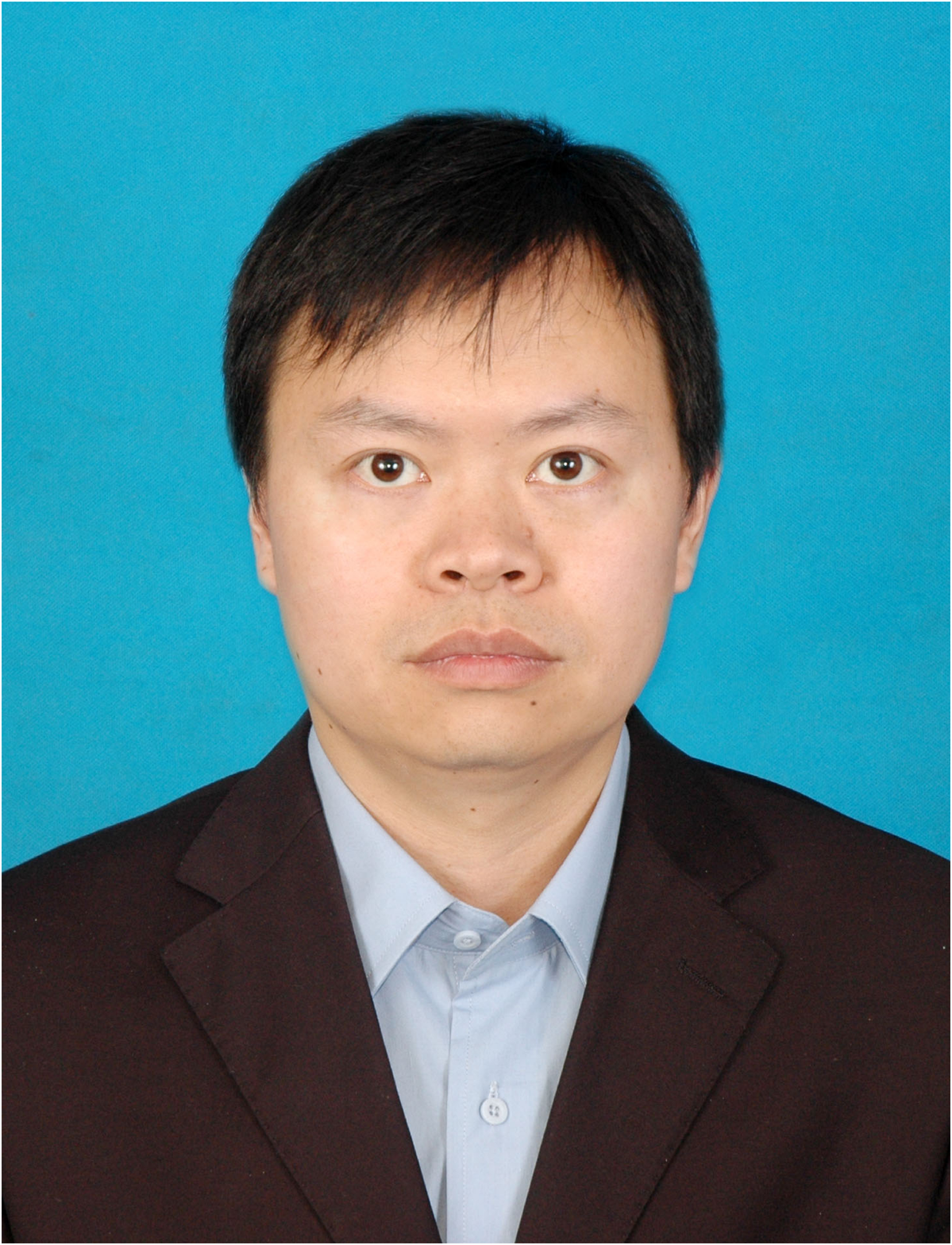}}]{Hongjun Wang}
 received his Ph.D. degree in computer science from Sichuan University of China in 2009. He is currently Associate Professor of the Key Lab of Cloud Computing and Intelligent Techniques in Southwest Jiaotong University. His research interests are machine learning, data mining and ensemble learning. He published over 30 research papers in journals and conferences and he is a member of ACM and CCF. He has been a reviewer for several academic journals.
\end{IEEEbiography}
\begin{IEEEbiography}[{\includegraphics [width=1in,height=1in,clip,keepaspectratio]{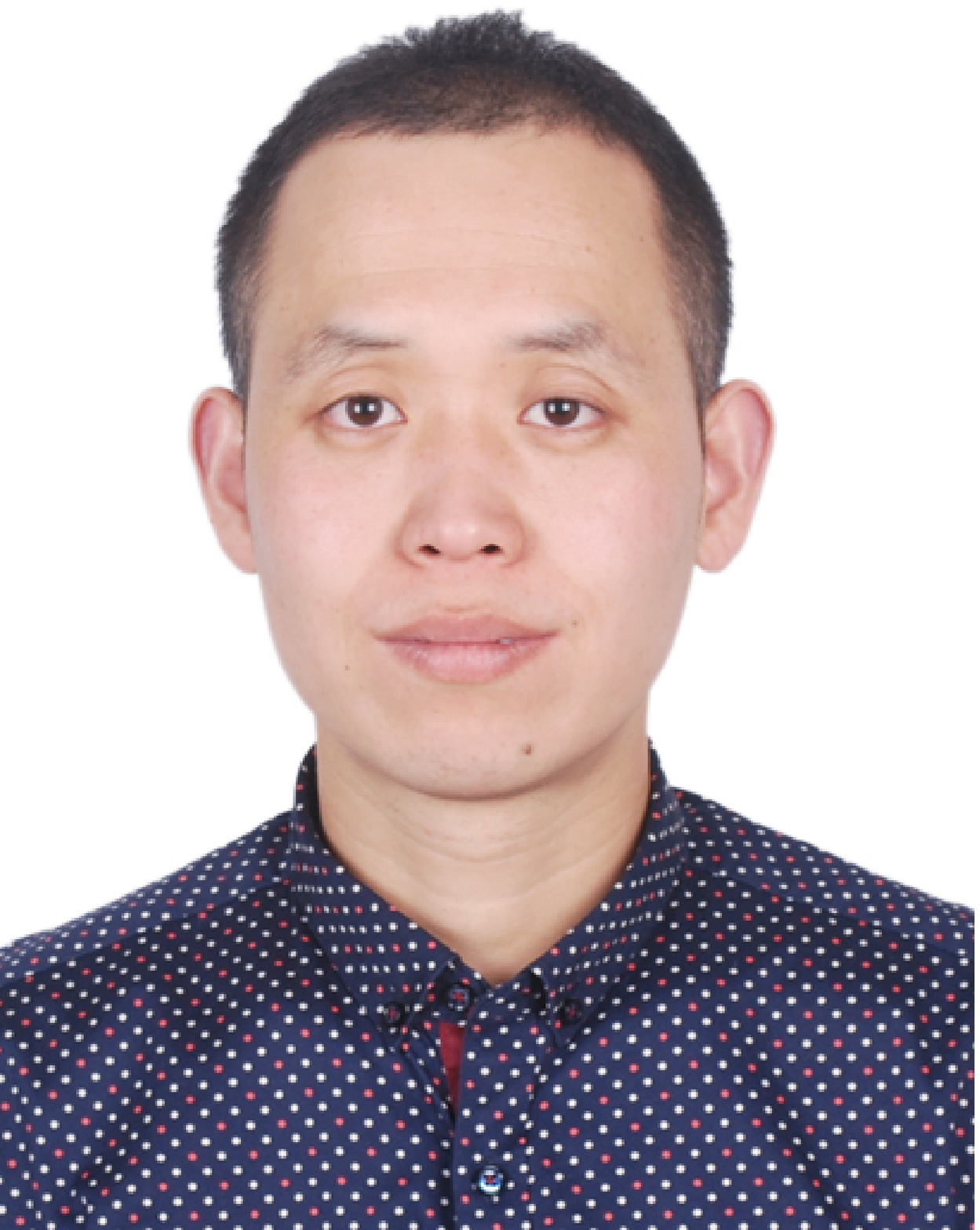}}]{Meng Hua}
 received his Ph.D. degree in Mathematics from Sichuan University of China in 2010. His research interests include belief revision, reasoning with uncertainty, machine learning, general topology.
\end{IEEEbiography}
\begin{IEEEbiography}[{\includegraphics [width=1in,height=1in,clip,keepaspectratio]{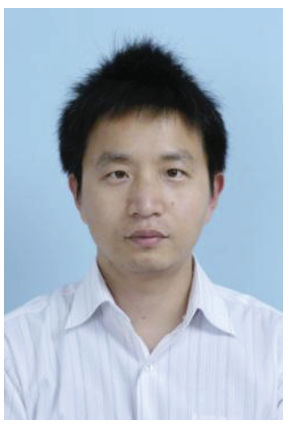}}]{Peng Jin}
received his BS, MS and Ph.D. in Computing Science from the Zhongyuan University of Technology, Nanjing University of Science and Technology, Peking University, respectively. From October 2007 to April 2008, he was a visiting student at the Department of Informatics, University of Sussex (Funded by China Scholarship Council); from August 2014 to February 2015, he is a visiting research fellow at the Department of Informatics, University of Sussex. Now, he is a professor of Leshan Normal University. His research interests include natural language processing, information retrieval and machine learning.
\end{IEEEbiography}
\begin{IEEEbiography}[{\includegraphics [width=1in,height=1in,clip,keepaspectratio]{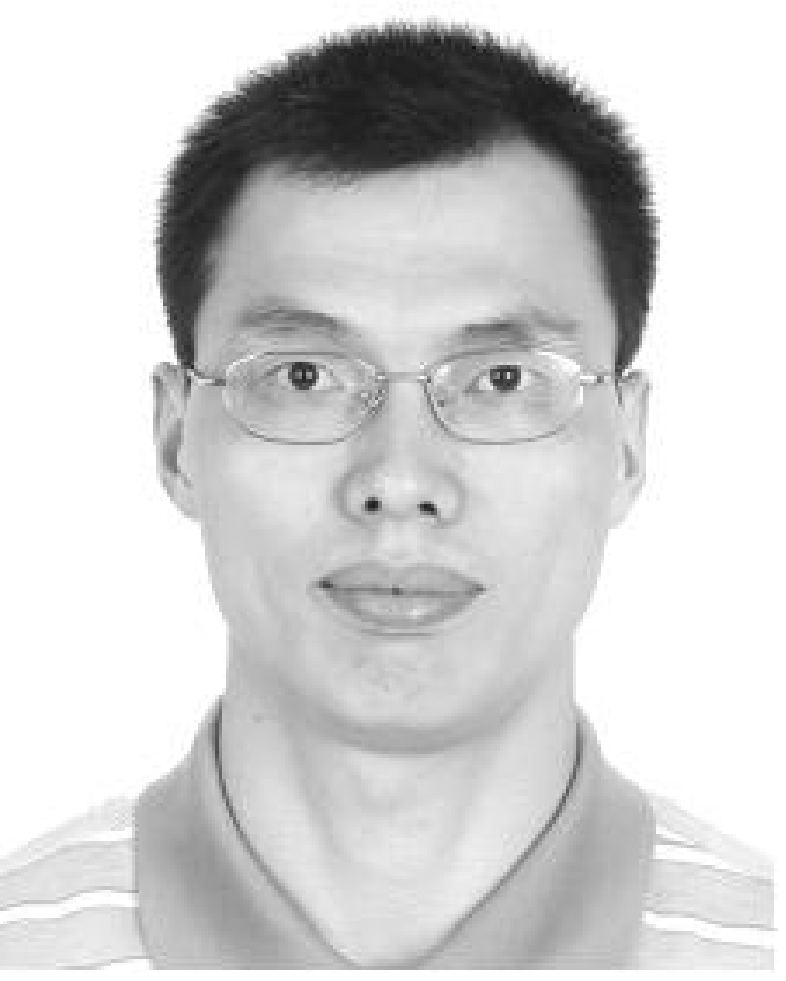}}]{Tianrui Li} (SM'11) received the B.S., M.S., and Ph.D. degrees in traffic information processing and control from Southwest Jiaotong University, Chengdu, China, in 1992, 1995, and 2002, respectively. He was a Post-Doctoral Researcher with Belgian Nuclear Research Centre, Mol, Belgium, from 2005 to 2006, and a Visiting Professor with Hasselt University, Hasselt, Belgium, in 2008; University of Technology, Sydney, Australia, in 2009; and University of Regina, Regina, Canada, in 2014. He is currently a Professor and the Director of the Key Laboratory of Cloud Computing and Intelligent Techniques, Southwest Jiaotong University. He has authored or co-authored over 150 research papers in refereed journals and conferences. His research interests include big data, cloud computing, data mining, granular computing, and rough sets. Dr. Li is a fellow of the International Rough Set Society.
\end{IEEEbiography}
\end{document}